\documentclass[acmlarge,screen=true]{acmart}

\usepackage{textcase}
\makeatletter
\g@addto@macro\@secfont{\MakeTextUppercase}
\makeatother
\usepackage{etoolbox}
\apptocmd{\thebibliography}{\interlinepenalty=10000\relax}{}{}

\usepackage{subfigure}
\usepackage[ruled,linesnumbered]{algorithm2e}
\usepackage{url}
\usepackage{caption}
\usepackage{graphicx}
\usepackage{float}
\usepackage{multirow} %
\usepackage{gensymb}
\usepackage{tabularx}
\usepackage{makecell}
\usepackage{enumitem}
\usepackage{diagbox}
\usepackage{utfsym}
\usepackage{tcolorbox}
\AtBeginDocument{%
  }

\setcopyright{cc}
\setcctype[4.0]{by-nc-nd}
\copyrightyear{2026}
\acmYear{2026}
\acmDOI{10.1145/3831638}
\acmJournal{IMWUT}
\acmVolume{10}
\acmNumber{3}
\acmArticle{159}
\acmMonth{9}

\begin{document}
\title{RF-HOI: Recognize Human-Object Interaction with Radio Frequency Signals}

\author{Lihao Wang}
\affiliation{%
  \institution{Johns Hopkins University}
  \city{Baltimore}
  \state{Maryland}
  \country{USA}
  }
\email{lwang231@jhu.edu}
\orcid{0000-0001-9290-491X}

\author{Linlu Gao}
\affiliation{%
  \institution{Johns Hopkins University}
  \city{Baltimore}
  \state{Maryland}
  \country{USA}
  }
\email{lgao25@jh.edu}
\orcid{0009-0004-9938-707X}

\author{Jiacan Yu}
\affiliation{%
  \institution{Johns Hopkins University}
  \city{Baltimore}
  \state{Maryland}
  \country{USA}
  }
\email{jyu197@jh.edu}
\orcid{0009-0005-6226-3193}

\author{Yanyu Lin}
\affiliation{%
  \institution{Johns Hopkins University}
  \city{Baltimore}
  \state{Maryland}
  \country{USA}
  }
\email{ylin185@jh.edu}
\orcid{0009-0004-3568-3849}

\author{Yifan Yin}
\affiliation{%
  \institution{Johns Hopkins University}
  \city{Baltimore}
  \state{Maryland}
  \country{USA}
  }
\email{yyin34@jhu.edu}
\orcid{0009-0008-5836-2342}

\author{Jianxin Wang}
\affiliation{%
  \institution{Johns Hopkins University}
  \city{Baltimore}
  \state{Maryland}
  \country{USA}
  }
\email{jwang616@jh.edu}
\orcid{0009-0008-4646-4819}

\author{Tianmin Shu}
\affiliation{%
  \institution{Johns Hopkins University}
  \city{Baltimore}
  \state{Maryland}
  \country{USA}
  }
\email{tianmin.shu@jhu.edu}
\orcid{0000-0001-5734-0653}

\author{Renjie Zhao}
\authornote{Corresponding author.}
\affiliation{%
  \institution{Johns Hopkins University}
  \city{Baltimore}
  \state{Maryland}
  \country{USA}
  }
\email{rjzhao@jhu.edu}
\orcid{0000-0003-2539-8580}

\renewcommand{\shortauthors}{Wang et al.}
\authorsaddresses{Authors' Contact Information: \href{https://orcid.org/0000-0001-9290-491X}{Lihao Wang}, Johns Hopkins University, Baltimore, Maryland, USA, lwang231@jhu.edu; \href{https://orcid.org/0009-0004-9938-707X}{Linlu Gao}, Johns Hopkins University, Baltimore, Maryland, USA, lgao25@jh.edu; \href{https://orcid.org/0009-0005-6226-3193}{Jiacan Yu}, Johns Hopkins University, Baltimore, Maryland, USA, jyu197@jh.edu; \href{https://orcid.org/0009-0004-3568-3849}{Yanyu Lin}, Johns Hopkins University, Baltimore, Maryland, USA, ylin185@jh.edu; \href{https://orcid.org/0009-0008-5836-2342}{Yifan Yin}, Johns Hopkins University, Baltimore, Maryland, USA, yyin34@jhu.edu; \href{https://orcid.org/0009-0008-4646-4819}{Jianxin Wang}, Johns Hopkins University, Baltimore, Maryland, USA, jwang616@jh.edu; \href{https://orcid.org/0000-0001-5734-0653}{Tianmin Shu}, Johns Hopkins University, Baltimore, Maryland, USA, tianmin.shu@jhu.edu; \href{https://orcid.org/0000-0003-2539-8580}{Renjie Zhao} (corresponding author), Johns Hopkins University, Baltimore, Maryland, USA, rjzhao@jhu.edu.}
\newcommand{\sysname}{\textit{\textsc{RF-HOI}}\xspace}
\newcommand{\etal}{\textit{et al}., }
\newcommand{\ie}{\textit{i}.\textit{e}., }
\newcommand{\eg}{\textit{e}.\textit{g}.\ }
\definecolor{responseblue}{RGB}{66, 109, 181}
\newcommand{\edited}{\color{black}}
\newcommand{\minor}{\color{black}}

\begin{abstract}
Recognizing Human-Object Interactions (HOI) is essential for intelligent systems, underpinning applications in virtual and augmented reality, embodied AI, and assistive robotics. However, vision-based HOI methods face challenges in privacy concerns and poor light conditions.
In this work, we introduce \sysname{}, the first framework that only uses radio frequency (RF) signals for HOI recognition. 
A key challenge of \sysname{} is that single-modality RF sensing is insufficient to recognize both actions and the objects being interacted with. \sysname{} addresses this through a novel modality fusion that combines mmWave radar and RFID, enabling simultaneous action recognition and target identification. Another challenge is limited training data across diverse setups, which impairs the generalizability of the recognition model. To overcome this, we develop a simulator that synthesizes multimodal RF data for diverse HOIs at scale, allowing us to fine-tune with only a small amount of real-world data. Experiment results show that \sysname{} outperforms all baselines, approaching vision model performance, and that our diverse synthetic training data can significantly boost our system's performance on real-world scenarios. These results highlight the potential of multimodal RF sensing for robust and privacy-preserving HOI recognition as well as the effectiveness of our RF data synthesis.

\end{abstract}


\maketitle

\section{Introduction}

\begin{figure*}[t]
  \centering
\includegraphics[width=\linewidth]{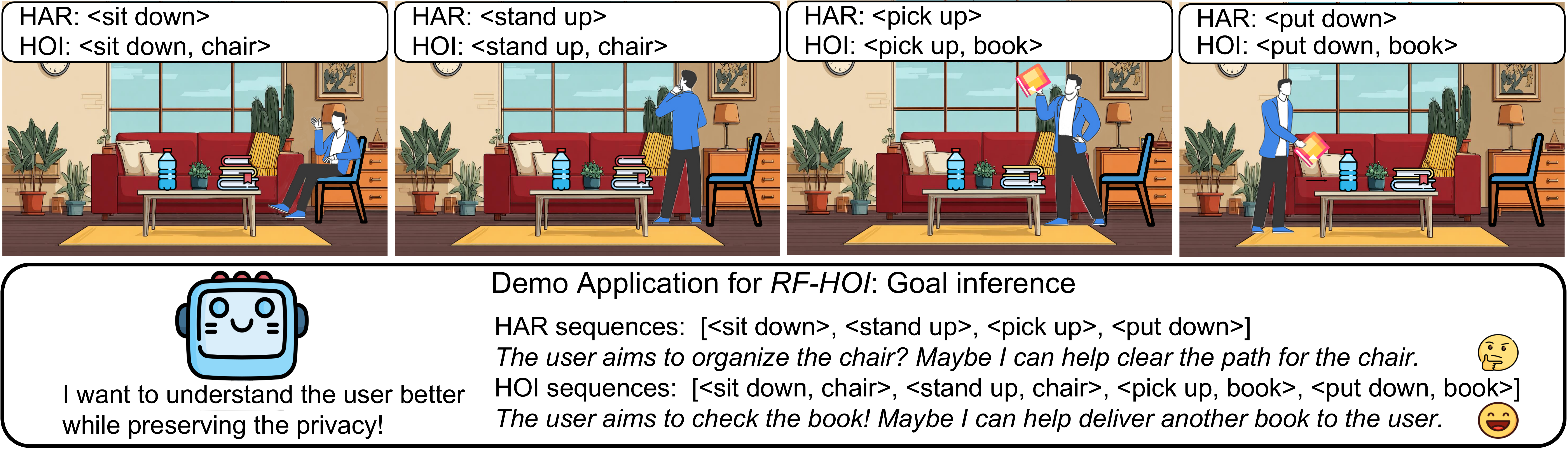}
  \caption{Using RF signals, \sysname{} can understand the user better in a privacy-preserving way by recognizing the user's interaction with objects. Compared with HAR, HOI recognition considers both human actions and target objects. As is shown in the figure, one of the applications of HOI recognition are the goal inference task. The agent can infer a more appropriate goal of the user based on HOI sequences, and then provide assistance correspondingly. \label{fig:intro}}
  \Description{\sysname{} can recognize the user action and identify the target at the same time, by fusing data from mmWave radar and RFID signals.}
\end{figure*}

Recognizing Human-Object Interactions (HOI) is foundational to human activity understanding tasks such as action anticipation \cite{mascaro2023hoiabot} and goal inference \cite{pei2011parsing,puig2023nopa,zhang2025autotom,ying2025siftom}. This capability is crucial for intelligent systems that are designed to interact with humans, including virtual agents in Virtual Reality (VR) \cite{holl2018efficient,canales2020performance} and Augmented Reality (AR) \cite{jain2023ubi}, and assistive robots \cite{savva2019habitat,puig2023nopa}. HOI recognition specifically seeks to \textit{simultaneously classify the human action and identify the interaction target}—the object being manipulated \cite{gupta2009observing,yao2010modeling,chao2015hico,yang2024open}. This joint recognition provides more information than the widely studied Human Activity Recognition (HAR), which considers only the action. For example, in Fig.~\ref{fig:intro}, when a user picks up a book, HOI recognizes a full tuple <pick up, book>, whereas HAR outputs only the action <pick up>.

While advances in computer vision have spurred extensive research on image- and video-based HOI recognition \cite{gupta2009observing,yao2010modeling,chao2015hico,liao2020ppdm,yang2024open,li2024disentangled}, these approaches face significant real-world challenges. Specifically, vision-based methods are constrained by privacy concerns \cite{zhao2025visual}, degraded performance in poor or variable lighting \cite{feng2024difflight}, and challenges in distinguishing visually similar objects \cite{chen2019destruction}. Such issues hinder their long-term and ubiquitous deployment, particularly in privacy-sensitive environments like bathrooms or bedrooms, or in variable lighting throughout the day. In contrast, recent advances in wireless sensing have demonstrated that Radio Frequency (RF) signals can enable effective user action classification \cite{singh2019radhar,zhang2021widar3,liu2022mtranssee,yang2023slnet,zhao2023cubelearn,cao2024mmclip,wang2024xrf55} and target identification \cite{pradhan2017rio,gao2019livetag,liu2021rfid,waghmare2023z}, while remaining robust to lighting conditions and maintaining user privacy. 

{\edited
Further, driven by the adoption of commercial products, wireless sensing systems increasingly have the potential to leverage existing infrastructure, thereby reducing deployment effort and hardware cost. For example, mmWave radars have been widely integrated into off-the-shelf smart home products by companies such as Xiaomi~\cite{mmWaveApp1} and Logitech~\cite{mmWaveApp2}.
In parallel, many real-world environments already deploy RFID tags as part of existing systems. In retail settings, retailers such as H\&M~\cite{RFIDApp1} and Decathlon~\cite{RFIDApp2} routinely attach RFID tags to individual products to support inventory management, customer assistance, and fast checkout. Beyond retail, RFID-tagged personal items are also common in commercial healthcare and assisted living environments~\cite{RFIDApp3,RFIDApp4}, where medication containers, daily necessities, and assistive tools are tagged for tracking and safety purposes. RFID tags have further been adopted in smart home environments~\cite{RFIDApp5,RFIDApp6}, where reusable containers or packages can be instrumented to support activity monitoring and context-aware assistance. 
Moreover, RFID readers can be implemented in lightweight and wearable forms, such as wrist-worn \cite{RFIDwearable} or hand-held \cite{RFIDhand} devices, which have been adopted in practical retail \cite{RFIDApp8} and healthcare \cite{RFIDApp7} deployments.
Building on these existing deployments, our system leverages RFID-tagged objects to enable  HOI recognition, reducing the need for additional sensing infrastructure while providing reliable object identification.
 }

In light of the above, we propose \sysname{}, the first exploration of HOI recognition using only RF signals.  In a privacy-preserving manner, \sysname{} focuses on single-user HOI recognition with commercial devices, offering a better user understanding in indoor scenarios.
Despite these advances, applying wireless sensing techniques to HOI recognition is non-trivial, since it faces two challenges.

\begin{figure*}[t]
  \centering
\includegraphics[width=\linewidth]{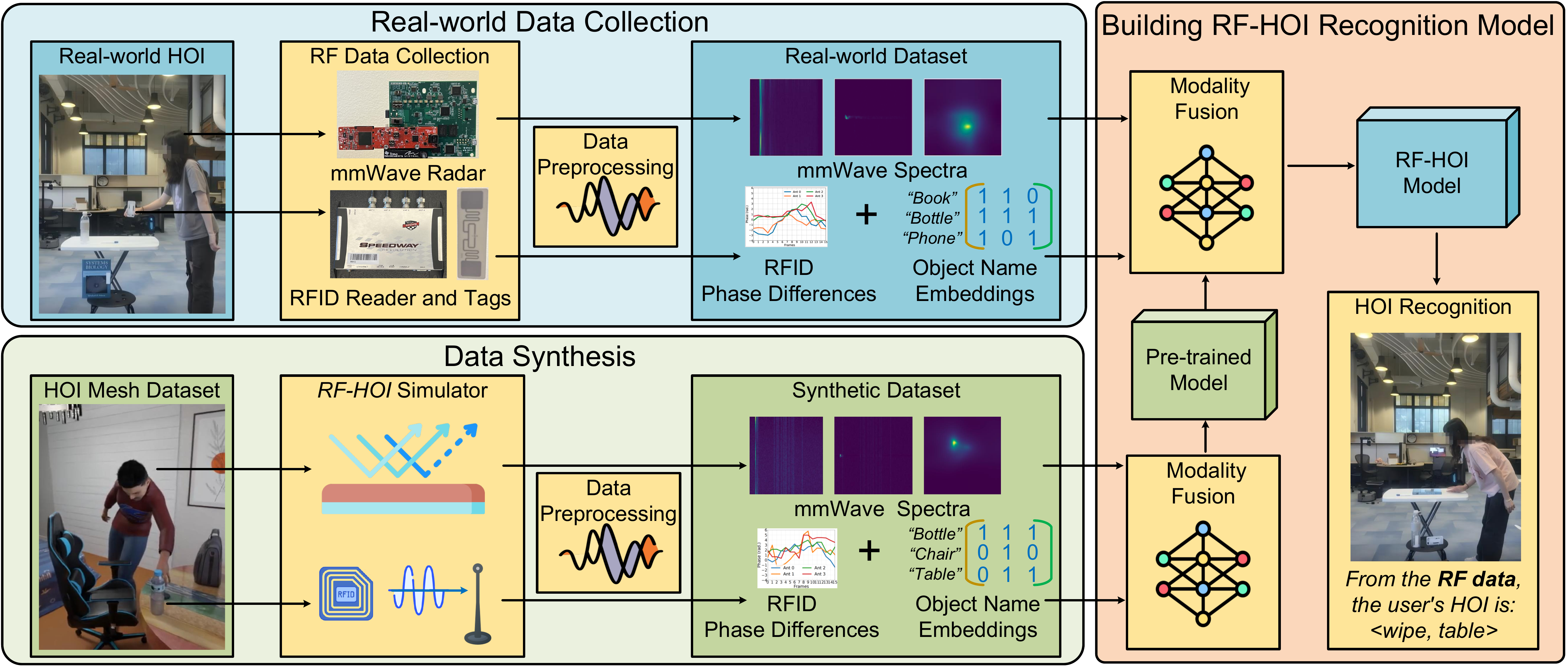}
  \caption{Overview of \sysname{} Framework. 
\label{fig:intro-overview}}
\vspace{-0.6cm}
  \Description{Overview of \sysname{} Framework. \sysname{} only uses RF signals to recognize HOI for indoor deployment. Leveraging the advantages of the mmWave radar and
RFID modalities, \sysname{} can recognize human actions and identify the target object at the same time, without assuming any devices on the user. To improve generalizability, \sysname{} also proposes a simulator that can synthesize diverse, high-quality data for both mmWave radar and RFID.}    
\end{figure*}

(1) \textit{Single modality design cannot be directly extended for HOI tasks.}
As to be detailed in Sec.~\ref{subsec:single_mod}, although single modality can address a subtask of HOI like using mmWave radar for HAR or RFID for target identification, none of them can address the tasks at the same time. %
Furthermore, naively combining the two modalities is insufficient for HOI because it will miss the mapping between the action and objects.
To grapple with these challenges, we propose a multimodal system that organically combines the advantages of the mmWave radar and RFID modalities for the HOI tasks. As illustrated in Fig. \ref{fig:intro-overview}, each candidate object is attached with an RFID tag, while the human does not wear any devices. We use an mmWave radar to collect signal changes introduced by human actions and an RFID reader to collect the signal changes of all RFID tags. %
 We devise a deep neural network to effectively fuse the two modalities. The model first applies modality-specific encoders to extract spatiotemporal features within each modality. Because HOI durations vary across scenarios, we use transformer-based temporal blocks to handle variable-length sequences. Then the model performs cross-modal fusion and decouples the output of actions and objects with two separate branches for better generalizability and interpretability.

(2) \textit{HOI recognition introduces much more costs on data collection.} 
To date, there is no dataset with object annotations for RF-only HOI recognition. Building such a dataset is costly for two reasons. First, the number of HOI categories far exceeds those in HAR \cite{singh2019radhar,liu2022mtranssee,wang2024xrf55}, inflating annotation costs \cite{li2024disentangled}. 
Second, for model generalizability, wireless sensing datasets have to cover diverse setups (e.g., user orientation, distance, body shape, and environment), which is especially important for HOI recognition since both actions and targets are affected by setups. Consequently, it is prohibitively expensive to assemble a sufficiently diverse, well-annotated real-world dataset for RF-based HOI recognition.  
To address this challenge, we first analyze the impacts of various factors to guide future large-scale data collection.  To improve model performance and generalizability with limited real data, we develop an RF signal simulator that generates synthetic HOI data using well-annotated 3D-mesh traces from recent works in computer vision\cite{jiang2023full, jiang2024scaling, jiang2024autonomous}. This enables the generation of diverse synthetic data, including 7,518 well-annotated multimodal samples. 
{\edited We highlight that the simulator enables the generation of synthetic data with expanded object categories. When these categories exhibit interaction patterns similar to those seen during training, a recognition model pre-trained on such synthetic data can generalize to them, even though they are not included in real-world fine-tuning.}
Finally, for evaluation purposes, we build a data collection platform and collect 3,615 real-world samples across 78 distinct setups.
We pre-train the proposed modality-fusion model on synthetic data and fine-tune it on a small subset of real-world samples to bridge the sim-to-real gap.

We conduct comprehensive evaluations using the collected real-world data. Results show that \sysname{} achieves an average accuracy of 96.97\%, with only a 1.96\% performance gap compared to the vision baselines. \sysname{} outperforms its non-fusion variant by 21.29\%, RFID-only by 32.90\%, and mmWave-only by 73.20\%. %
Moreover, incorporating synthetic data boosts performance by an average of 12.00\% across various real-world training set sizes and by 12.09\% across different numbers of seen setups.

Our contributions can be summarized as follows:
\begin{itemize}
    \item To the best of our knowledge, we are the first to address HOI recognition only using RF signals, providing a privacy-preserving alternative to vision-based methods. %
    \item We introduce a novel multimodal approach that combines mmWave radar and RFID sensing for HOI recognition. We design a modality fusion network that processes spatiotemporal features and outputs decoupled HOI recognition.
    \item {\edited We develop a multimodal simulator that can synthesize diverse, high-quality RF data, enabling improved HOI recognition performance and supporting generalization to unseen object categories with similar interaction patterns, especially when real-world RF data is limited.}
    \item We build a real-world prototype using commercial devices for experiments across different distances, orientations, environments, and users. Extensive evaluations demonstrate the effectiveness and generalizability of \sysname{} in the real world.
\end{itemize}

\section{Challenges and Motivation}
\subsection{Deficiency in Single Modality}\label{subsec:single_mod}

\begin{table*}
\centering
\caption{Compare different sensing settings. RF-HOI fuses both mmWave and RFID modalities. \label{tab:singlemodal}}
\begin{tabular}{cccccc}
\toprule
\makecell{Modality} & \makecell{Device assistance} & \makecell{Fine-grained \\ action capture} & \makecell{Target \\ identification} & \makecell{User \\ convenience} & \makecell{HOI \\ recognition} \\
\hline
\multirow{2}{*}{mmWave} 
    & Device-free HAR \cite{singh2019radhar,wang2021m,liu2022mtranssee,zhao2023cubelearn,cao2024mmclip}    & \textbf{Yes} & Low & \textbf{High} & Low \\
\cline{2-6}
    & Tags on objects \cite{soltanaghaei2021millimetro, bae2024supersight,lu2023millimeter,bae2022omniscatter, li2019ferrotag} & No & Medium & \textbf{High} & Low\\
\hline
\multirow{3}{*}{RFID} 
    & Tags as background  \cite{wang2018modeling,huang2019id,sun2024lodihar,wang2024xrf55}                       & No & Low & \textbf{High} & {Low}\\
\cline{2-6}
    & Tags on objects  \cite{li2015idsense,spielberg2016rapid,he2016deep,zhang2019shopeye}        & No & Medium & \textbf{High} & Low \\
\cline{2-6}
    & Tags on humans \cite{wang2016toward,jin2018towards,yang2022environment,wang2025generative} & \textbf{Yes} &  Low & Low & Low \\
\hline
RF-HOI& Tags on objects               & \textbf{Yes} & \textbf{High} & \textbf{High} & \textbf{High} \\
\bottomrule
\end{tabular}
\end{table*}
 
For effective HOI recognition, a sensing system must \textit{capture fine-grained action features}, \textit{identify the target object}, and \textit{offer user convenience}. However, as summarized in Table~\ref{tab:singlemodal}, widely used single-modality approaches—such as mmWave-only or RFID-only—fall short of meeting all these requirements.
 
\textbf{mmWave suffers for target identification.} While mmWave radar, like vision-based approaches, can capture fine-grained action features in a convenient device-free manner, it is largely limited to HAR tasks \cite{singh2019radhar,wang2021m,liu2022mtranssee,zhao2023cubelearn,cao2024mmclip,lin2026active,monjur2026mmweaver}. Target identification is challenging for mmWave in device-free settings, as commercial mmWave radars lack the resolution to directly distinguish objects \cite{adhikari2022mishape}. Even advanced solutions with larger radar arrays \cite{zhang2020mmeye} or metasurfaces \cite{wang2025high} typically infer object identity from contours and can only handle one object at a time \cite{chen2022target,he2023fusang}, making them unsuitable for HOI scenarios with multiple candidate objects.  Attaching mmWave tags to objects \cite{li2019ferrotag, soltanaghaei2021millimetro,lu2023millimeter,bae2022omniscatter, bae2024supersight} may support object identification via tag ID and localization, and could be an option in the future with hardware improvements and commercialization. At present, however, these solutions are impractical because they require specialized hardware design and their tags may be expensive or require external energy. 

\begin{figure}[t]
\centering
\subfigure[The photo of <pick up, phone>]{
\includegraphics[width=0.4\linewidth]{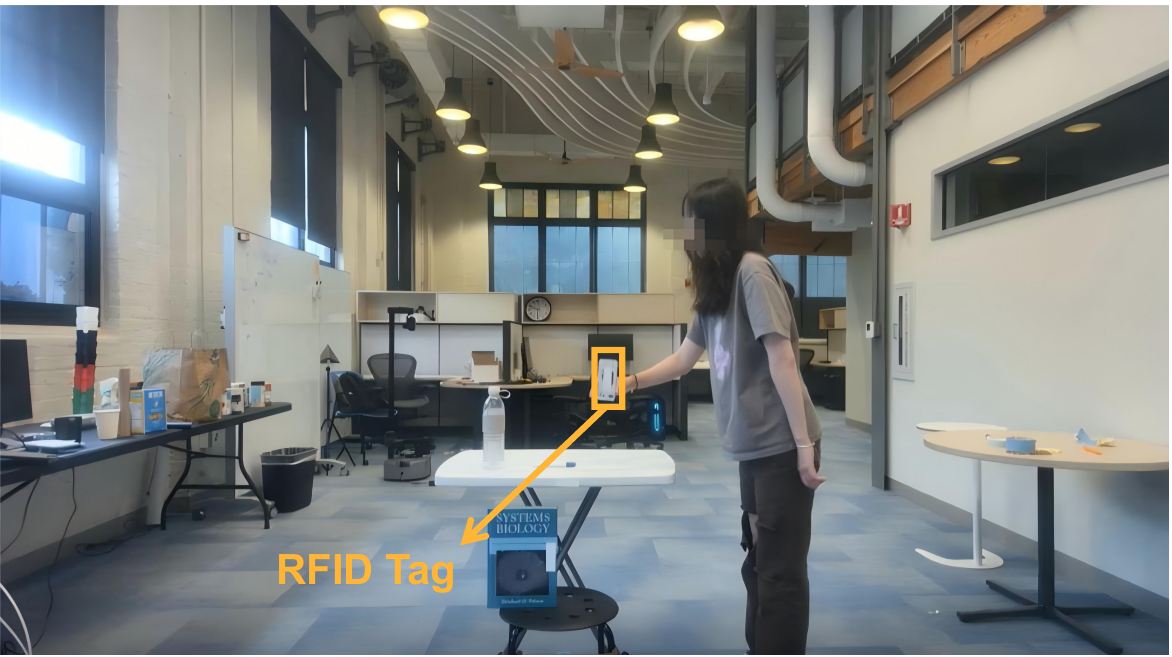} 
}
\subfigure[The photo of <wipe, table>]{
\includegraphics[width=0.4\linewidth]{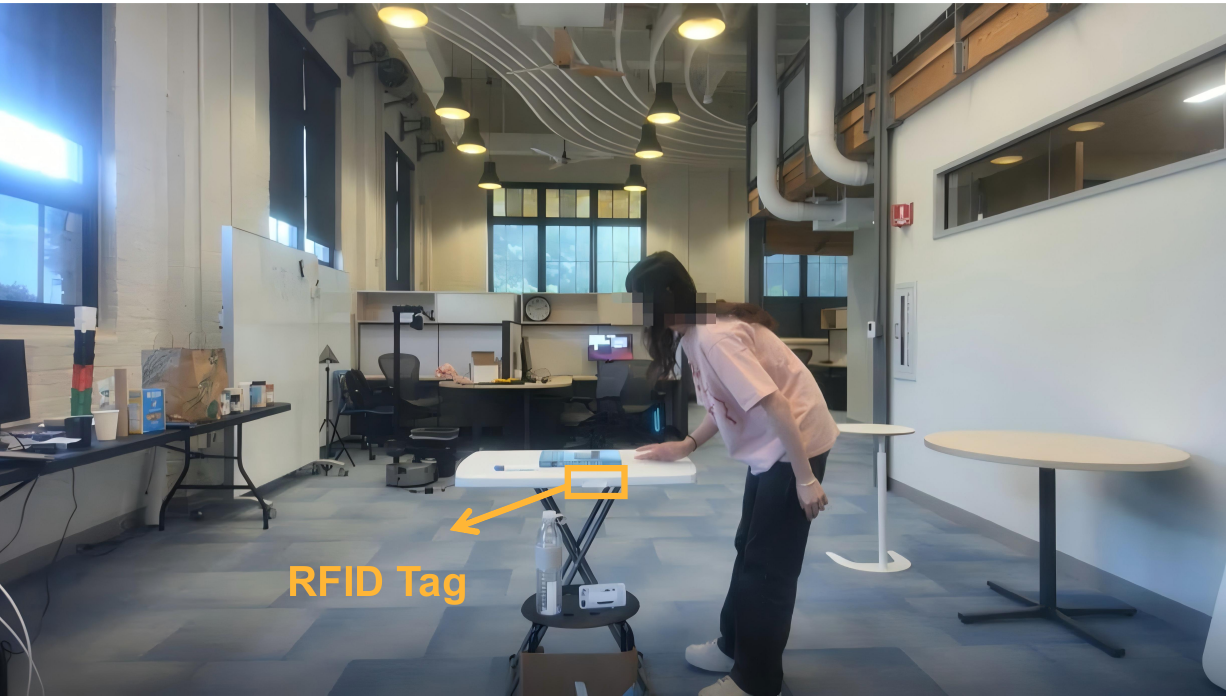} 
}
\\
\subfigure[The tag attached to the phone in <pick up, phone>]{
\includegraphics[width=0.4\linewidth]{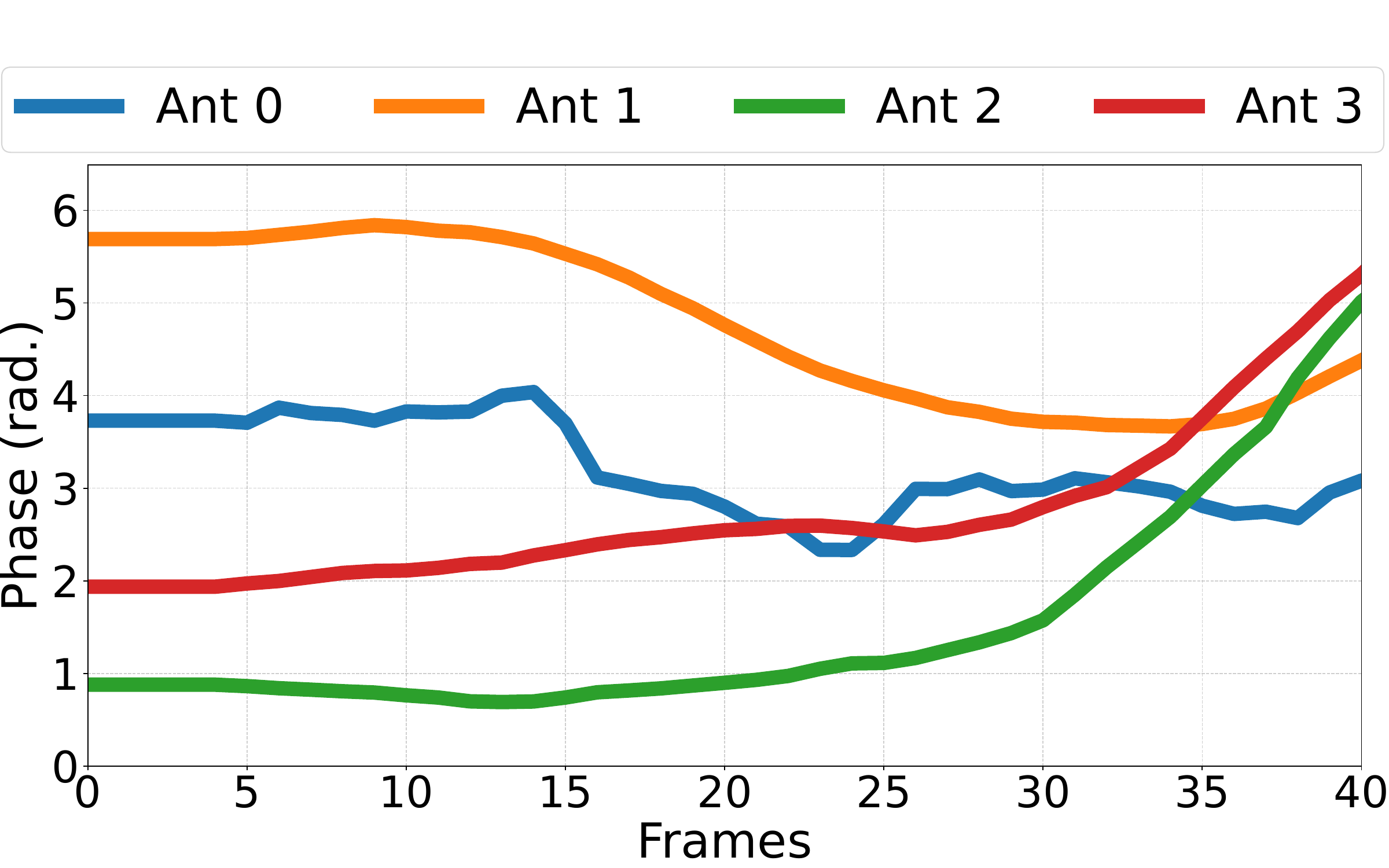} 
}
\subfigure[The tag attached to the table in <wipe, table>]{
\includegraphics[width=0.4\linewidth]{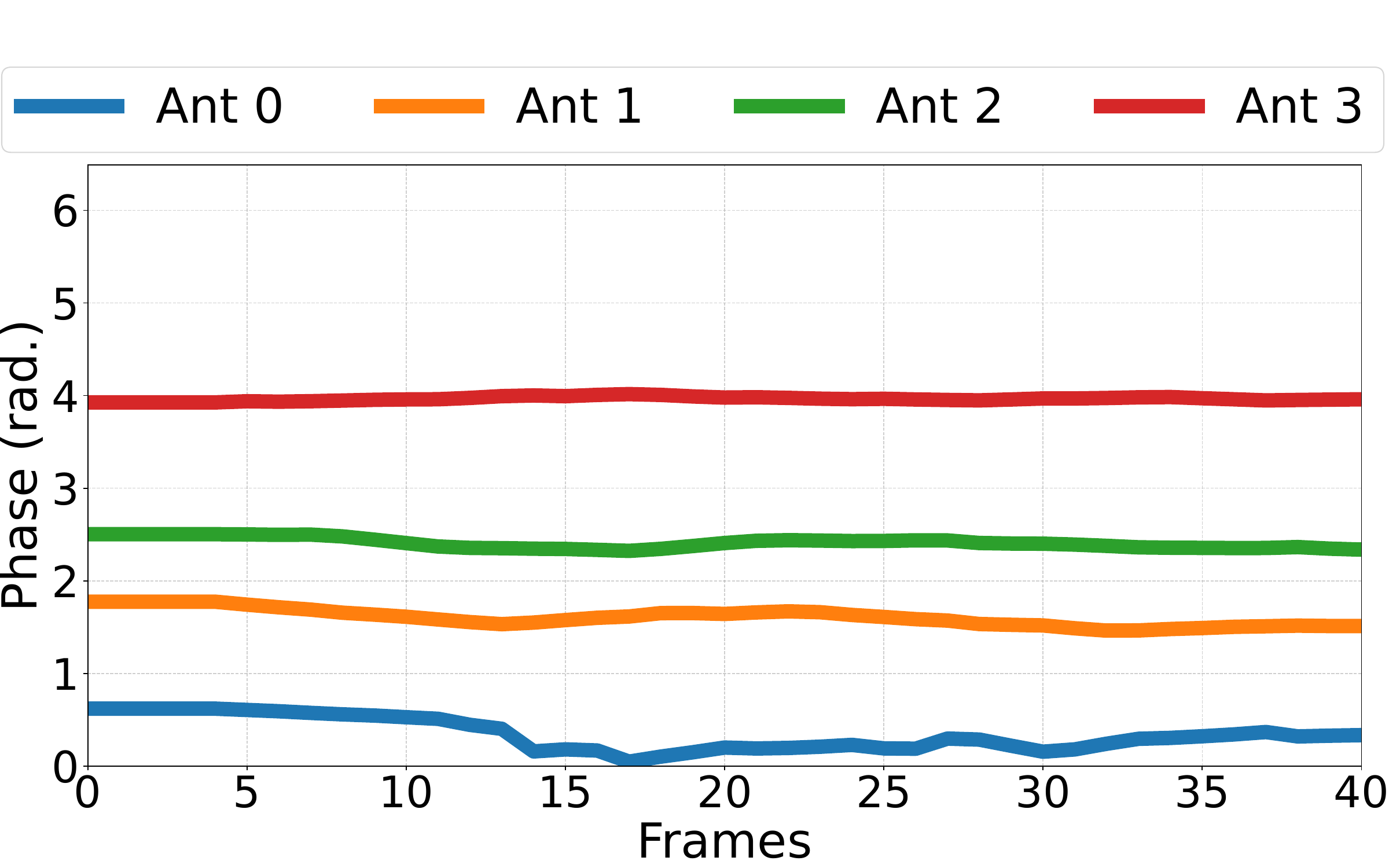} 
}
\caption{Photos and corresponding RFID signal phases of the tag attached to the target object in two different interactions. 
\label{fig:rfidphase}}
\Description{Photos and RFID signal phases of the tag attached to the target object in two different interactions.}
\end{figure}
\textbf{RFID suffers for capturing action features.} While attaching RFID tags enables target identification, RFID-based methods face significant limitations in action recognition. First, placing large arrays of RFID tags in the background \cite{wang2018modeling,huang2019id,sun2024lodihar,wang2024xrf55,feng2026rfusion} introduces a high deployment cost, has a limited update rate due to the limited RFID reading capacity, and needs to constrain the activity between reader and tags. Second, attaching tags to target objects \cite{li2015idsense,spielberg2016rapid,he2016deep,zhang2019shopeye} enables target identification only when the object is moved, but fails to capture action features or recognize interactions where the user does not touch the tag or move the object—situations common in HOI tasks. For example, as is shown in Fig.~\ref{fig:rfidphase}, using a four-antenna RFID reader, we observe distinct phase changes during the <pick up, phone> interaction when the tagged object was moved, but minimal phase variation during <wipe, table> when the tag (affixed to a stationary table) was not directly manipulated. Without discernible patterns in RFID signals, accurate action classification becomes infeasible for RFID-only approaches. Finally, although directly attaching tags to the human body \cite{wang2016toward,jin2018towards,yang2022environment,wang2025generative,wu2026rf,wang2026rfgat} can capture fine-grained actions, this approach is impractical for real-world deployment as it is inconvenient for users.

Therefore, we argue that it is essential for \sysname{} to integrate both mmWave and RFID sensing for HOI recognition—leveraging mmWave for robust action classification, RFID for accurate target identification, and their complementary contextual information to enhance overall performance. We will present our multimodal data preprocessing pipeline in Sec.~\ref{sec:preproc} and our fusion network design in Sec.~\ref{sec:network}.

\subsection{High Data Collection Cost for HOI \label{sec:whysyn}}

\sysname{} faces three key challenges in applying empirical recognition models:

\textbf{The novel RF modalities require customized data and annotation.}
Unlike prior works on HAR \cite{li2025consense, dai2025babel}, existing datasets cannot be repurposed, and there is no RF signal-based HOI dataset available. Most HOI datasets are designed for vision-based approaches and contain only images or videos. Meanwhile, existing RF-based HAR datasets typically lack object interaction scenarios and do not provide the necessary HOI annotation.

\begin{figure}[t]
    \centering
        \begin{minipage}[t]{0.4\linewidth}
        \centering
        \includegraphics[width=\linewidth]{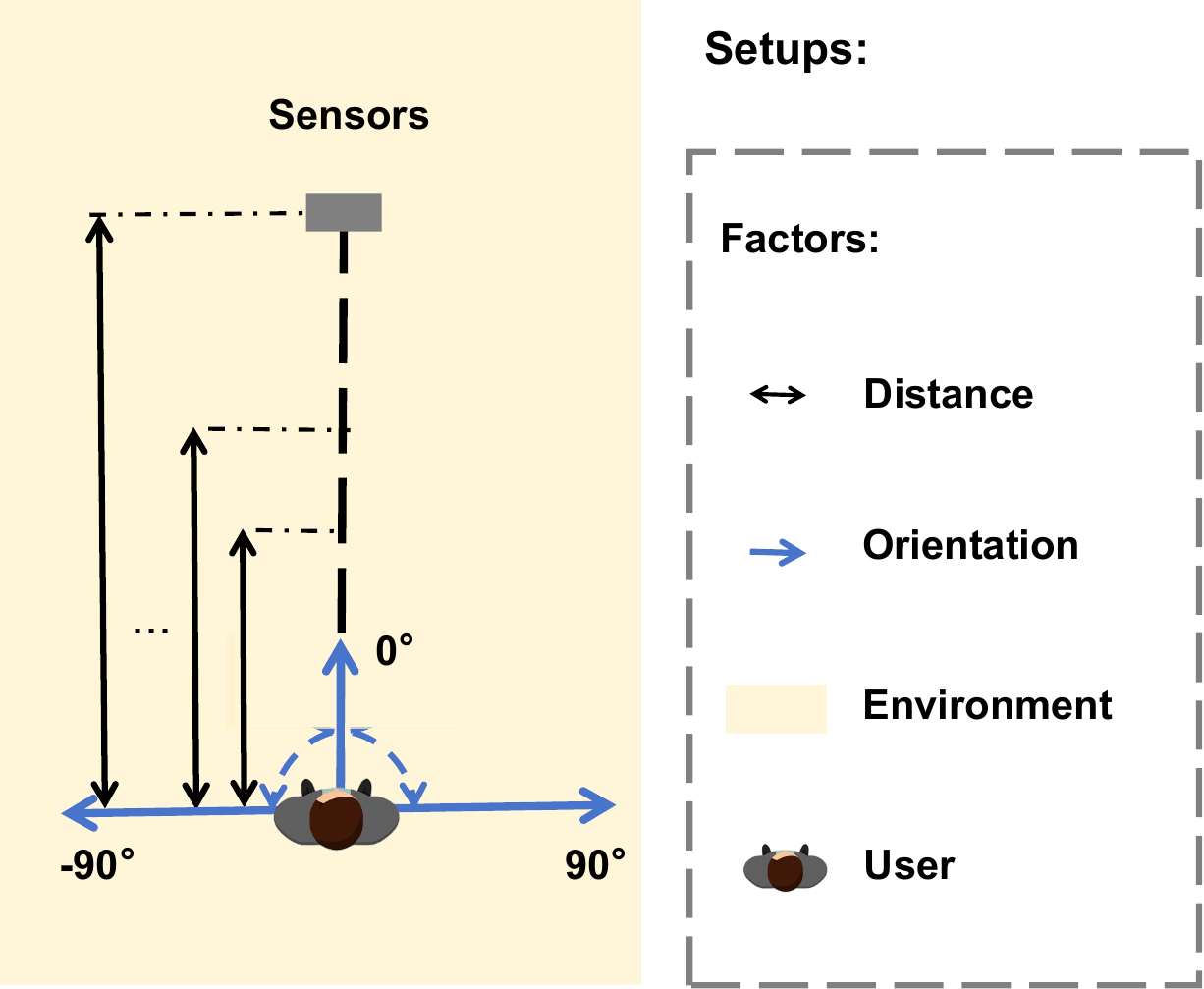}
        \caption{Factors that may affect model performance. A setup is a combination of the four factors. \label{fig:factors}}
   \Description{Definition of user orientations. }
    \end{minipage}
      \hspace{0.03\linewidth}
    \begin{minipage}[t]{0.4\linewidth}
        \centering
        \includegraphics[width=\linewidth]{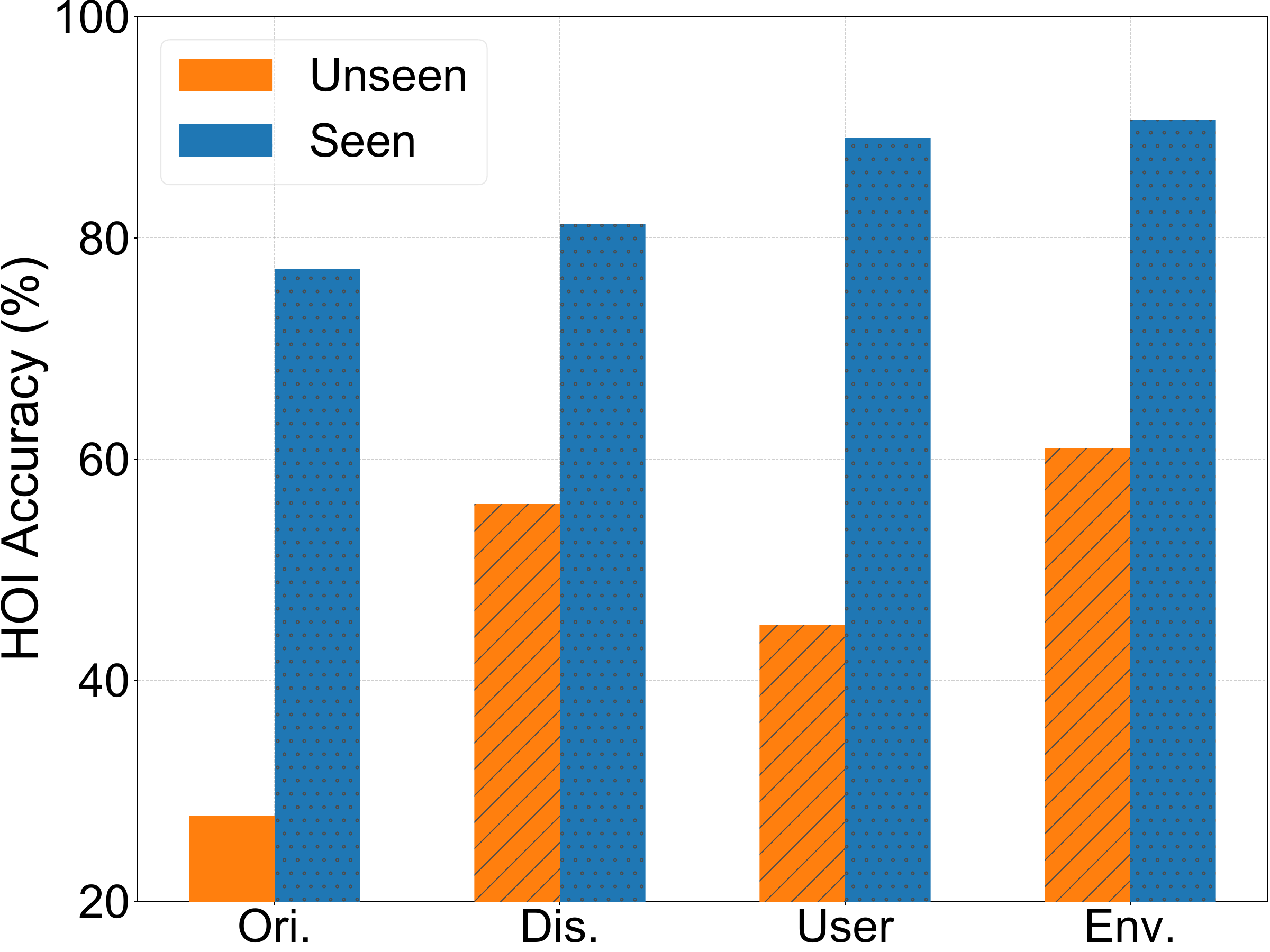}
        \caption{Impact of unseen factors (Ori: orientation, Dis: distance, Env: environment).\label{fig:exp_unseendom}}
   \Description{Impact of unseen factors.}
    \end{minipage}
\end{figure}

\textbf{Data collection needs to cover diverse setups.}
For comprehensive evaluations, data collection must include a wide range of \textit{setups} — combinations of user orientation, distance to the device, background environment, and body shape \cite{zhang2021widar3,rfgen}, as illustrated in Fig. \ref{fig:factors}. Inadequate coverage of these factors in the training set can significantly impact HOI recognition performance. 
To demonstrate this, we conducted a preliminary study using data that varies across two environments, two users, three distances, and two orientations. For each factor, we reserved all samples with one specific value as the unseen factor from the training set and used them for testing, while the baseline included all factors in training. As shown in Fig.~\ref{fig:exp_unseendom}, HOI accuracy drops by 49.41\% for unseen orientation, 25.32\% for unseen distance, 44.06\% for unseen user, and 29.69\% for unseen environment, highlighting the importance of diverse setup coverage. 

\textbf{HOI tasks introduce many categories with the combination of actions and objects.}
However, despite previous works on HAR \cite{wang2024xrf55,zhang2021widar3,cui2023milipoint}, collecting sufficient and diverse data for HOI recognition is even more challenging. Considering the same number of actions, HOI recognition usually has more categories than HAR, since one action may have multiple candidate target objects. As a consequence, we have to collect and annotate more samples if we want to cover every possible setup for each HOI category \cite{li2024disentangled}. Suppose we want to cover three environments, five users, four distances, and seven orientations for seven actions with three candidate target objects on average. Then, if we repeat each HOI category five times in each setup, the total number of samples we need to collect will be the product of the above numbers, \ie $
    3 \times 5 \times 4 \times 7 \times 7  \times 3 \times 5 = 44,100$.
In fact, this number has exceeded the size of many large-scale benchmarks for HAR, with only a few factors considered.

Given the high cost of real-world data collection, and inspired by numerous HAR studies leveraging synthetic data \cite{cao2024mmclip,rfgen,deng2023midas,deng2023midas++,ahuja2021vid2doppler,li2024sbrf}, \sysname{} adopts synthetic data to enhance model performance with limited real-world samples, thereby reducing data collection effort. We detail the simulator design and the integration of synthetic data in Sec.~\ref{subsec:method_synth}.

\section{Methods}

\subsection{Overview}

As shown in Fig. \ref{fig:intro-overview}, \sysname{} comprises three main components: real-world data collection, data synthesis, and HOI recognition modeling. We first collect mmWave and RFID data in real-world settings, followed by RF signal preprocessing through our \textbf{data preprocessing module}. Subsequently, a \textbf{modality fusion model} is designed to integrate the complementary strengths of both sensing modalities. To address the limitation of real-world data, we use our proposed \textbf{\sysname{} simulator} to generate large-scale synthetic data. The model is initially pre-trained on this synthetic data to enhance generalizability, then fine-tuned using a small subset of real-world samples, and finally evaluated in real-world scenarios.

\subsection{Data Preprocessing \label{sec:preproc}}
\sysname{} leverages both mmWave and RFID signals to identify human actions and target objects. However, these raw signals are often noisy, high-dimensional, and complex-valued, which complicates the design of suitable neural networks \cite{zhao2023cubelearn}. To address this, we introduce a dedicated data preprocessing module in \sysname{} that processes the raw signals, enabling the effective application of established model architectures such as convolutional neural networks (CNNs) and transformers \cite{vaswani2017attention}.
\subsubsection{Preprocessing mmWave Data}
\begin{figure}[t]
\centering
\subfigure[RT heatmap]{
\includegraphics[width=0.2\linewidth]{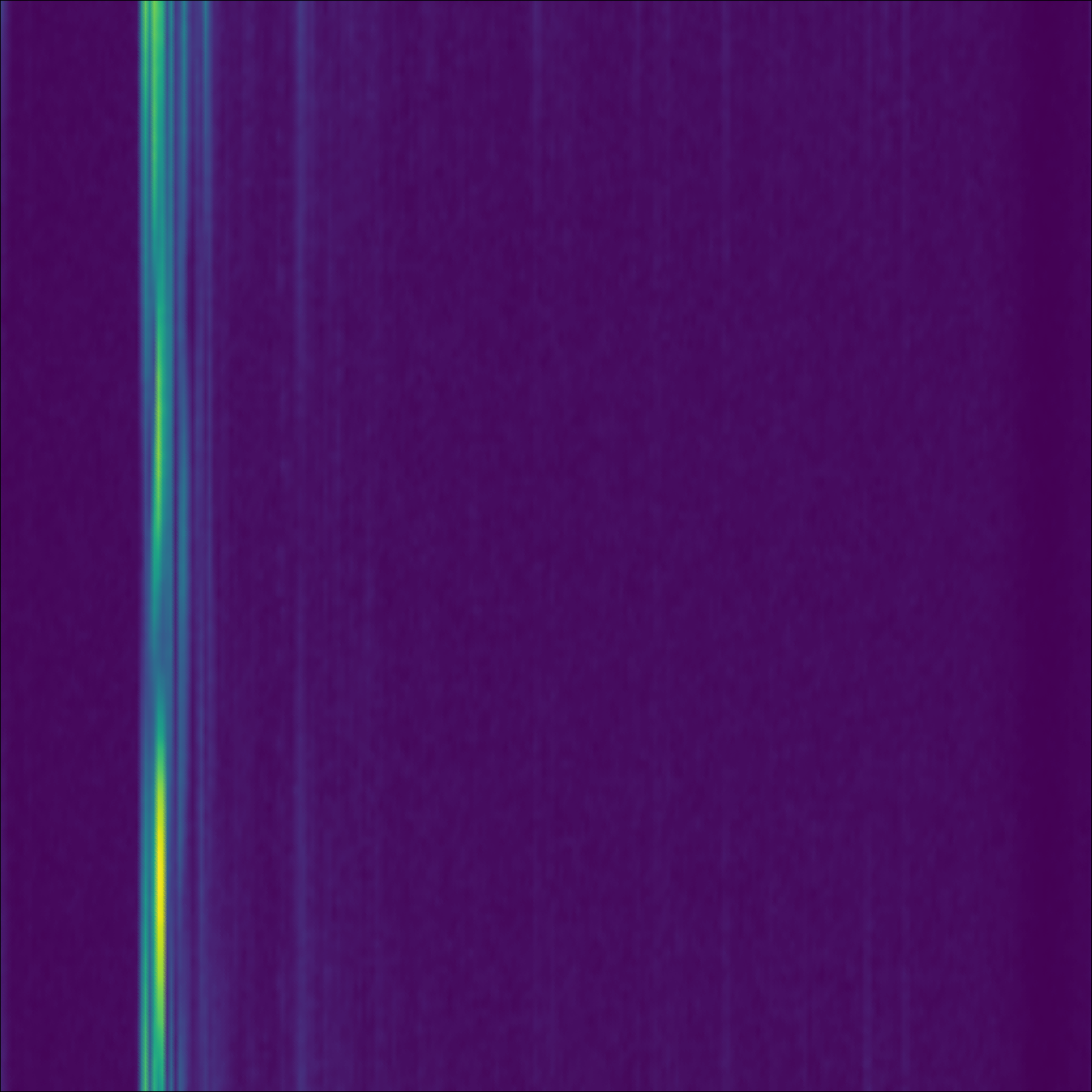 
}
}
\hspace{0.03\linewidth}
\subfigure[RD heatmap]{
\includegraphics[width=0.2\linewidth]{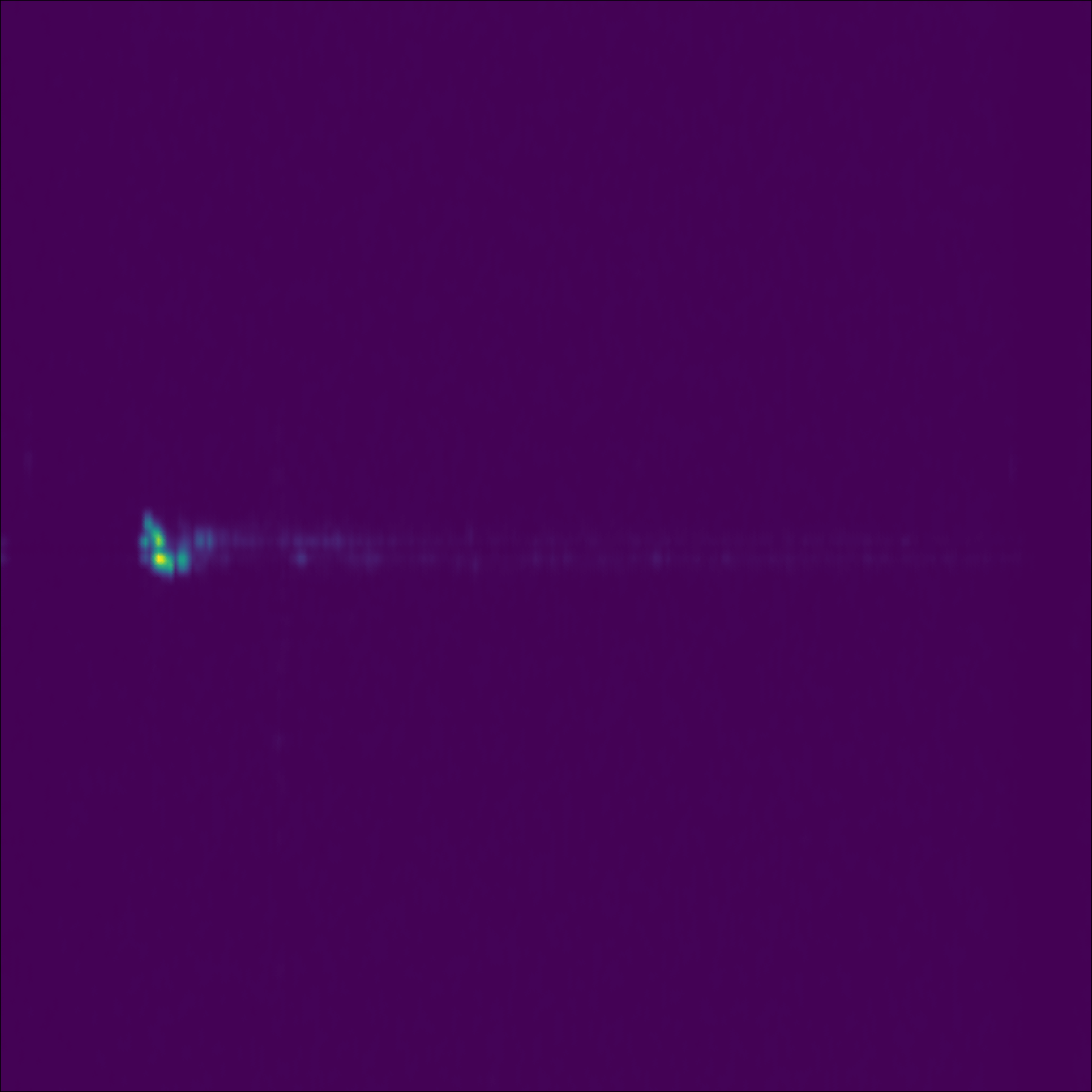
}
}
\hspace{0.03\linewidth}
\subfigure[AoA heatmap]{
\includegraphics[width=0.2\linewidth]{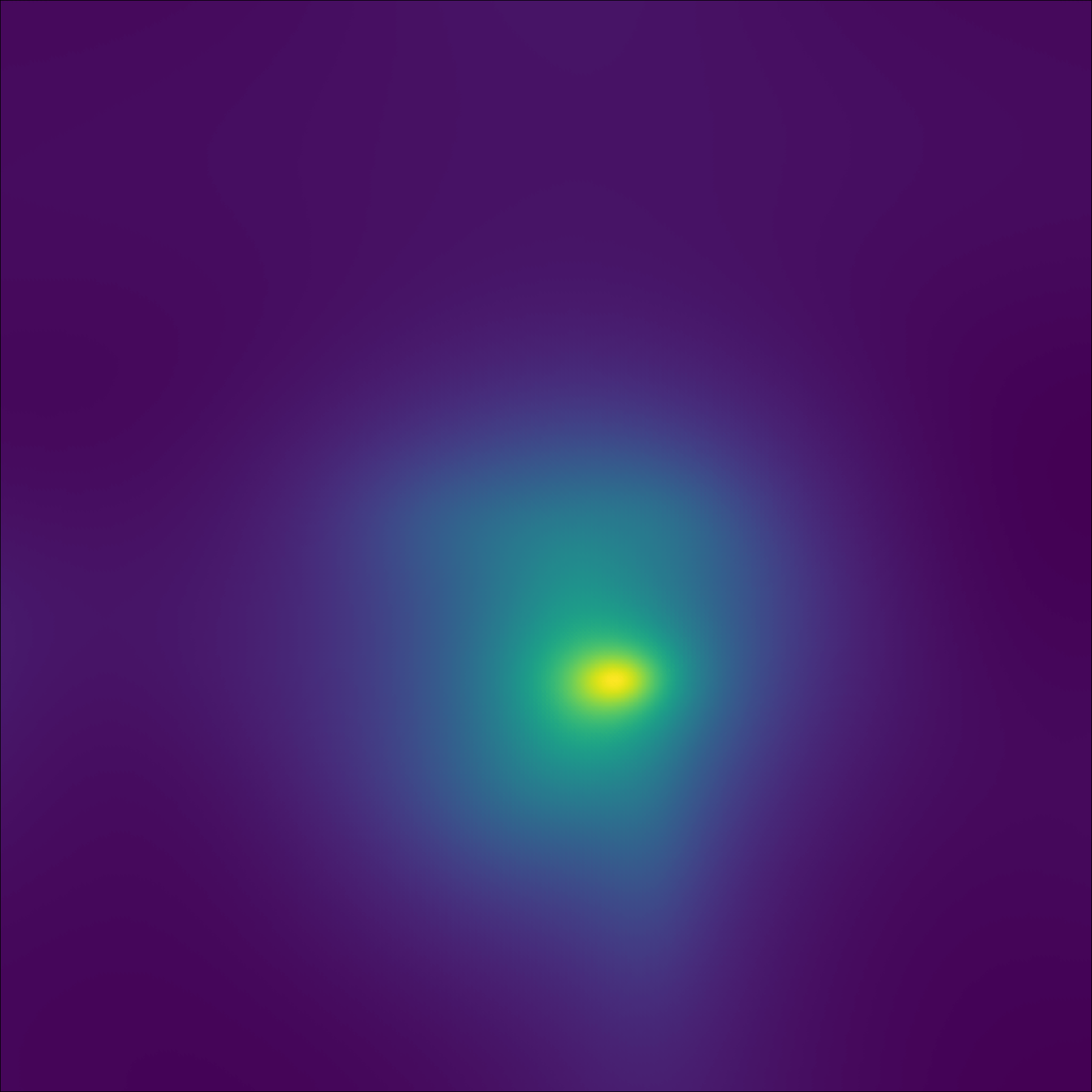 
}
}
\caption{Examples for mmWave heatmaps. \label{fig:mmSpec}}
\Description{Examples for mmWave heatmaps. }
\end{figure}
{\minor We adopt spectrum-based mmWave representations as a task-driven design for RF-HOI, using heatmaps to capture dense temporal measurements of motion-related range and velocity responses. 
We acknowledge that mmWave point clouds~\cite{singh2019radhar,xue2021mmmesh,zhang2024super} provide an alternative representation and remain a promising direction for future work.}
Specifically, the raw mmWave radar data consists of Intermediate Frequency (IF) signals with dimensions ($N_{frame}^{M}$, $N_{chirp}$, $N_{Tx}$, $N_{Rx}$, $N_{R}$),  representing the number of frames, chirps, transmitting antennas, receiving antennas, and samples per chirp, respectively. To begin, we apply a static clutter removal algorithm \cite{xue2021mmmesh} to mitigate background interference. Following standard processing pipelines \cite{iovescu2020fundamentals}, we then compute a 2D Range-Time (RT) heatmap for distance features using Range-FFT, and a 2D Range-Doppler (RD) heatmap for velocity features from each radar frame using Doppler-FFT. Finally, we extract the Angle of Arrival (AoA) heatmap using the MUSIC algorithm \cite{schmidt1986multiple}. As illustrated in Fig.~\ref{fig:mmSpec}, this process yields three 2D heatmaps per radar frame, which can be directly fed into the proposed neural network model.

\subsubsection{Preprocessing RFID Data}
Suppose each candidate object is equipped with an RFID tag. The RFID reader generates a table where each entry records the timestamp, tag ID, antenna ID, signal amplitude, and signal phase. 

From this table, we first extract phase frames for each antenna–tag pair at each time step. \sysname{} primarily utilizes the signal phase, as it is more fine-grained and reliable than amplitude \cite{magnago2019ranging,bu2020rf}. To account for the asynchronous nature of RFID readings—resulting in different frame counts across pairs—we use nearest neighbor interpolation for alignment \cite{wei2016gyro}, followed by a moving average for smoothing.
Note that modeling the RFID phase $\varphi(t)$ is challenging due to unknown multipath conditions, and accurate RFID localization typically requires expensive hardware \cite{liang2023rf}. However, since our focus is on kinematic features of the target objects, we use the phase difference between consecutive frames, $\Delta \varphi (t)$, as it effectively indicates movement direction by its sign. 

Additionally, to leverage semantic information from RFID, we extract a tag table mapping tag IDs to object names, and convert these names to word embeddings using spaCy \cite{honnibal2020spacy}. This allows us to input meaningful object semantics into the neural network.

\subsection{Modality Fusion Network \label{sec:network}}
\begin{figure}[t]
  \centering
  \includegraphics[width=0.8\linewidth]{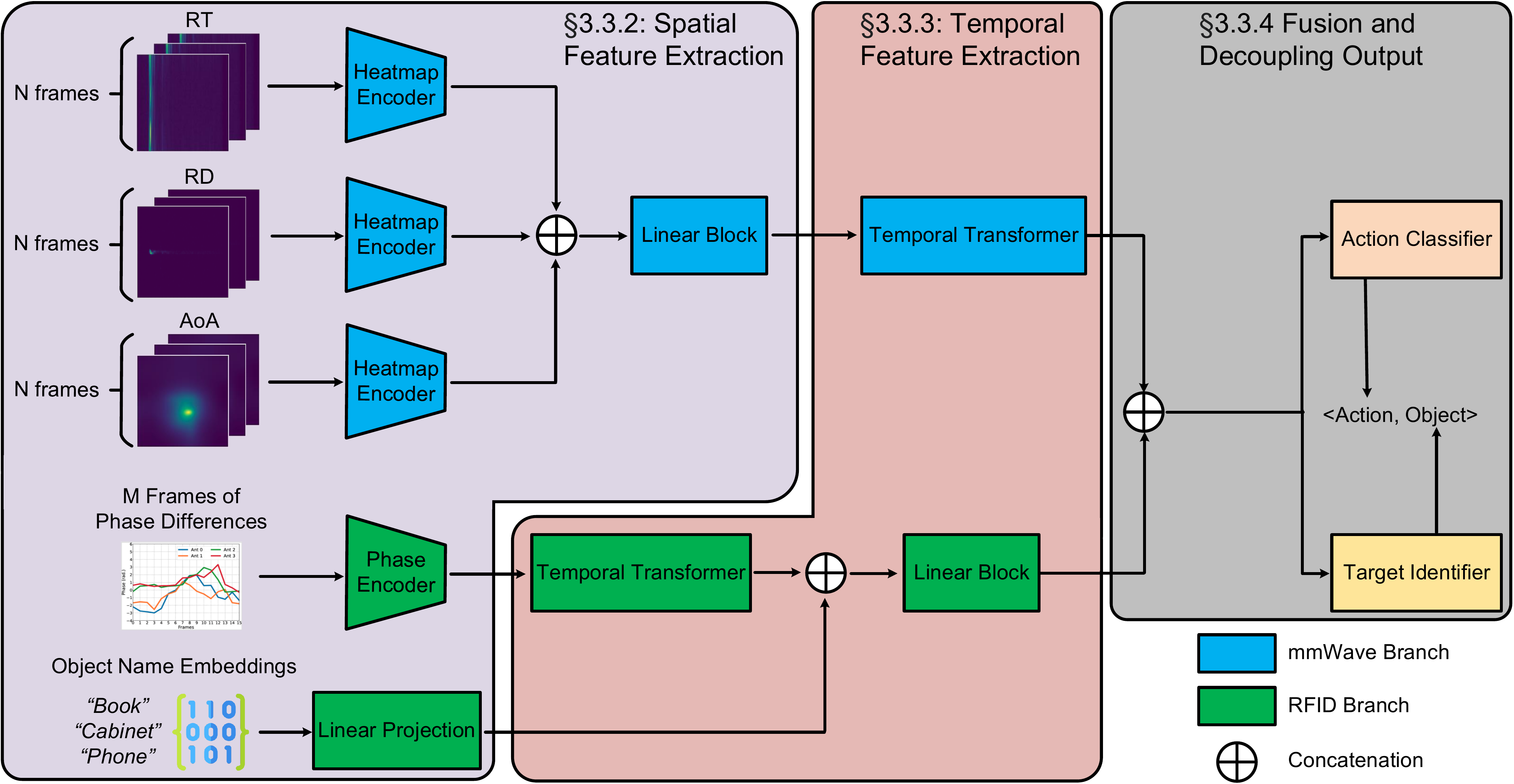}
  \caption{Structure of the modality fusion network. We carefully designs spatial and temporal feature extraction for each modality, and then fuses the two asynchronous modalities at the feature level. The outputs are decoupled by two branches to improve generalizability and interpretability.\label{fig:nn}}
  \Description{Structure of the modality fusion network. }
\end{figure}
\subsubsection{Challenges} 
For each sample, after preprocessing, the data includes a matrix of ($N_{frame}^{M}$, 3, $N_{map}$, $N_{map}$) for mmWave heatmaps, a matrix of ($N_{frame}^{R}$, $N_{tag}$, $N_{ant}$) for RFID phase differences, and object name embeddings with a shape of ($N_{tag}$, $N_{embed}$), 
where $N_{map}$ is the heatmap size, $N_{frame}^{R}$ is the number of RFID phase difference frames, $N_{tag}$ is the number of tags,  $N_{ant}$ is the antenna number, and $N_{embed}$ is the embedding dimension.  

Fusing these two modalities is challenging due to their heterogeneity in both spatial and temporal dimensions.
In the spatial dimension, mmWave data provides three heatmaps, while RFID data offers phase differences $\Delta \varphi$ for each tag-antenna pair, along with object name embeddings. The input dimensions and underlying physical meanings differ significantly, which raises challenges in how to design the model to encode the information of each modality.

In the temporal dimension, mmWave radar has a fixed frame rate, but RFID readings are asynchronous and vary over time \cite{wei2016gyro}. As a result, each radar frame may correspond to a different number of RFID frames, and real-world HOI samples may have varying lengths and frame counts. Therefore, it is difficult to process and fuse both modalities' temporal features at the frame level. 

Moreover, directly outputting HOI categories with a single branch would limit the generalizability of the model and cannot support unseen HOI categories (unseen action-object pair) when the constituent actions and objects are seen. 
For example, the model cannot recognize <pick up, book> even when it has seen <pick up, bottle> and <open, book> in the training data.
In addition, coupling the action and object in one output leads to poor interpretability since it is difficult to analyze the performance on action recognition and target identification separately \cite{zhou2022human,li2024disentangled}.

To address these challenges, we propose a fusion network, as illustrated in Fig.~\ref{fig:nn}.
We first extract spatial and temporal features independently for each modality, and then fuse these representations for joint action classification and target identification. Further, the proposed network decouples the output of actions and objects with two separate branches for a better generalizability and interpretability.
\subsubsection{Spatial Features Extraction}
After preprocessing, each radar frame contains three heatmaps of size $N_{map} \times N_{map}$. To extract features related to range, velocity, and angle, we feed each heatmap into a dedicated heatmap encoder. As illustrated in Fig.~\ref{fig:mmSpec}, each heatmap resembles an image, making convolutional neural networks (CNNs) an effective choice for pattern extraction, as CNNs excel at capturing patterns in mmWave heatmaps \cite{xue2023towards}. Each heatmap encoder consists of two single-channel CNN layers followed by a max-pooling layer. The outputs from the three encoders are then concatenated and passed through a two-layer linear block to integrate features across all heatmaps.

For the RFID branch, we obtain a matrix of shape ($N^{R}_{frame}$, $N_{tag}$, $N_{ant}$), comprising the $\Delta \varphi$ phase differences for each tag-antenna pair at each frame. We observe that the phase differences from one tag to multiple antennas form a coordinate-like vector related to the tag's position. Inspired by prior work with 3D coordinates \cite{xue2021mmmesh}, our phase encoder uses linear layers to project the input into a high-dimensional space, providing better capacity to distinguish motion patterns. To ensure stable training and effective feature normalization, we use Layer Normalization \cite{ba2016layer} instead of the commonly used Batch Normalization \cite{ioffe2015batch} within each block. It is because Layer Normalization normalizes along the feature dimension only, while  Batch Normalization operates across batches and can disrupt temporal dependencies, potentially reducing model performance.

\subsubsection{Temporal Features Extraction}
To address the varying frame rate of RFID data, we extract temporal features independently within each modality and then fuse them via concatenation. 

For the mmWave branch, after spatial feature extraction, we obtain a matrix of shape $(N^{M}_{\text{frame}},\, N^{M}_{\text{dim}})$, where $N^{M}_{\text{frame}}$ is the number of frames and $N^{M}_{\text{dim}}$ is the feature dimension for each frame. We add trigonometric positional encodings~\cite{vaswani2017attention} and process the sequence through three Transformer Encoder layers. To summarize temporal information, mean pooling is applied across the time dimension, resulting in a vector of size $N^{M}_{\text{dim}}$ for each sample.

For the RFID branch, we apply a similar temporal transformer to obtain a vector of dimension $N^{R}_{\text{dim}}$. The word embeddings of object names are projected using a linear layer into the same feature space, $N^{R}_{\text{dim}}$. We concatenate the temporal features and projected embeddings, and further align the vector dimension to $N^{M}_{\text{dim}}$ using a linear block, preparing the features for fusion with the mmWave branch.

\subsubsection{Fusion and Decoupling Output} After processing within each modality, we concatenate the features from mmWave and RFID modalities and feed them into two output branches: an action classifier and a target identifier. In this way, both branches utilize information from both modalities for decision-making, while disentangling the two outputs for a better generalizability and interpretability. Each output branch is composed of linear layers and produces either the action category or the target object, respectively. Let $loss_{\text{action}}$ and $loss_{\text{target}}$ denote the cross-entropy losses for the action classifier and target identifier. The total loss is defined as their weighted sum:
\begin{equation}
    loss_{\text{total}} = loss_{\text{action}} + \alpha \cdot loss_{\text{target}},
    \label{equ:loss}
\end{equation}
where $\alpha$ is a hyperparameter that balances the two losses. Finally, the model outputs the predicted action category and the tag ID of the target object among the input RFID matrix.

\subsection{Generating Synthetic Multimodal RF Data}\label{subsec:method_synth}
\begin{figure}[t]
  \centering
  \includegraphics[width=0.7 \linewidth]{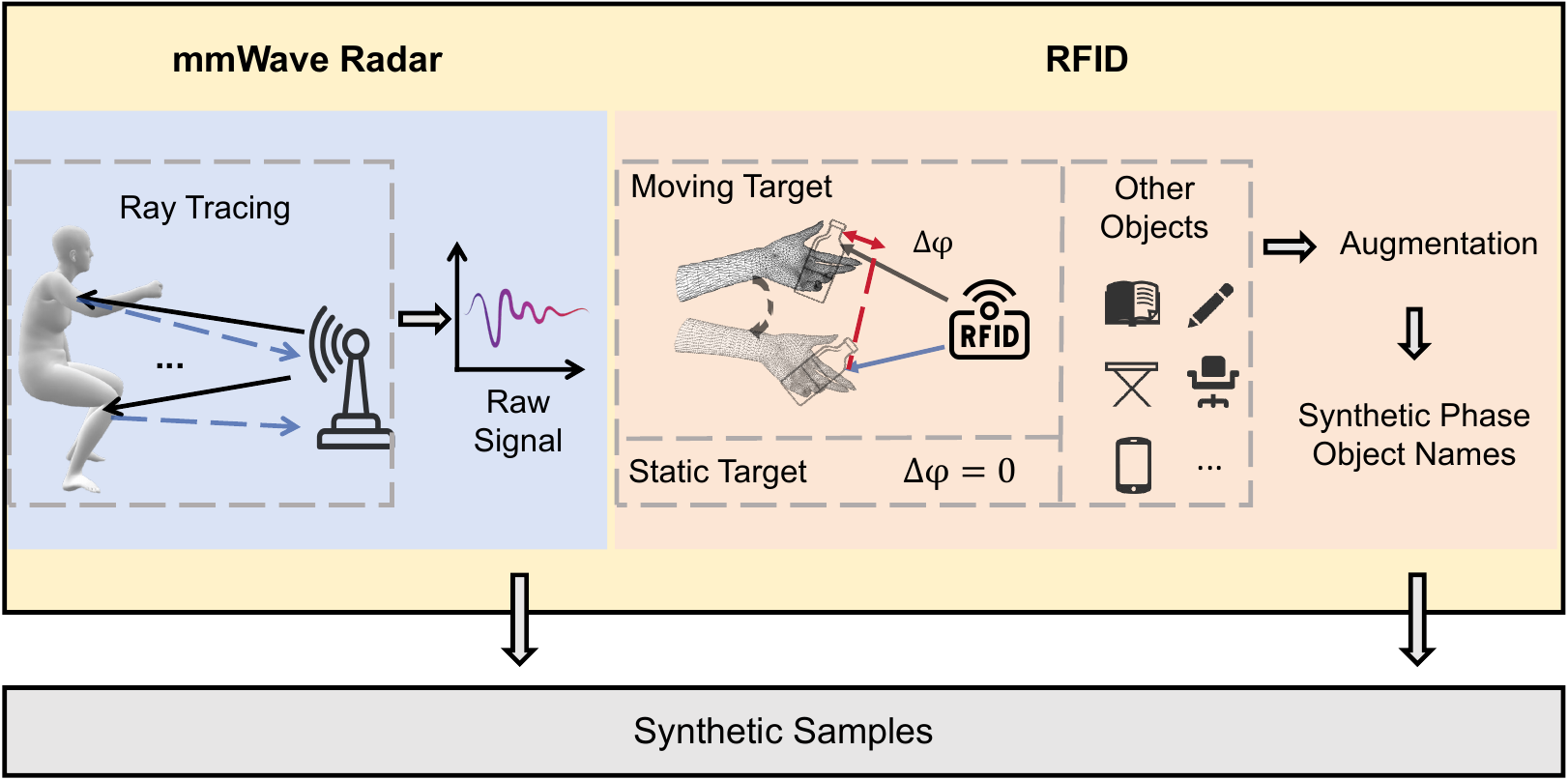}
  \caption{Overview of the RF signal simulator. Our simulator can synthesize both mmWave and RFID data given a mesh sample. \label{fig:simulator}}
 \Description{Structure of the modality fusion network }
\end{figure}
To boost model performance with limited real-world training data and enhance generalizability to unseen factors, we propose generating large amounts of synthetic data using a custom-built simulator. As illustrated in Fig.~\ref{fig:simulator}, the proposed simulator supports multimodal RF data synthesis, \ie both mmWave and RFID data. In this section, we detail the design of our simulator.
\subsubsection{Determine Principles of Simulation}
Existing RF signal simulators for wireless sensing are typically based on either data-driven methods~\cite{ahuja2021vid2doppler,deng2023midas,chi2024rf} or physics-based methods~\cite{korany2019xmodal,deng2023midas++,rfgen,cao2024mmclip,li2024sbrf}. Although data-driven simulators are versatile, their models are usually trained on non-HOI datasets and lack interpretability. As a result, synthesizing HOI data using data-driven approaches may lead to inconsistent representations between modalities, such as mismatched object movements and user actions. To ensure both explainability and reproducibility, we adopt a physics-based approach to build our RF signal simulator.

\subsubsection{Prepare Input to the Simulator} \label{sec:inputsim}
We choose ray tracing to simulate mmWave signals, as it is the most widely used technique among physics-based methods. However, ray tracing requires well-annotated 3D mesh data as input, which is particularly challenging for HOI tasks. Unlike HAR tasks—where generative models can produce human mesh data—generating meshes for HOI is more difficult due to the added complexity from diverse objects and their geometries~\cite{diller2024cg, xu2025interact,peng2025hoi}. This complexity can cause mismatches between human motion and real object movements, making datasets from motion capture systems preferable for realism.

Fortunately, several HOI datasets with human mesh have been released in the computer vision community~\cite{jiang2023full, jiang2024scaling, jiang2024autonomous}. Since our simulator can accept input from any HOI dataset that provides 3D mesh data and HOI category annotations, it is compatible with datasets either from motion capture or generative models in the future. Among the options, we select the TRUMANS dataset~\cite{jiang2024scaling} as the most suitable, because it is fully public and richly annotated with action names, target object names, and details about each interaction—whereas other datasets generally lack object information.

\subsubsection{Build the Simulator}
After preparing the mesh dataset, we develop a physics-based simulator for \sysname{} that jointly synthesizes mmWave and RFID data. Specifically, our mmWave signal synthesis module builds upon RFGen~\cite{rfgen}, an open-source ray tracing framework originally designed for HAR. Instead of using generative models for human mesh, we adapt this framework to accept 3D mesh data from public HOI mesh datasets. {\edited Although the heatmap representations used in our system contain reflection intensity information, we do not aim to precisely model absolute radar cross section (RCS) ~\cite{richards2005fundamentals} in simulation. Instead, all heatmaps are normalized to the [0,1] range using min–max normalization, which substantially reduces the influence of absolute intensity values that depend on distance, angle, and material properties. This design choice encourages the model to focus on relative spatiotemporal patterns and motion-induced structures that are more consistent between simulated and real-world data, rather than relying on absolute reflection strength. We note that spectrum-based representations have been widely adopted in prior radar sensing work~\cite{joshi2024towards,cao2024mmclip,yan2025mmexpert}, and these studies have also demonstrated that simulated radar data can effectively enhance downstream performance when combined with appropriate domain adaptation strategies.}

However, \sysname{} requires not only mmWave data but also RFID data, which is not supported by any existing tools. Moreover, ray tracing cannot be directly applied for RFID simulation due to two reasons: (1) Ray tracing introduces large errors with RFID frequencies due to geometric approximations~\cite{talbi2001simulation}, as RFID operates at much lower frequencies than mmWave; and (2) Accurate object coordinates are needed to correctly position meshes for ray tracing, yet only a small portion of the TRUMANS dataset includes such annotations. This constraint would significantly reduce the number of usable samples for ray tracing if applied to RFID.

\begin{figure*}
\centering
\subfigure[x-axis]{
\includegraphics[width=0.305\linewidth]{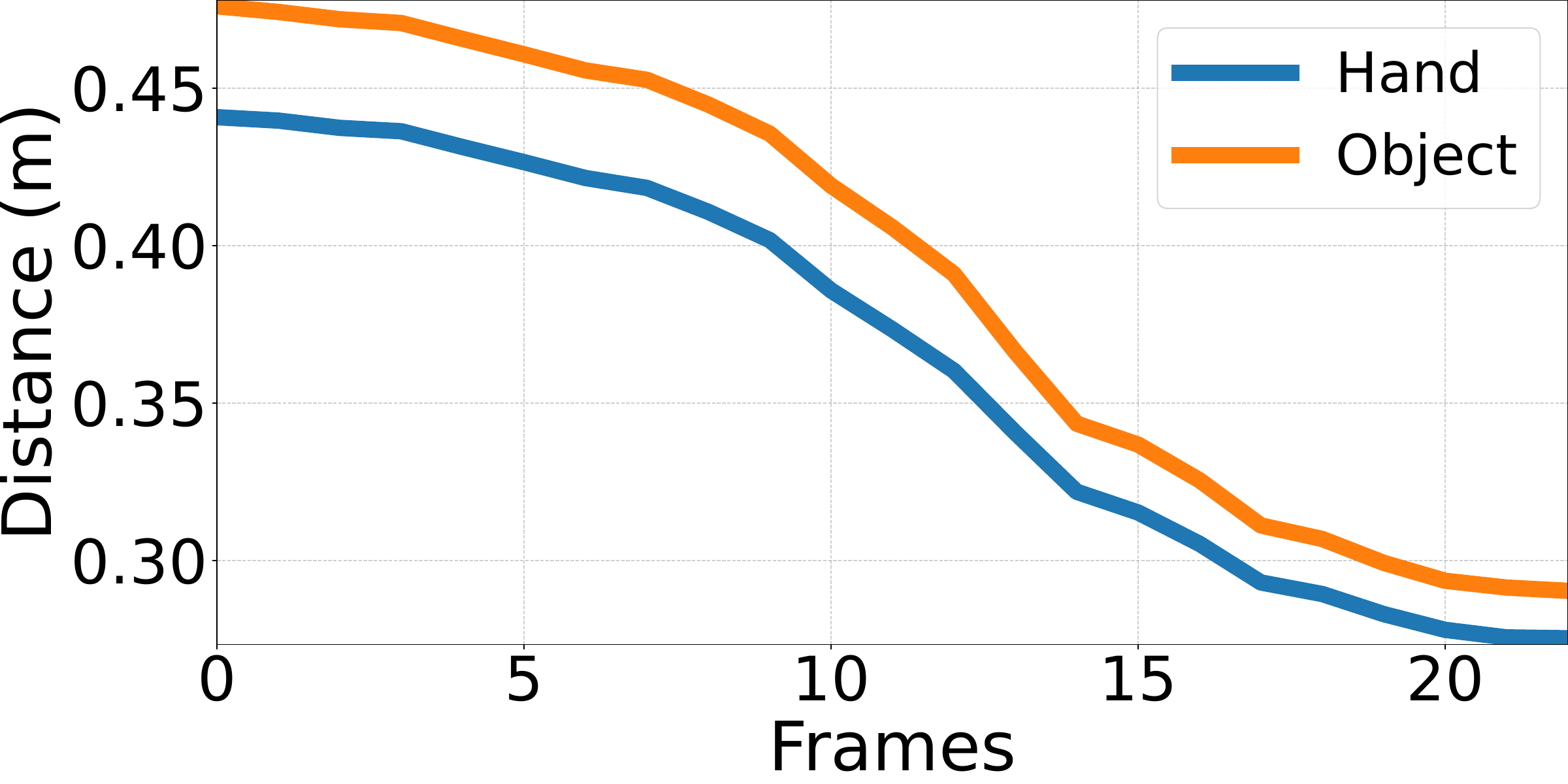 
}
}
\hfill
\subfigure[y-axis]{
\includegraphics[width=0.305\linewidth]{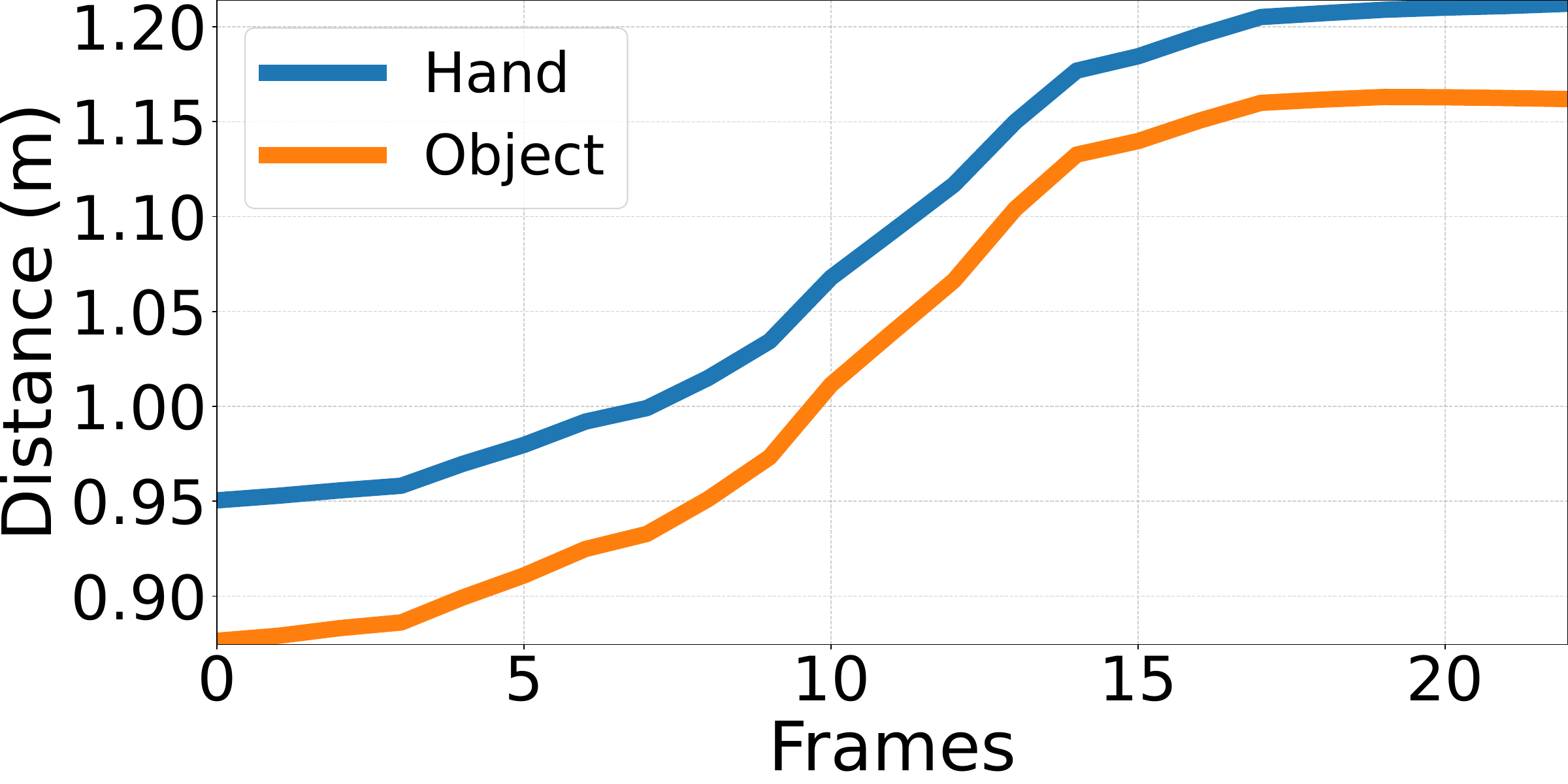 
}
}
\hfill
\subfigure[z-axis]{
\includegraphics[width=0.305\linewidth]{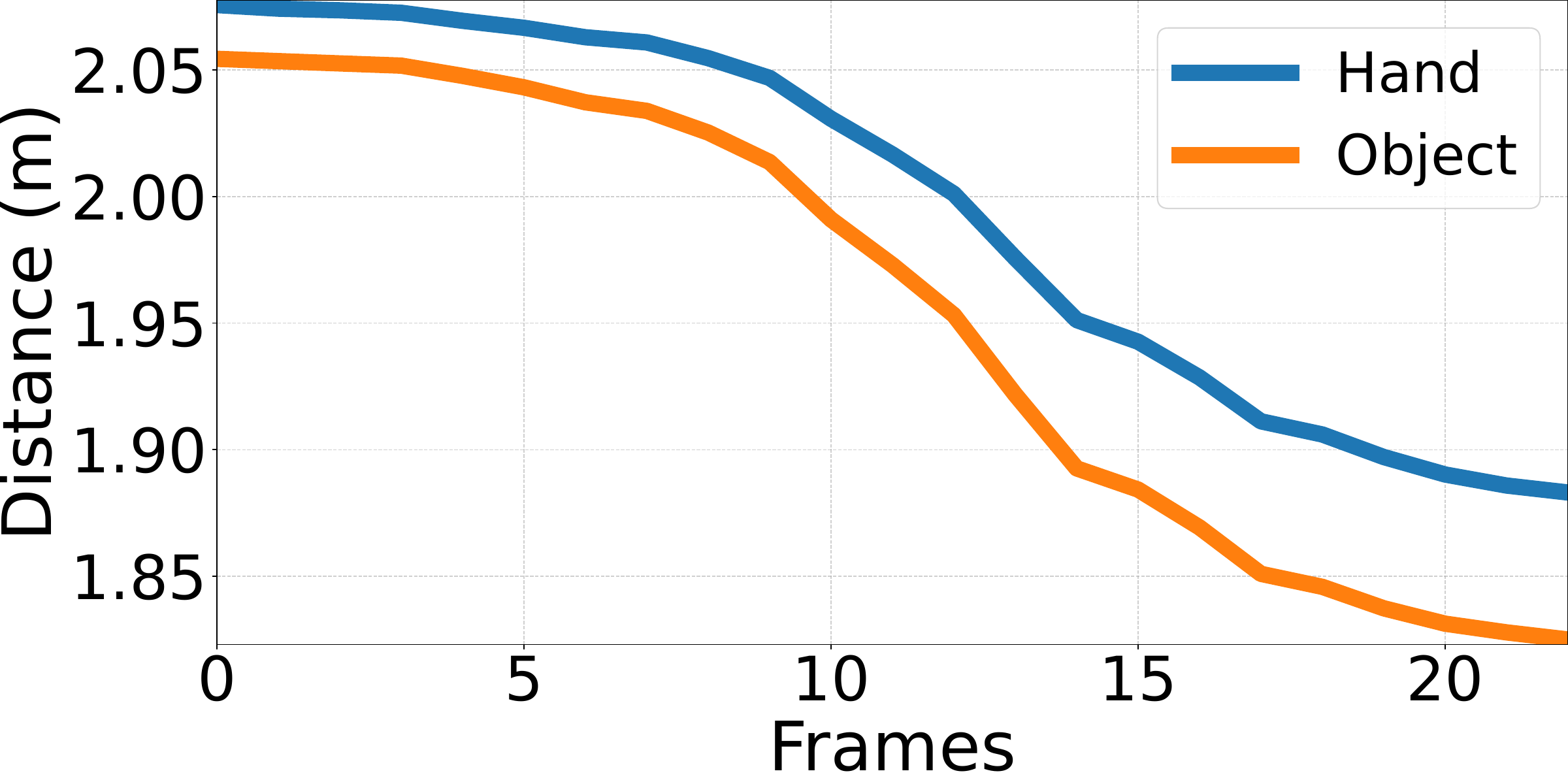 
}
}
\caption{The coordinates in three axes of the object and the hand in a `pick up' action. \label{fig:HandObjLoc}}
\Description{The trajectories of the object and the hand in a `pick up' action.}
\end{figure*}
\begin{figure}
\centering
\subfigure[Real $\Delta \varphi$]{
\includegraphics[width=0.4\linewidth]{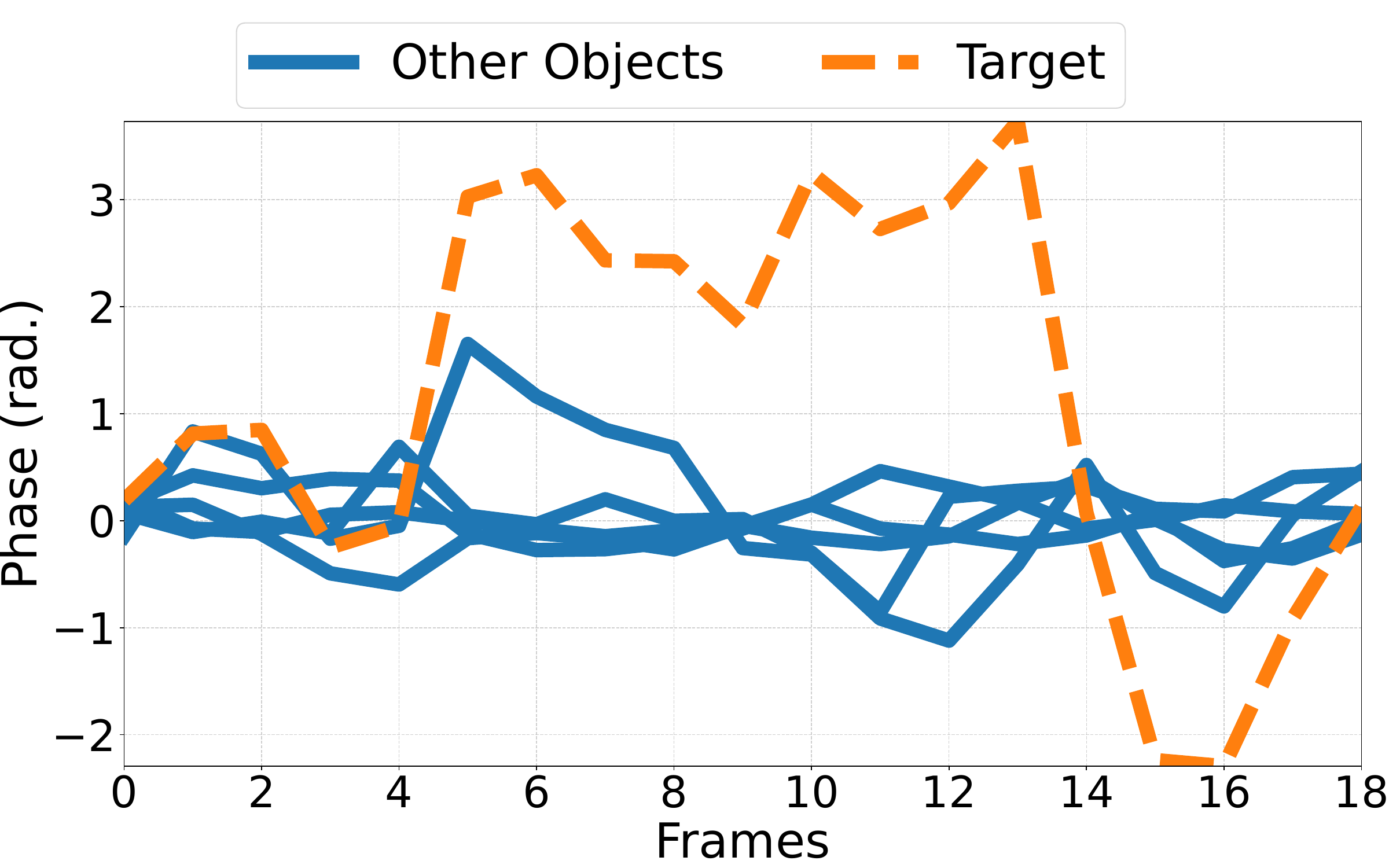}
}
\hspace{0.03\linewidth}
\subfigure[Synthetic $\Delta\varphi$]{
\includegraphics[width=0.4\linewidth]{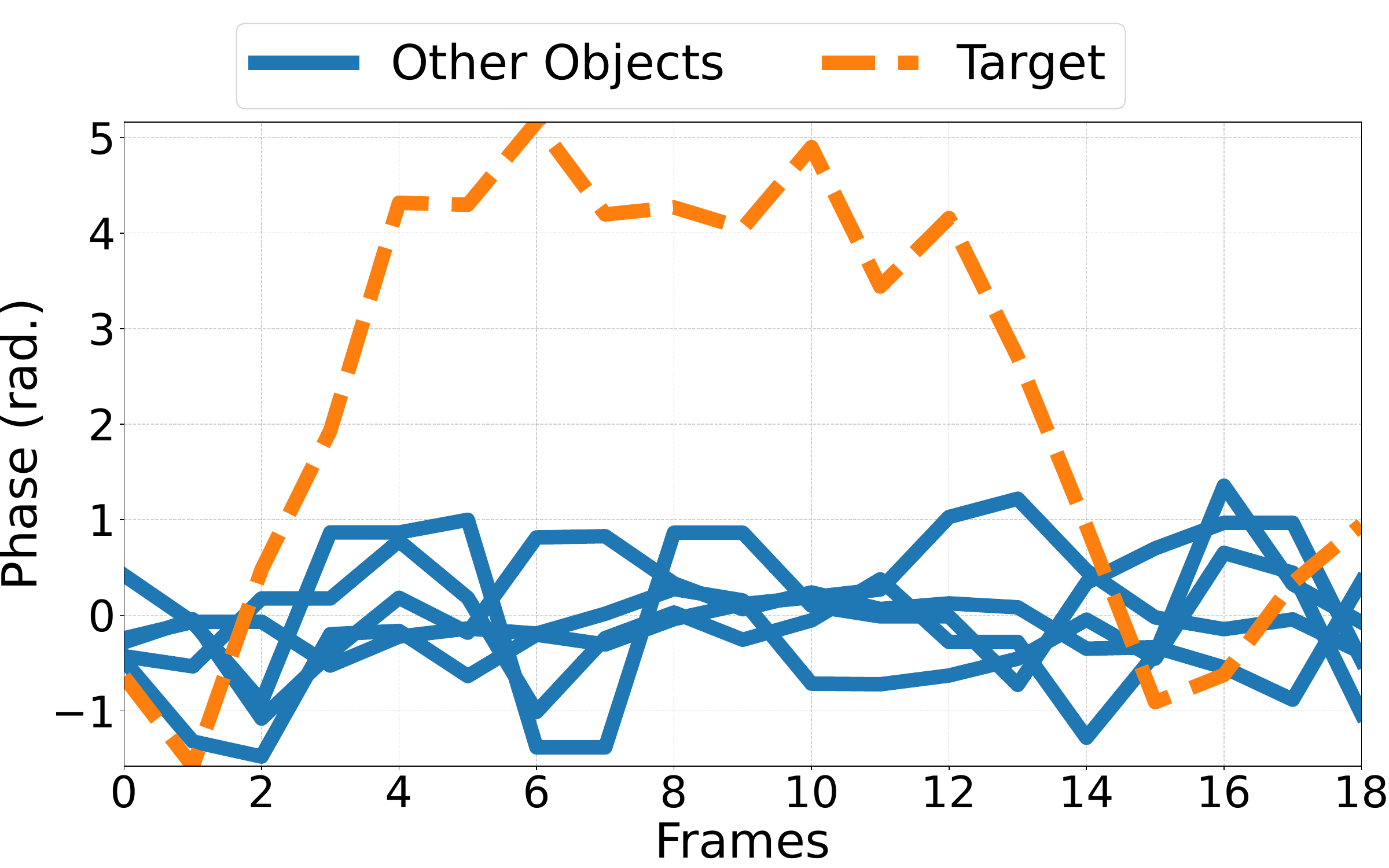} 
}
\caption{Comparing normalized tag phase difference in a `pick up' action. \label{fig:rfiddeltaphase}}
\Description{Photos and RFID signal phases of the tag attached to the target object in two different interactions.}
\end{figure}

To address the challenges above, we design the RFID data synthesis module to directly simulate RFID signal phase differences based on physical principles, focusing on object movement. We use phase differences ($\Delta \varphi$) to capture the kinematic features relevant to our model. In this way, we do not require explicit object position annotations, since $\Delta \varphi (t)$ depends solely on the distance change between adjacent frames. For static objects, we set $\Delta \varphi = 0$. For moving objects, the TRUMANS dataset provides information on which hand the user uses during interaction, and the SMPL-X human body mesh~\cite{pavlakos2019expressive} supplies precise hand locations. As shown in Fig.~\ref{fig:HandObjLoc}, the hand and the target object's trajectories are highly similar during interaction. Thus, we use the corresponding hand location as a proxy for the object's location to calculate $\Delta \varphi$. Additionally, for non-target candidate objects, we assign them random object names from TRUMANS and set $\Delta \varphi=0$ for their phases.

To bridge the gap between synthetic and real-world data, we apply three data augmentation steps: (1) Simulate packet loss by randomly dropping samples, emulating EPC Gen2 RFID standard collisions~\cite{xie2019implementation}; (2) Add Gaussian white noise to the synthesized phase signals; and (3) Normalize the data to reduce the impact of outliers. Fig.~\ref{fig:rfiddeltaphase} presents a comparison of our synthesized $\Delta\varphi$ during a pick-up action with real-world measurements (using one antenna and six objects). In both cases, the target's $\Delta\varphi$ is clearly distinguishable from those of other candidates.

{\edited
\begin{algorithm}[t]
\caption{Synthetic object category expansion}
\label{alg:object_expansion}
\KwIn{
Original object categories $\mathcal{O}_{old}$; new object categories $\mathcal{O}_{new}$; 
object-agnostic action list $\mathcal{A}$; original synthetic dataset $\mathcal{D}$
}
\KwOut{Expanded synthetic dataset $\mathcal{D}^{*}$}

$\mathcal{O}_{all} \leftarrow \mathcal{O}_{old} \cup \mathcal{O}_{new}$\;
$\mathcal{D}^{*} \leftarrow \mathcal{D}$\;

\ForEach{sample $(a, o) \in \mathcal{D}^{*}$}{
    \If{$a \in \mathcal{A}$}{
        Randomly sample a new object category $o' \sim \mathcal{O}_{all}$\;
        Replace the original object category $o$ with $o'$\;
        Update the corresponding object name embedding using $o'$\;
    }
}
\Return{$\mathcal{D}^{*}$}
\end{algorithm}
\subsection{Generalization to New Object Categories}\label{subsec:method_newobj}
In HOI recognition, the space of object categories is typically much more diverse than that of action categories, as similar interaction dynamics (e.g., \textit{pick up} or \textit{put down}) can involve a wide range of objects. This makes generalization to unseen object categories particularly challenging.

\subsubsection{Challenges.} The difficulty mainly arises from two aspects. First, in conventional formulations that treat each HOI as an independent class, the model is required to recognize previously unseen action--object pairs, which correspond to unseen categories at test time. Second, wireless signals are highly affected by environmental factors and object properties, making it non-trivial to disentangle motion-related patterns from object-specific variations without explicit modeling. To address these challenges, we design \sysname{} to support generalization to new object categories from both the model and simulation perspectives.

\subsubsection{Model Perspective.}
An intuitive approach to HOI recognition is to treat each HOI category as an independent category and train a single-branch classifier over all combinations. However, such a formulation cannot generalize to unseen object categories, as each new action--object pair corresponds to an unseen category. 
In contrast, our model adopts a decoupled output design that separates action recognition from object identification. This design is motivated by the observation that many HOI categories share common interaction dynamics across different categories of objects. The object branch predicts the index of the target object within the candidate set for each sample, rather than assigning a fixed global category label.

In addition, the RFID branch uses an input representation consisting of phase-based temporal features together with object identity embeddings. For interaction primitives such as \textit{pick up} and \textit{put down}, the motion-induced phase variations are primarily determined by interaction dynamics rather than specific object instances. As a result, objects with similar interaction patterns can share transferable representations.

This enables action representations to generalize across objects, while object identity is inferred separately from the RFID branch.

\subsubsection{Simulation Perspective.}
To further support generalization beyond the limited real-world dataset, we extend object categories at the simulation level. Based on the observation that many interaction primitives are largely object-agnostic in terms of motion dynamics, the simulation pipeline enables systematic reassignment or expansion of object categories associated with the same interaction primitives. In practice, each action can be paired with either existing or newly introduced object categories during RF data synthesis, without requiring additional motion capture or mesh reconstruction.

\subsubsection{Putting Everything Together.}
To extend \sysname{} to unseen object categories, we first expand the synthetic dataset, as summarized in Algorithm~\ref{alg:object_expansion}. Specifically, we generate synthetic samples using our simulator and update the object categories with the expanded object set. The resulting synthetic dataset is used to pre-train the model, followed by fine-tuning on the same real-world dataset containing only the original object categories. The model is then evaluated on real-world samples with unseen object categories.
}

\section{System Implementation}

\begin{figure}
\centering
\subfigure[Hardware: mmWave radar, RFID antenna and camera \label{fig:hardware}]{
\includegraphics[width=0.4\linewidth]{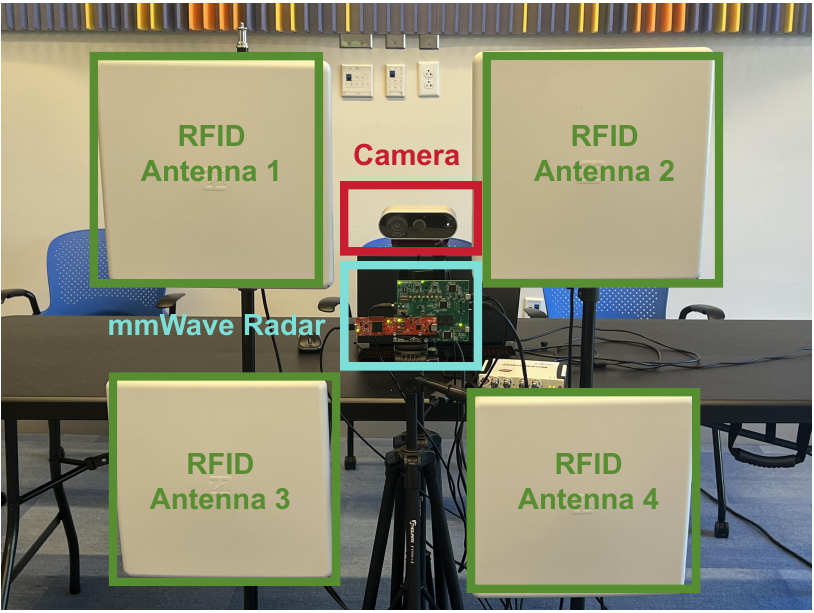} 
}
\subfigure[Scene A: Working space  (few multipath) \label{fig:scenea}]{
\includegraphics[width=0.4\linewidth]{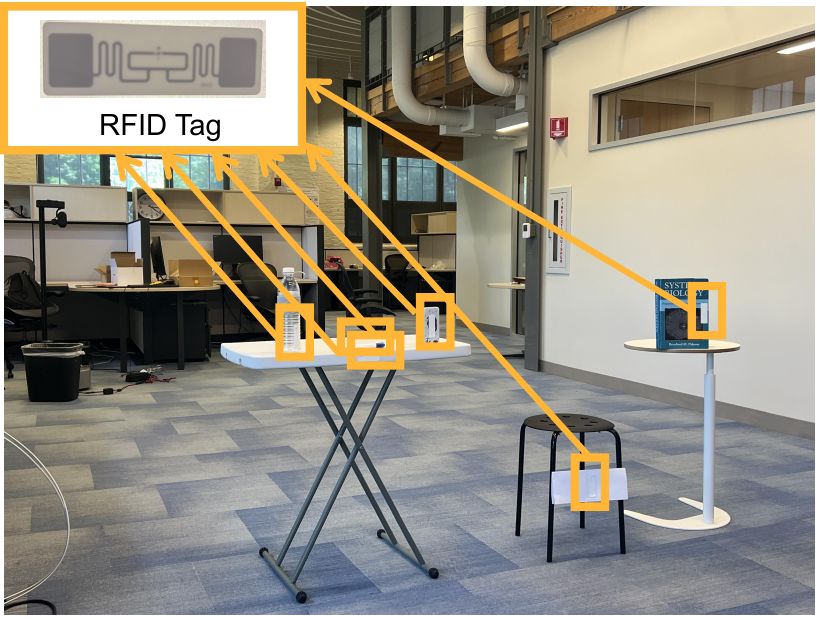} 
}
\subfigure[Scene B: Storage room (medium multipath) \label{fig:sceneb}]{
\includegraphics[width=0.4\linewidth]{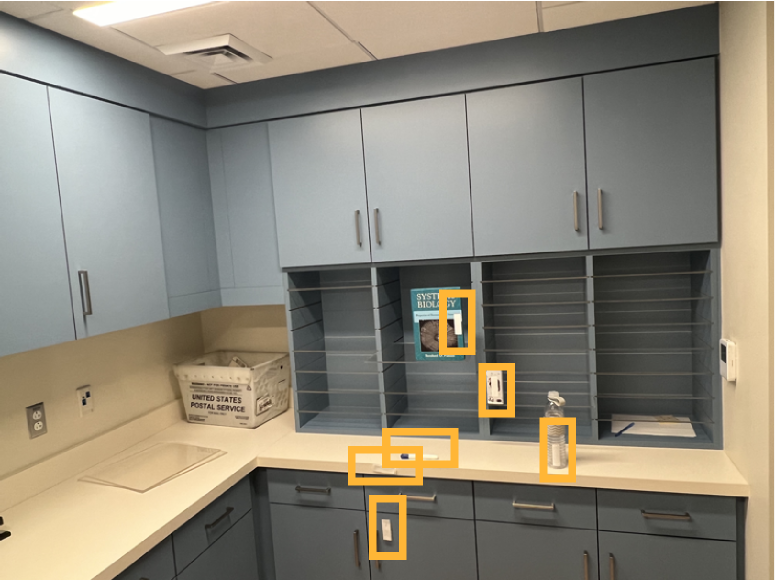} 
}
\subfigure[Scene C: Kitchen (rich multipath) \label{fig:scenec}]{
\includegraphics[width=0.4\linewidth]{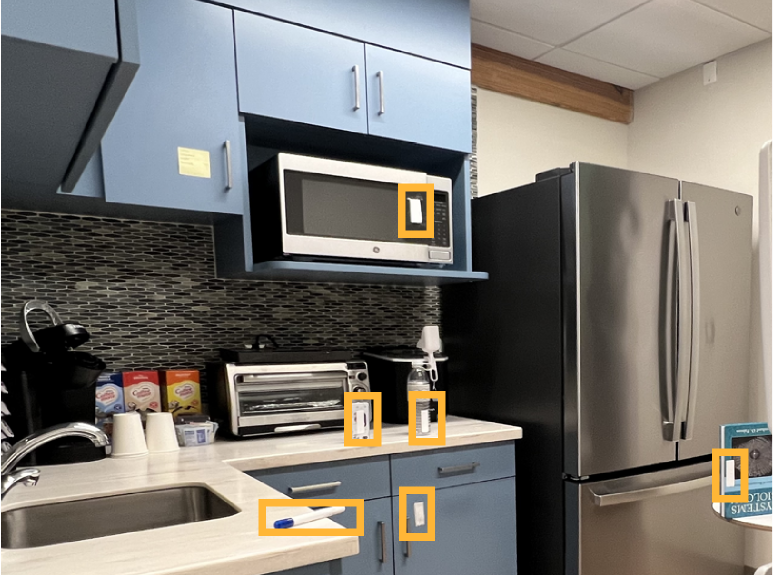} 
}
\caption{Hardware platform and experiment setups. \sysname{} is evaluated in three different scenes with various multipath. \label{fig:setups}}
\Description{Experiment setup.}
\end{figure}

\subsection{Data Collection Platform}
\subsubsection{Hardware} 
We build the \sysname{} hardware platform using commercial components to collect real-world data. As shown in Fig.~\ref{fig:hardware} and Fig.~\ref{fig:scenea}, our setup includes a TI AWR1843AOP mmWave radar~\cite{TIradar}, an Impinj R420 RFID reader~\cite{RFIDreader} with four antennas~\cite{RFIDant}, passive UHF RFID tags~\cite{RFIDtags}, and a Femto Bolt RGB camera~\cite{RGBcamera}. In Fig.~\ref{fig:setups}, we mark the attachment of RFID tags with yellow rectangles. The camera and the mmWave radar are placed closely at about one meter above the ground, surrounded by the RFID antennas.

The mmWave radar transmits Frequency Modulated Continuous Wave (FMCW) signals with a frame length of 0.1 seconds, operating as a 3TX 4RX MIMO system (emulating a $3 \times 4$ virtual array with $1/2\,\lambda$ antenna spacing). Each frame consists of 128 chirps (65~$\mu$s duration, 256 sampling points per chirp), using a 3.9~GHz bandwidth swept from 77 to 80.9~GHz. This configuration supports a maximum sensing range of 11~meters, range resolution of 4.3~cm, maximum detectable velocity of 4.5~m/s, and velocity resolution of 7.1~cm/s. As is shown in Fig.~\ref{fig:hardware}, the radar and RGB camera are mounted 1 meter above the ground, while the RFID antennas are placed around the radar with 45~cm spacing.

The RFID reader is FCC-compliant and performs frequency hopping between 902.75 and 927.25~MHz across 50 channels. This induces phase discontinuities, which we calibrate by profiling the initial phase readings as described in~\cite{wei2016gyro}.

\subsubsection{Software} 
We developed data collection software in Python to control all devices and collect multimodal data. The RFID data acquisition module uses public Python APIs~\cite{RFIDAPI,cameraAPI}. Since the mmWave radar lacks a Python API but streams data via Ethernet, we capture data in real-time by sniffing the Ethernet interface. We employ multi-processing to accelerate data handling and compress raw images to JPEG format for memory efficiency. With these solutions, we enable real-time data collection from all sensors using a Dell G15 laptop~\cite{laptop} with 16~GB of memory.

\subsection{Real-world Data Collection}

\begin{figure*}[t]
  \centering
\includegraphics[width=\linewidth]{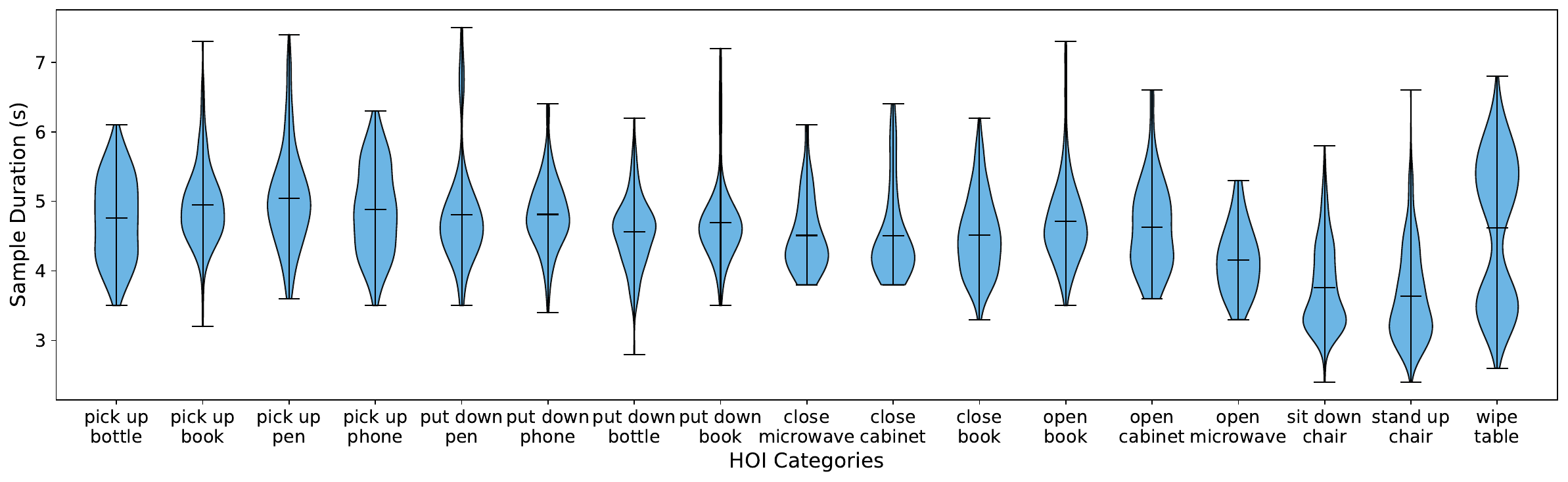}
  \caption{Real-world sample durations of 17 HOI Categories. The proposed modality fusion network is designed to support samples with various durations. \label{fig:duration}}
  \Description{HOI Categories. }
\end{figure*}

As shown in Fig.~\ref{fig:setups},  to evaluate the robustness of \sysname{}, we collect real-world data across three distinct scenes: a working space, a storage room, and a kitchen. Depending on the density of non-target objects, the three environments exhibit different degrees of static multipath interference. The working space has few multipaths, the storage room has medium multipaths, and the kitchen has rich multipaths.

{\edited In total, the dataset contains 17 HOI categories, including seven actions and eight objects, selected based on the interaction taxonomy of the TRUMANS dataset~\cite{jiang2024scaling}. 
While TRUMANS includes additional HOI categories, we focus on those with sufficient samples to ensure reliable learning of interaction dynamics. 
Rather than exhaustively enumerating all interactions, we focus on fundamental action primitives (e.g., pick, put, open, and close) that capture core motion patterns and serve as building blocks for more complex behaviors. This design facilitates learning transferable representations and supports generalization to unseen HOI compositions and object categories. }
During data collection, five other candidate objects are included in addition to the target object. 
One instance of each object type is present at a time, each equipped with an RFID tag.
Fig.~\ref{fig:duration} shows the distribution of the duration of all samples in each category. We can see that in the real world, the duration distribution varies both within and across categories, which highlights the importance of supporting variable-length samples when we design the neural network.

We recruited five users (both male and female, heights ranging from 163~cm to 191~cm). For each sample, one user stands in front of the sensors and performs a single $<$action, object$>$ tuple.  The subject-to-sensor distance varies between 1~m and 3~m,  within a $\pm$45-degree range in front of the sensors.
Each action is collected in seven user orientations, ranging from $-90^\circ$ to $90^\circ$ (see Fig.~\ref{fig:factors}). Data from mmWave radar, RFID, and RGB camera are acquired for each sample.

In addition to HOI categories, each sample is annotated with its setup: a combination of environment, user identity, user-sensor distance, and user orientation. Data are collected across 78 unique setups, with each user repeating interactions 5 to 10 times in each setup. {\edited In total, the data collection process spans approximately 1.5 months and yields 3,615 real-world samples (617 GB).} \footnote{The data collection was approved by the IRB of the authors’ institution.} The data are randomly partitioned, with 30\% used for training and 70\% for testing. {\edited  This main dataset is used for overall evaluation of \sysname{}, including ablation studies and assessing robustness across diverse setups. 
In addition, we collect an extra set of 625 real-world samples dedicated exclusively to generalization evaluation, which are strictly excluded from any training or model selection process. 
This additional dataset is used to evaluate the model under more challenging conditions, including unseen object categories (five new objects leading to 10 unseen HOI categories), human-induced occlusions (from hands or body), and dynamic environments with multi-user interference.}

\subsection{Data Synthesis Detail}
We utilize the TRUMANS dataset~\cite{jiang2024scaling} to provide mesh data for our simulator. TRUMANS contains 15 hours of motion-captured data with fine-grained HOI annotations, featuring four female and three male participants and a variety of target objects—including both static items (e.g., chairs) and dynamic items (e.g., bottles).

Prior to data synthesis, we filter out samples that do not match the categories shown in Fig.~\ref{fig:duration}, and further remove around 20\% of samples that are either too short or too long. To enrich and expand the dataset, we systematically vary the distance between the human body mesh and the sensors from 1 to 3~m, and place sensors at different orientations relative to the human mesh for each sample. 

Using the \sysname{} simulator, we generate 7,518 synthetic samples comprising mmWave and RFID data with well-annotated labels. {\edited 
The simulator also supports flexible object category assignment for generating diverse HOI samples, which is later used to study generalization to unseen objects. 
}
By default, the synthetic dataset is randomly split in half for training and validation. 

\subsection{Model Implementation and Training}
We implement our proposed model (Sec.~\ref{sec:network}) using PyTorch~\cite{paszke2019pytorch}, setting $N_{map}=256$, $N_{embed}=300$, $N^{M}_{dim}=1024$, and $N^{R}_{dim}=128$.
Each Transformer Encoder Layer in the temporal transformer block has two heads, and we use LeakyReLU~\cite{maas2013rectifier} as the activation function. The hyperparameter $\alpha$ is set to $0.5$ to balance the two losses.

Different from some prior works~\cite{cao2024mmclip,wang2024xrf55,zhao2023cubelearn}, our model handles variable-length sequence inputs by leveraging transformer blocks for temporal feature extraction. Since samples are processed in batches during training, we pad all samples within a batch to the same length and provide both the padded sequences and corresponding padding masks to the model.

For pre-training, the learning rate is set to $1\times10^{-5}$. We save the model weights after each epoch and select the checkpoint with the best performance on the validation set for subsequent fine-tuning. During fine-tuning, we use a learning rate of $1\times10^{-6}$. In both stages, AdamW~\cite{loshchilov2017decoupled} serves as the optimizer. Unless otherwise specified, all model training and evaluation are conducted on a single NVIDIA A100 GPU.

\section{Experiments}
\subsection{Metrics} 
In our experiment, for each sample, the model must classify the action among seven categories and identify the target from six candidate objects. A sample's HOI recognition is considered correct only if both the predicted action and target object match the ground truth. We define the following metrics to evaluate performance:\\
(1) \textbf{HOI Accuracy} is the ratio of test samples for which both the action and target object are recognized correctly.\\
(2) \textbf{Action Accuracy} is the ratio of test samples for which the action classified correctly, independent of the target object identification.\\
(3) \textbf{Object Accuracy} is the ratio of test samples for which the target object is identified correctly,  independent of the action classification.

\subsection{Advantages of Modality Fusion}
\begin{figure*}
	\centering
		\subfigure[HOI accuracy \label{fig:exp_modal_hoi}]{
    \includegraphics[width=0.315\linewidth]{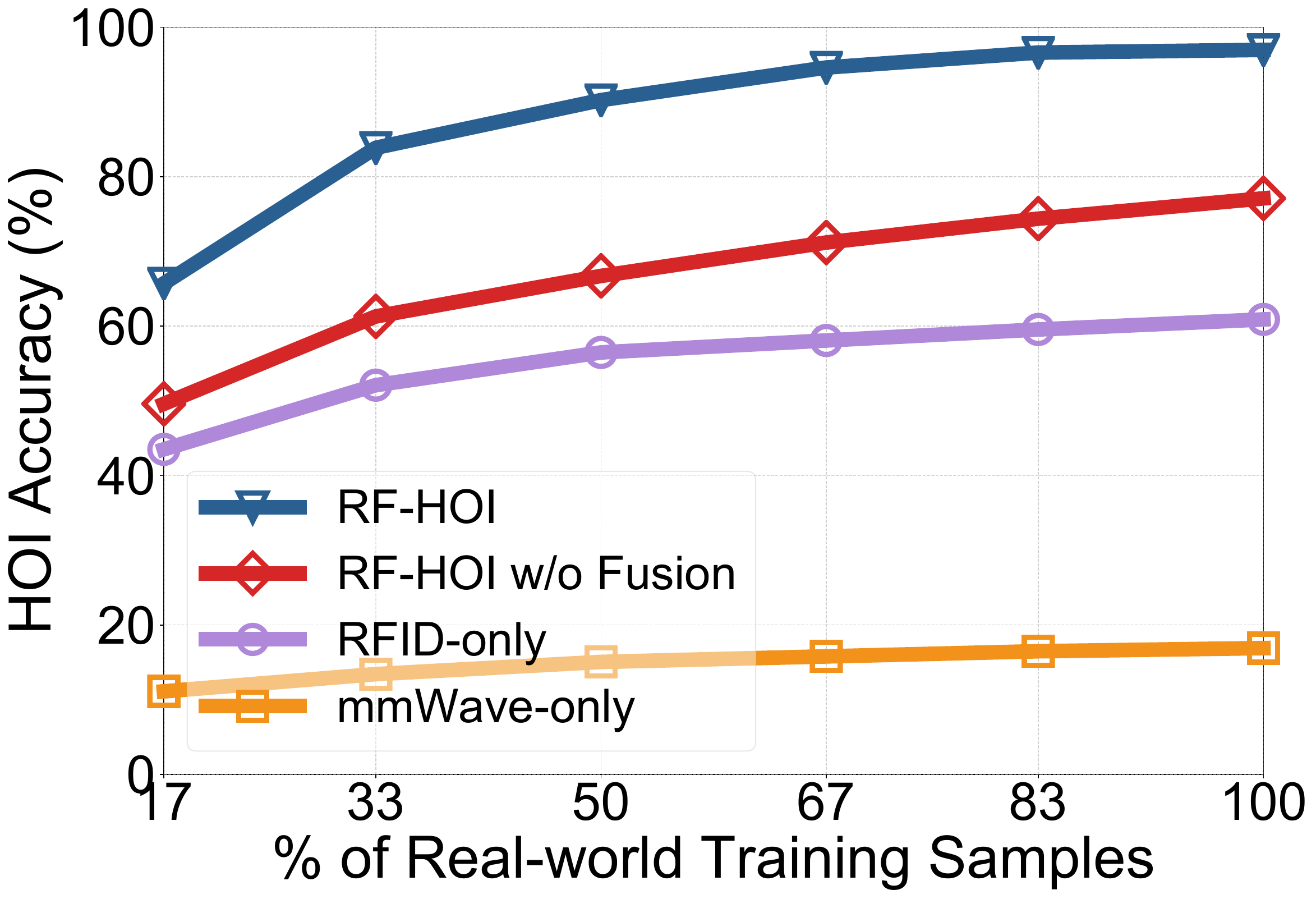} 
            }
            \subfigure[Action accuracy \label{fig:exp_modal_action}]{
            \includegraphics[width=0.315\linewidth]{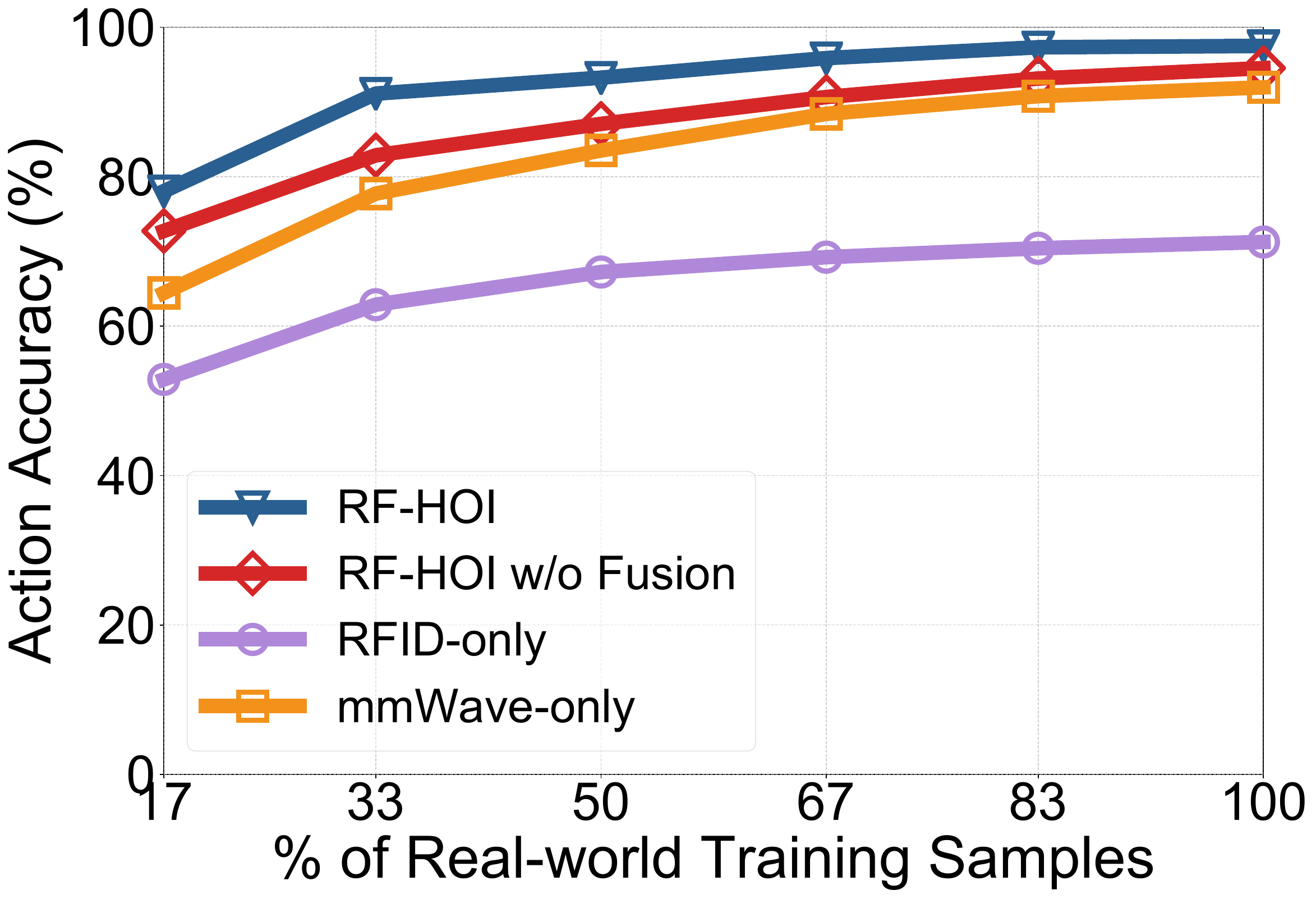} 
            }
            \subfigure[Object accuracy \label{fig:exp_modal_obj}]{
            \includegraphics[width=0.315\linewidth]{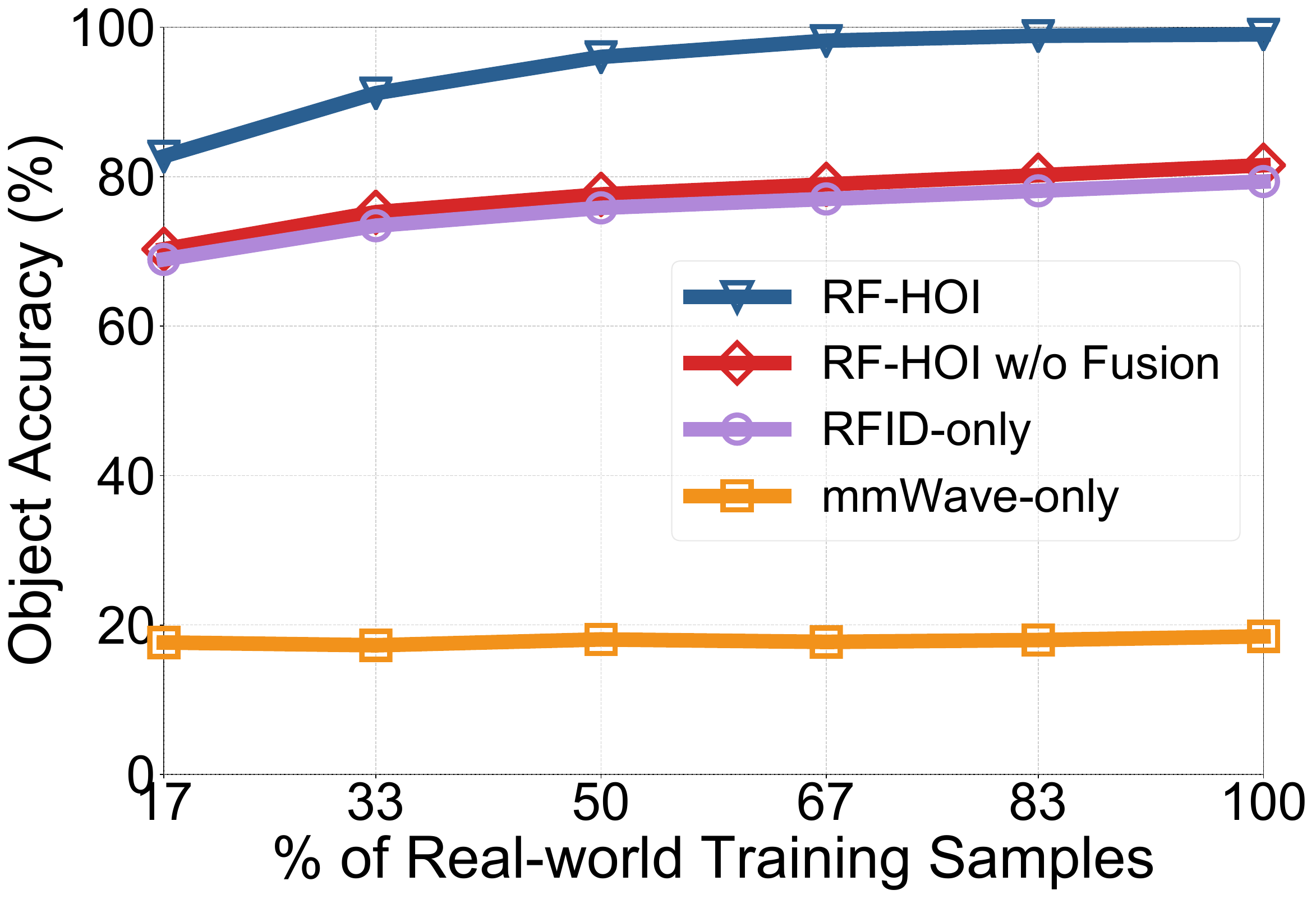}
            }
            \caption{Effectiveness of modality fusion. \label{fig:exp_modality}}

          \Description{Comparison between modalities. }
\end{figure*}
We first evaluate the effectiveness of our modality fusion network by comparing the proposed \sysname{} with the following methods:\\
(1) \textbf{\sysname{} w/o Fusion}: The concatenation of the two modality features is removed; the action classifier receives only mmWave features, while the target identifier receives only RFID features.\\
(2) \textbf{RFID-only}: Only RFID features are used for both the action classifier and target identifier.\\
(3) \textbf{mmWave-only}: Only mmWave features are used for both the action classifier and target identifier.

For each baseline, we first pre-train the model with the same amount of synthetic data as \sysname{}, then fine-tune the model on the same amount of real-world data. We vary the percentage of real-world training samples used in fine-tuning and repeat each experiment with five random seeds, reporting the average result to reduce variance due to randomness.

As is shown in Fig.~\ref{fig:exp_modal_hoi}, increasing the amount of real-world training data leads to performance improvements for all baselines, with \sysname{} consistently outperforming others. On average, the HOI accuracy of \sysname{} exceeds that of \sysname{} w/o Fusion, RFID-only, and mmWave-only by 21.29\%, 32.90\%, and 73.20\%, respectively. Specifically, when using all real-world training samples, \sysname{} reaches 96.97\% HOI accuracy, which is 19.87\%, 36.07\%, and 80.06\% higher than \sysname{} w/o Fusion, RFID-only, and mmWave-only, respectively.

To further elucidate the advantages of \sysname{}, we present the action accuracy in Fig.~\ref{fig:exp_modal_action} and the object accuracy in Fig.~\ref{fig:exp_modal_obj}. To be more specific, in terms of the action accuracy,  \sysname{} outperforms that of \sysname{} w/o Fusion, RFID-only, and mmWave-only by
5.35\%, 26.54\%, and 9.38\%, correspondingly. Similarly, with regard to object accuracy, \sysname{} outperforms that of \sysname{} w/o Fusion, RFID-only, and mmWave-only by 17.03\%, 18.90\%, and 76.47\%, correspondingly.
Across both metrics, \sysname{} consistently outperforms all other baselines, confirming the effectiveness of modality fusion for both action classification and target identification. Notably, using only mmWave yields high action accuracy (exceeding 90\%), but its object accuracy is close to random guessing among six candidate objects (approximately 17\%; see Fig.~\ref{fig:exp_modal_obj}). This limitation results in poor overall HOI accuracy. %
Conversely, using only RFID achieves higher object accuracy but significantly lower action accuracy than all other methods. Moreover, since features of interactive objects may reveal some action patterns and action patterns can help narrow the candidate targets, \sysname{} consistently achieves higher action accuracy than mmWave-only and higher Object accuracy than RFID-only.

Collectively, these results underscore the necessity of modality fusion and validate the effectiveness of the modal fusion design in \sysname{}.

\subsection{Close Performance Between \sysname{} and Vision}
Next, we evaluate the performance gap between \sysname{} and vision-based HOI recognition. Our vision baseline utilizes RGB videos as input, extracting image features with a pre-trained ResNet~\cite{he2016deep} and modeling temporal features using transformer blocks. Following InteractNet~\cite{gkioxari2018detecting}, the baseline includes two output branches for action and target object identification, respectively. Each experiment is repeated with five different random seeds to mitigate random variance.

\begin{figure}[t]
    \centering
        \begin{minipage}[t]{0.4\linewidth}
        \centering
        \includegraphics[width=\linewidth]{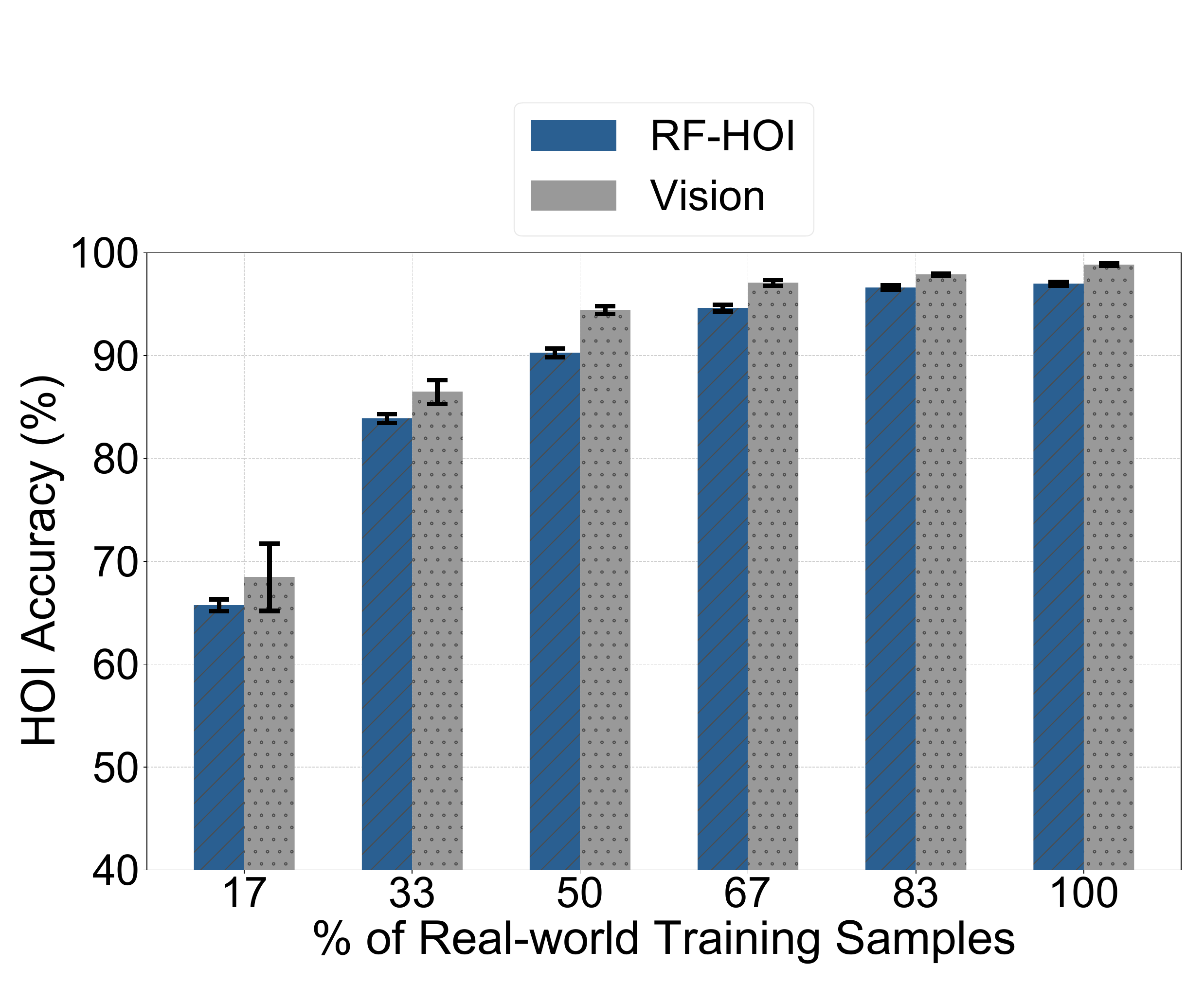}
        \caption{Compare \sysname{} with the vision model in terms of HOI accuracy. \label{fig:exp_vision}}
    \end{minipage}
        \hspace{0.03\linewidth}
    \begin{minipage}[t]{0.4\linewidth}
        \centering
        \includegraphics[width=\linewidth]{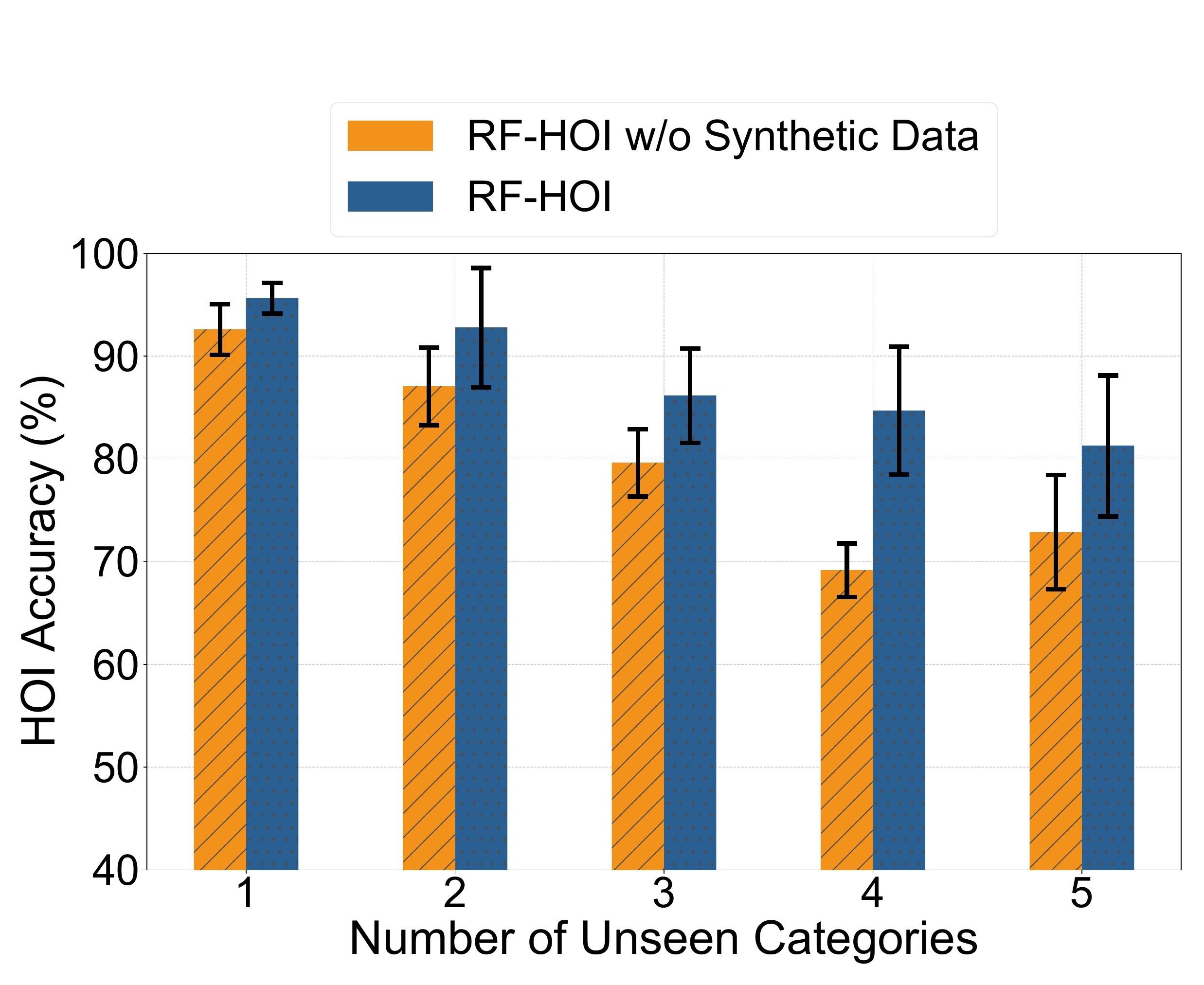}
\caption{\edited Generalizability to unseen HOI categories.  \label{fig:exp_ctg_test}}
          \Description{Test robustness on unseen HOI categories. }
    \end{minipage}
\end{figure}

Fig.~\ref{fig:exp_vision} compares the results of \sysname{} and the vision-based method across varying percentages of real-world training samples. While the video baseline performs slightly better than \sysname{}, the average gap is only 2.50\%. 
Furthermore, as the training set size increases, the gap narrows—reaching just 1.96\% when using 100\% of the training data. Nevertheless, \sysname{} maintains key advantages over the vision baseline by avoiding privacy concerns and remaining robust under poor ambient lighting conditions.

\subsection{Effectiveness of Synthetic Data}

As RF data collection for HOI recognition is expensive, synthetic data is important to enhance \sysname{}'s generalizability. In this section, we evaluate the effectiveness of the proposed simulator for multimodal RF data synthesis. 
For a fair comparison, we fine-tune \sysname{} on the same real-world training set used by the baseline, while the baseline trains the proposed \sysname{} model from scratch, without any synthetic data or pre-trained weights. 

\begin{figure}[t]
 \begin{minipage}[t]{0.4\linewidth}
        \centering
        \includegraphics[width=\linewidth]{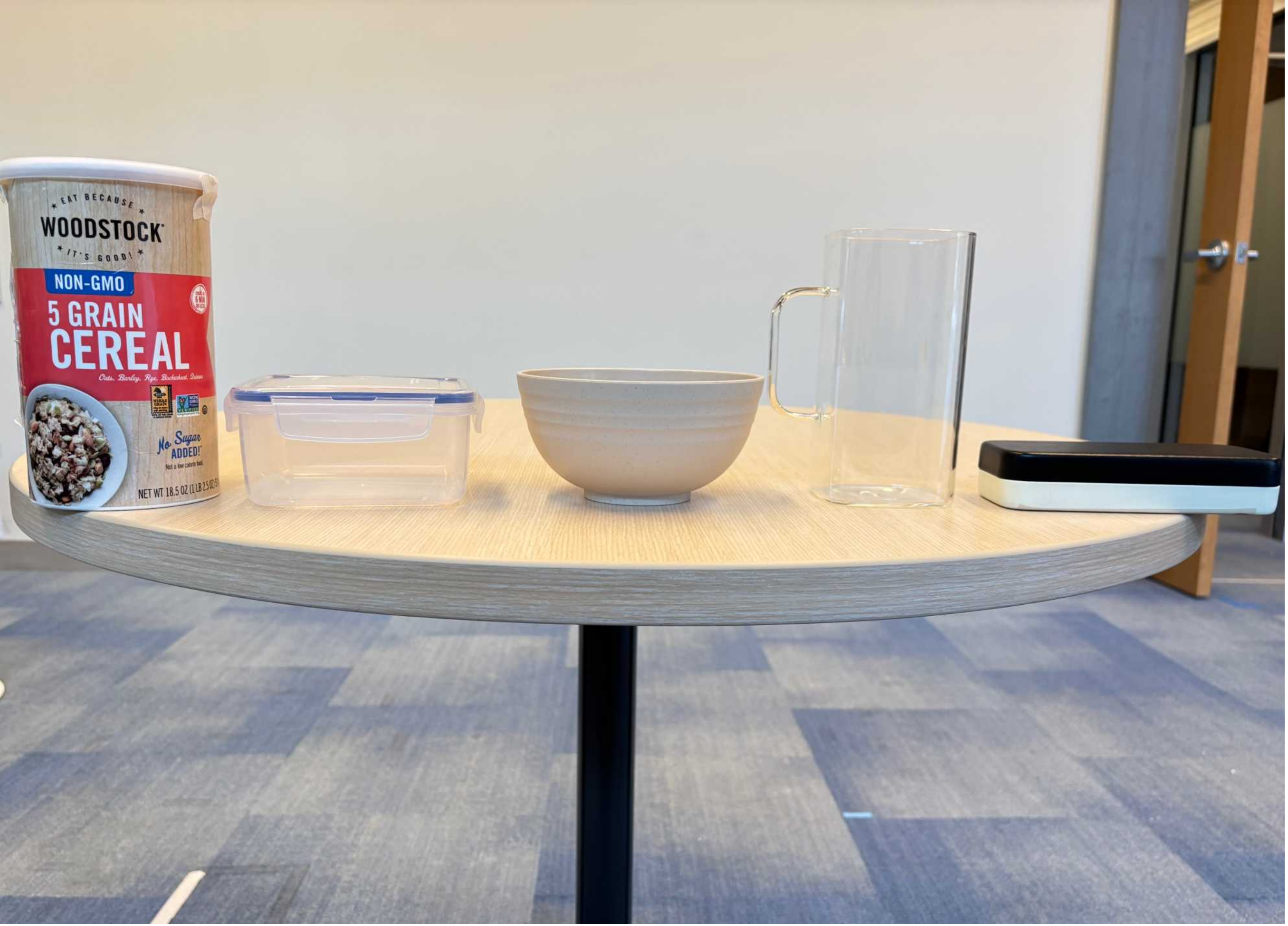}
\caption{\edited Five unseen objects: jar, lunchbox, bowl, cup, and eyeglass case. \label{fig:newobjphoto}}
          \Description{Test robustness on unseen objects. }
    \end{minipage}
            \hspace{0.03\linewidth}
 \begin{minipage}[t]{0.4\linewidth}
        \centering
        \includegraphics[width=\linewidth]{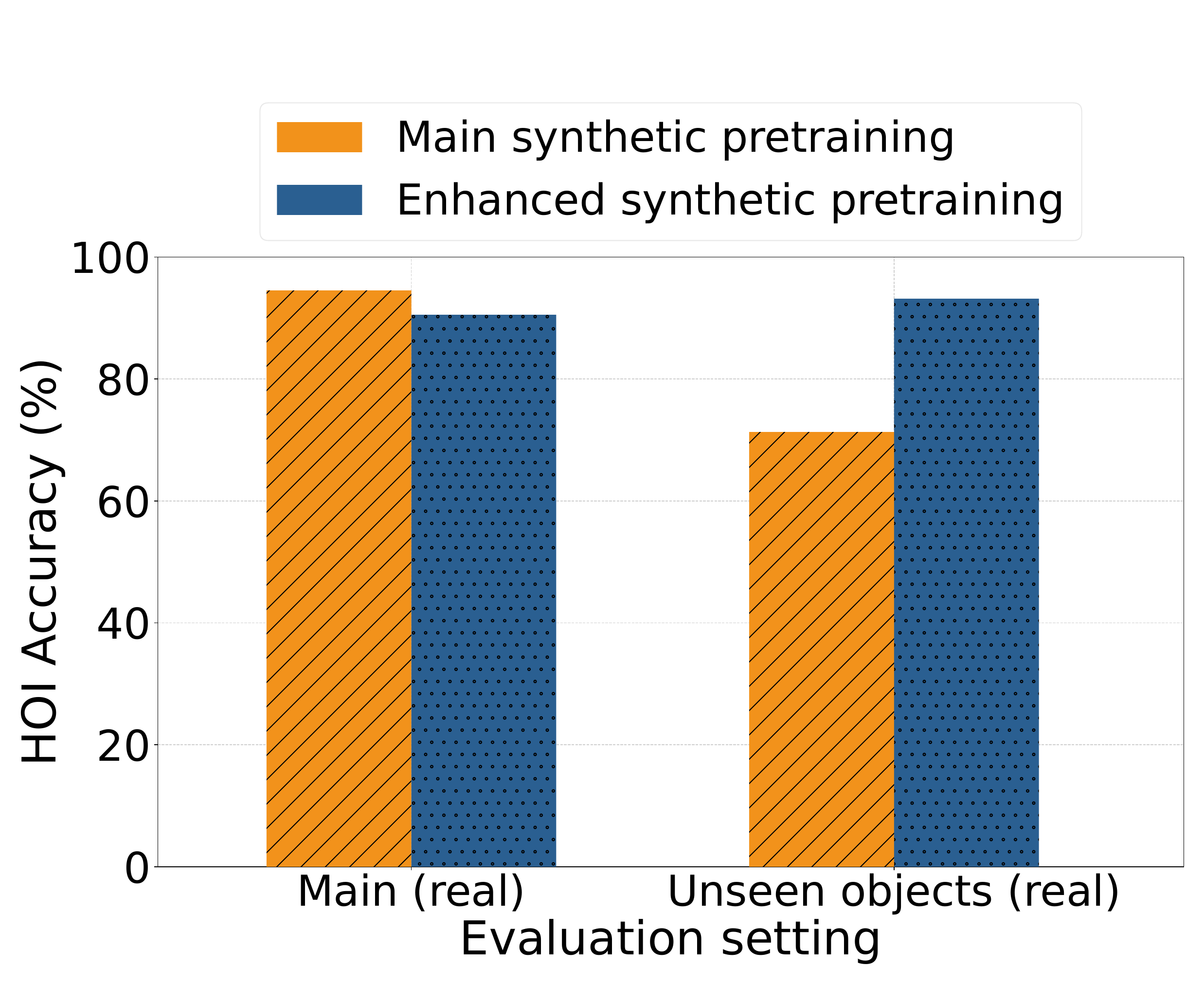}
\caption{\edited Generalizability to unseen object categories. \label{fig:exp_unseen_obj}}
          \Description{Test robustness on unseen objects. }
    \end{minipage}
\end{figure}
{\edited 
\subsubsection{Generalizability to Unseen HOI Categories}

First, benefiting from the decoupled output branches of actions and objects in our model design, we evaluate whether \sysname{} can generalize to unseen HOI categories, i.e., unseen action-object pairs, when the constituent actions and objects have been observed during training. This setting focuses on compositional generalization, which is important for scaling HOI recognition beyond a predefined set of categories.

To this end, we retain all HOI categories in the synthetic dataset, which provides broad coverage over action-object pairs. In contrast, for the real-world dataset, we randomly select varying numbers of HOI categories as unseen and reserve them exclusively for testing, while the remaining categories are used for training. Importantly, we ensure that all action categories and object categories remain present in the training data, so that only the action-object pairs are unseen.
For a fixed number of unseen HOI categories, we repeat the random selection five times to reduce sampling bias. Fig.~\ref{fig:exp_ctg_test} compares HOI accuracy between \sysname{} and the baseline. On average, \sysname{} outperforms the baseline by 7.84\% in HOI accuracy and maintains an accuracy of 81.25\% even when five HOI categories are completely unseen during training.

The results indicate that our design on decoupling action and object predictions enables effective compositional generalization, and that synthetic data further enhances this capability by expanding the coverage of the action-object pairs space.

\subsubsection{Generalizability to Unseen Object Categories}
Second, we evaluate generalization to unseen object categories. In this experiment, we focus on `pick up' and `put down' actions, where the target objects never appear in the real-world training data, and where the underlying motion patterns are largely consistent across different objects.

As shown in Fig. \ref{fig:newobjphoto}, we introduce five additional objects: jar, lunchbox, bowl, cup, and eyeglass case. This corresponds to 10 unseen HOI categories in total, formed by five unseen objects and two actions (`pick up' and `put down'). We collect 145 real-world samples for evaluation only, covering three orientations, and do not use these samples in any training or model selection process. This evaluation set is constructed to specifically assess generalization to unseen object categories. In addition, we generate an enhanced synthetic dataset of 7,518 samples with expanded HOI categories following our design in Sec.~\ref{subsec:method_newobj}, which provides sufficient coverage for pre-training with extended object diversity.
Models are pre-trained either on the original synthetic dataset or on the enhanced synthetic dataset, and are fine-tuned using the same original real-world training data.

The results are shown in Fig.~\ref{fig:exp_unseen_obj}. Here, ``Main (real)'' and ``Unseen objects (real)'' denote evaluation on the original real-world dataset and the newly collected unseen-object dataset, respectively, while the two bars correspond to models pre-trained on the original and enhanced synthetic datasets.
When pre-trained on the original synthetic dataset, the model achieves high HOI accuracy on seen objects (94.5\%) but suffers a substantial drop to 71.3\% on unseen objects. In contrast, pre-training on the enhanced synthetic dataset significantly improves accuracy on unseen objects to 93.1\%, while maintaining comparable performance on seen objects (90.5\%).  The slight decrease on seen objects reflects a natural trade-off when expanding object coverage during pre-training, rather than a degradation of the model’s recognition capability.

These results demonstrate that \sysname{} can effectively generalize to new object categories without additional real-world training data, provided that the underlying action patterns are similar and that object-level expansion is appropriately incorporated at the simulation stage.

}
\begin{figure*}
\centering
\subfigure[HOI accuracy]{
\includegraphics[width=0.315\linewidth]{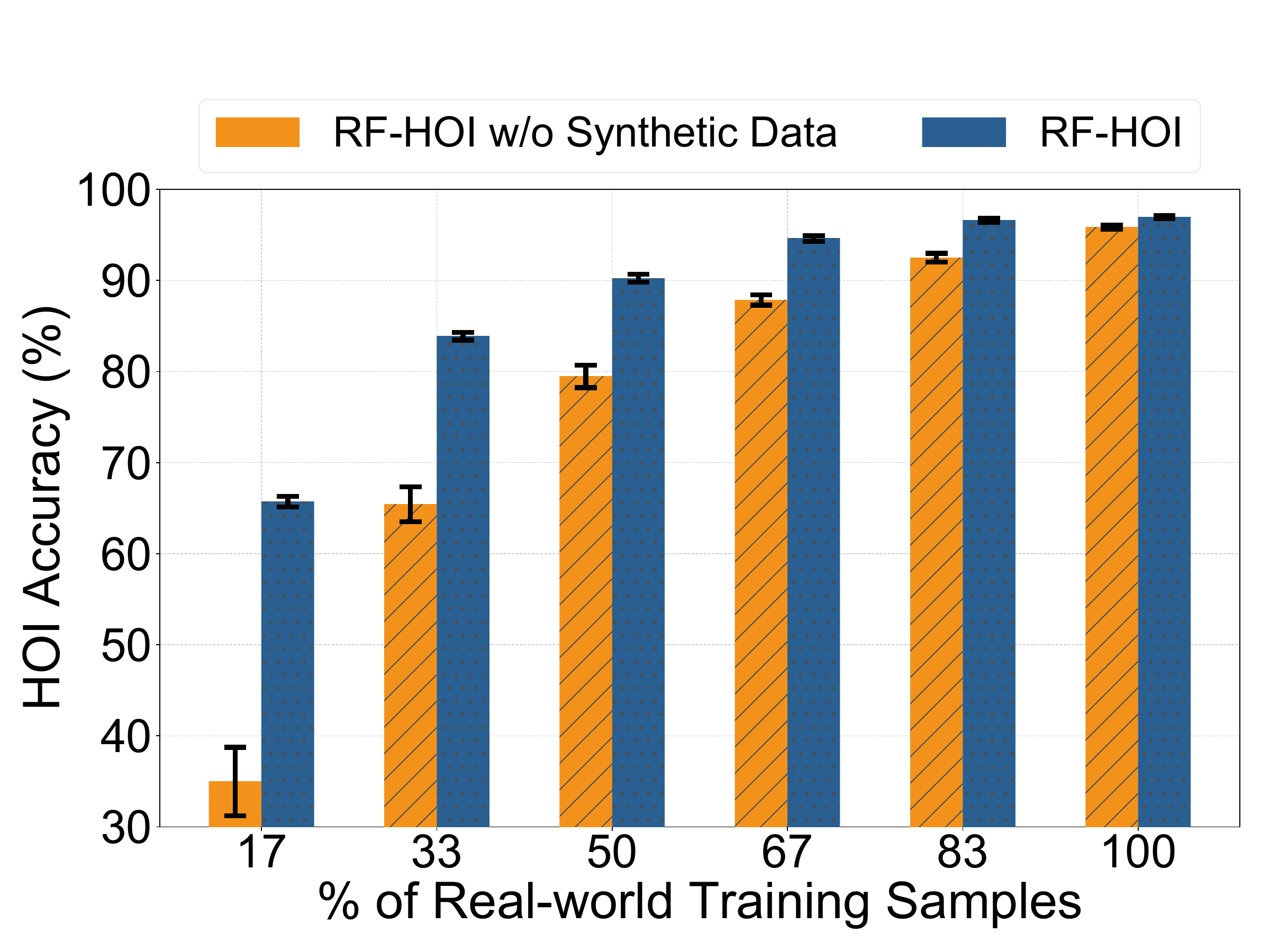} 
}
\subfigure[Action accuracy]{
\includegraphics[width=0.315\linewidth]{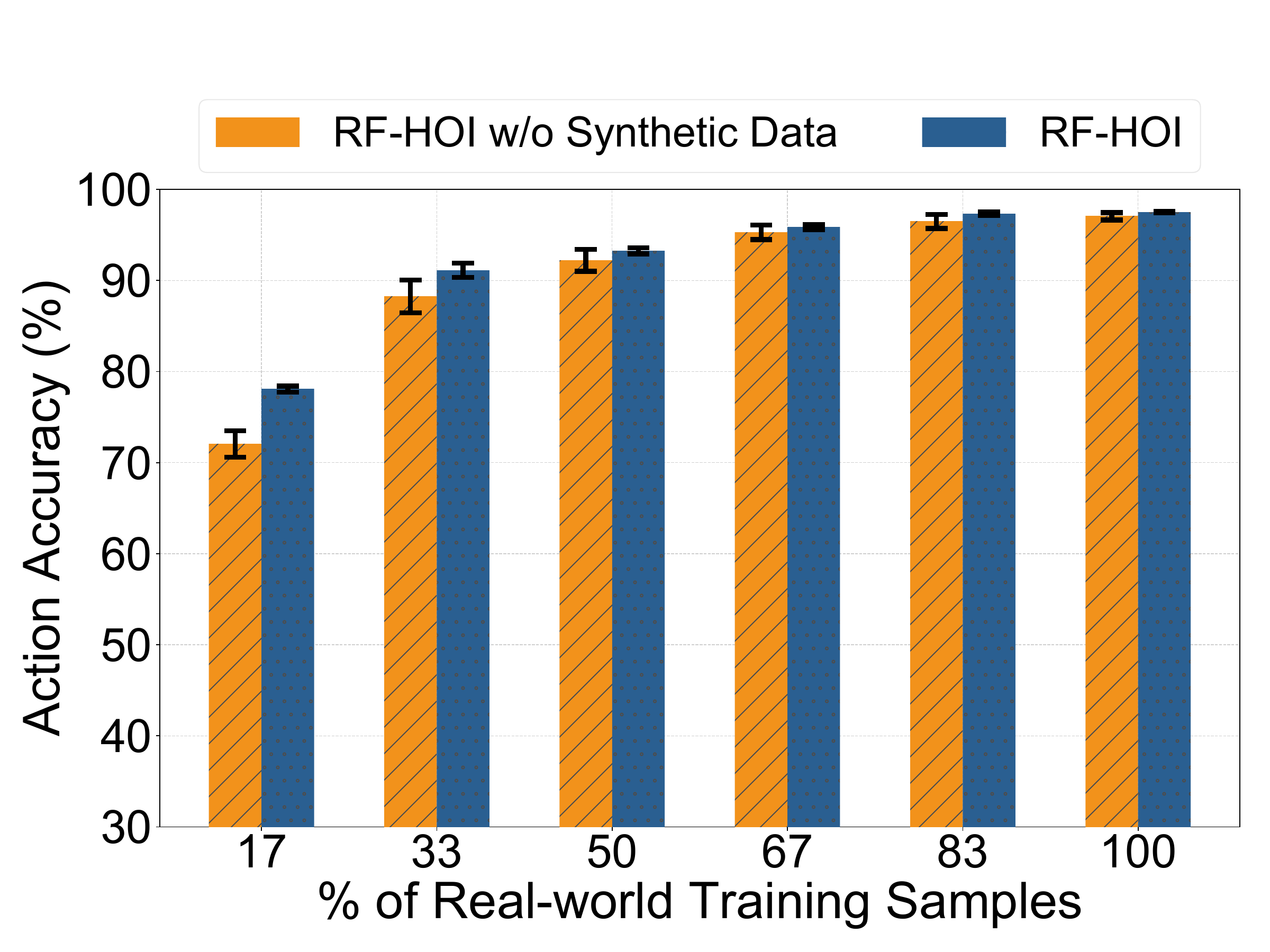} 
}
\subfigure[Object accuracy]{
\includegraphics[width=0.315\linewidth]{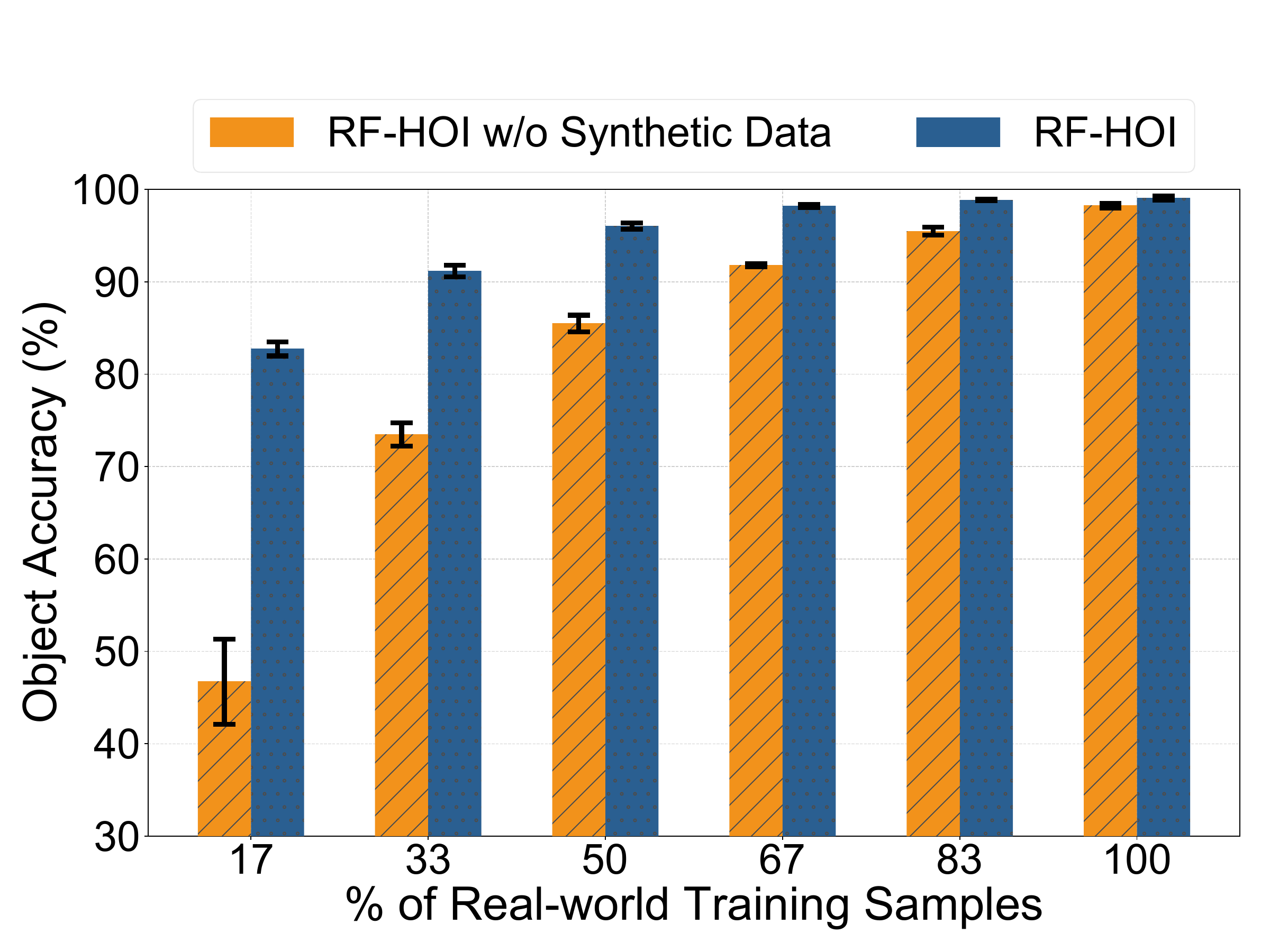}
}
\caption{Impact of training set size. \label{fig:exp_syn_size}}
\Description{Impact of training set size.}
\end{figure*}

\begin{figure*}
\centering
\subfigure[HOI accuracy]{
\includegraphics[width=0.315\linewidth]{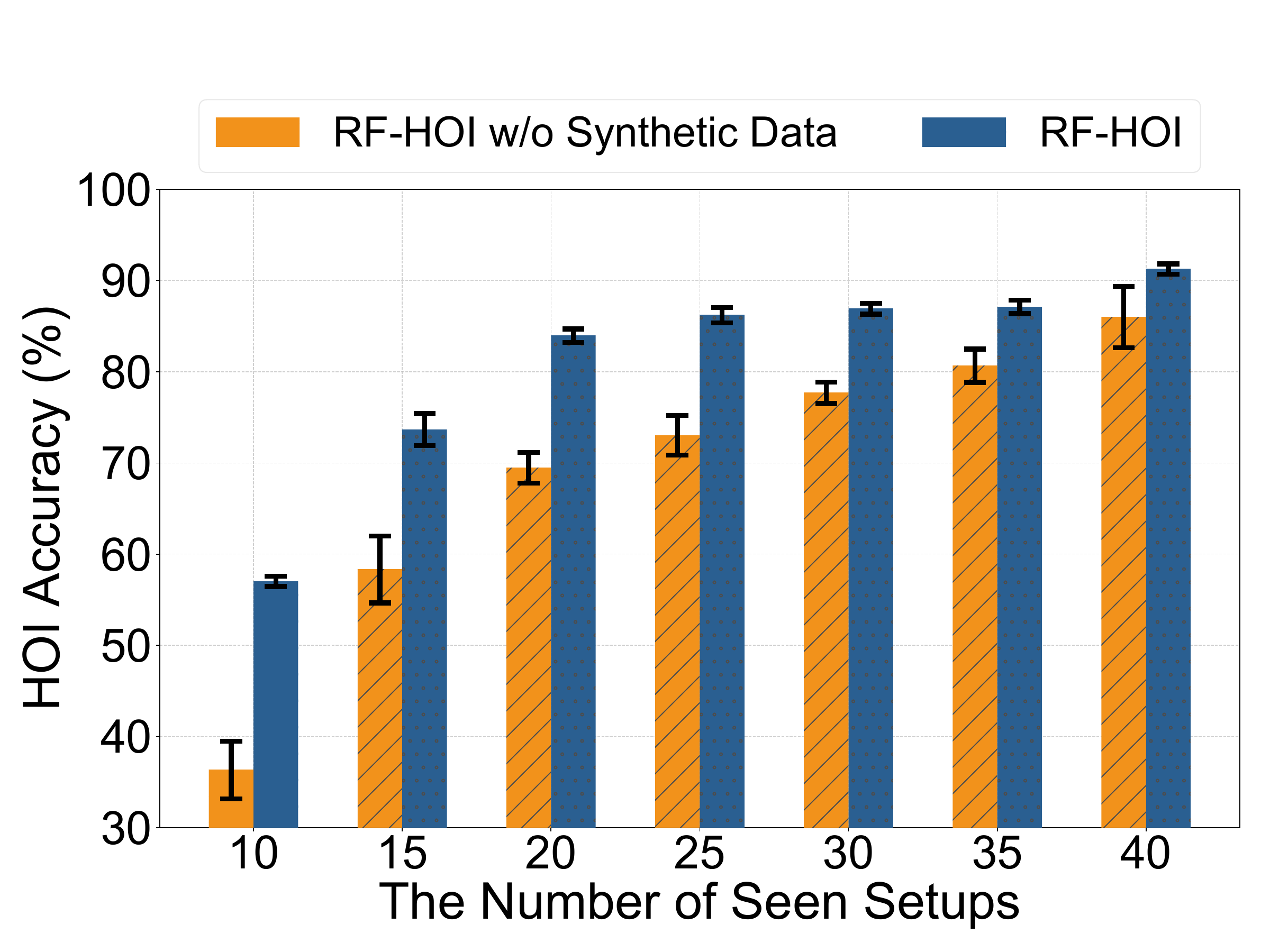} 
}
\subfigure[Action accuracy]{
\includegraphics[width=0.315\linewidth]{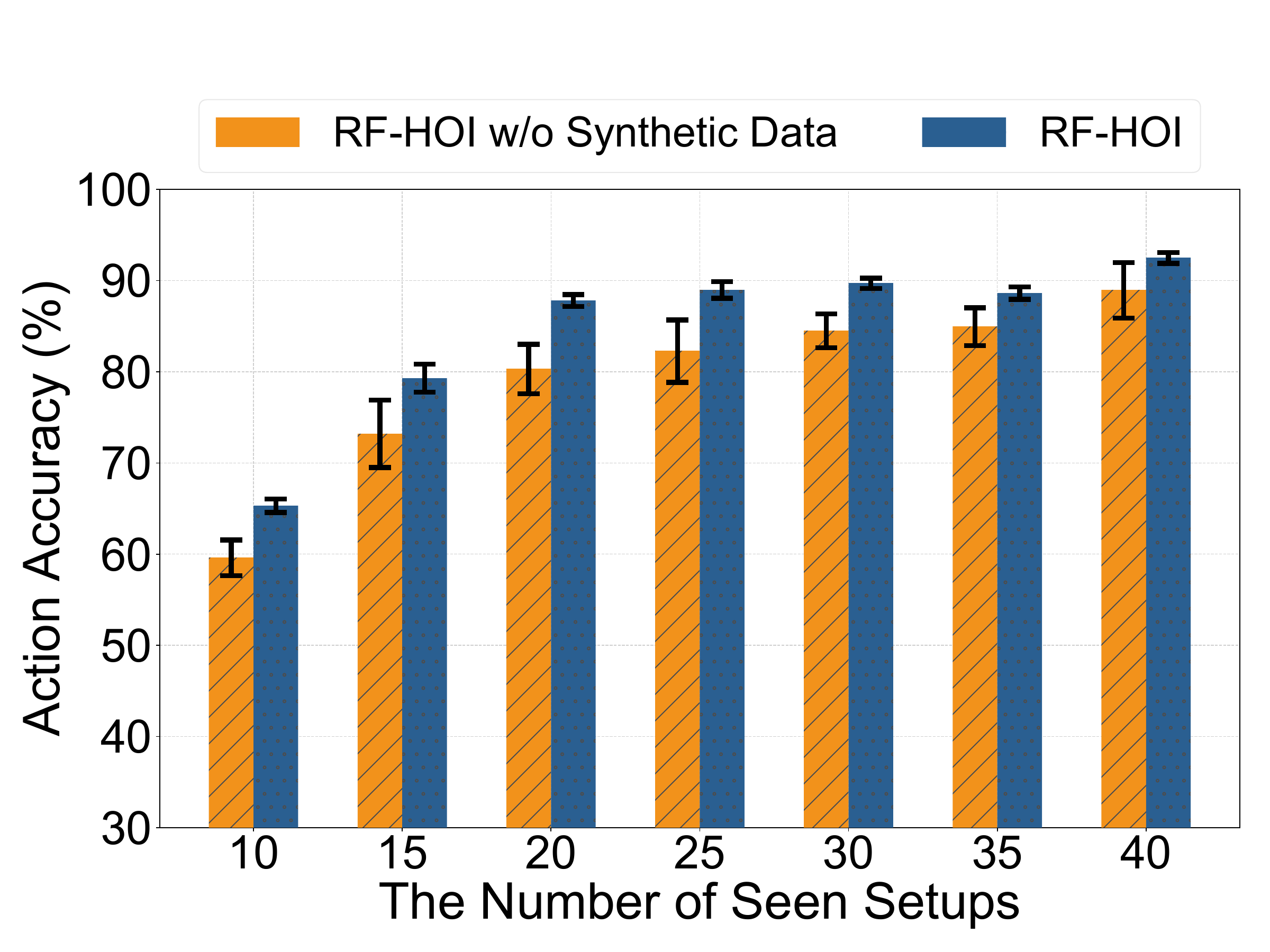} 
}
\subfigure[Object accuracy]{
\includegraphics[width=0.315\linewidth]{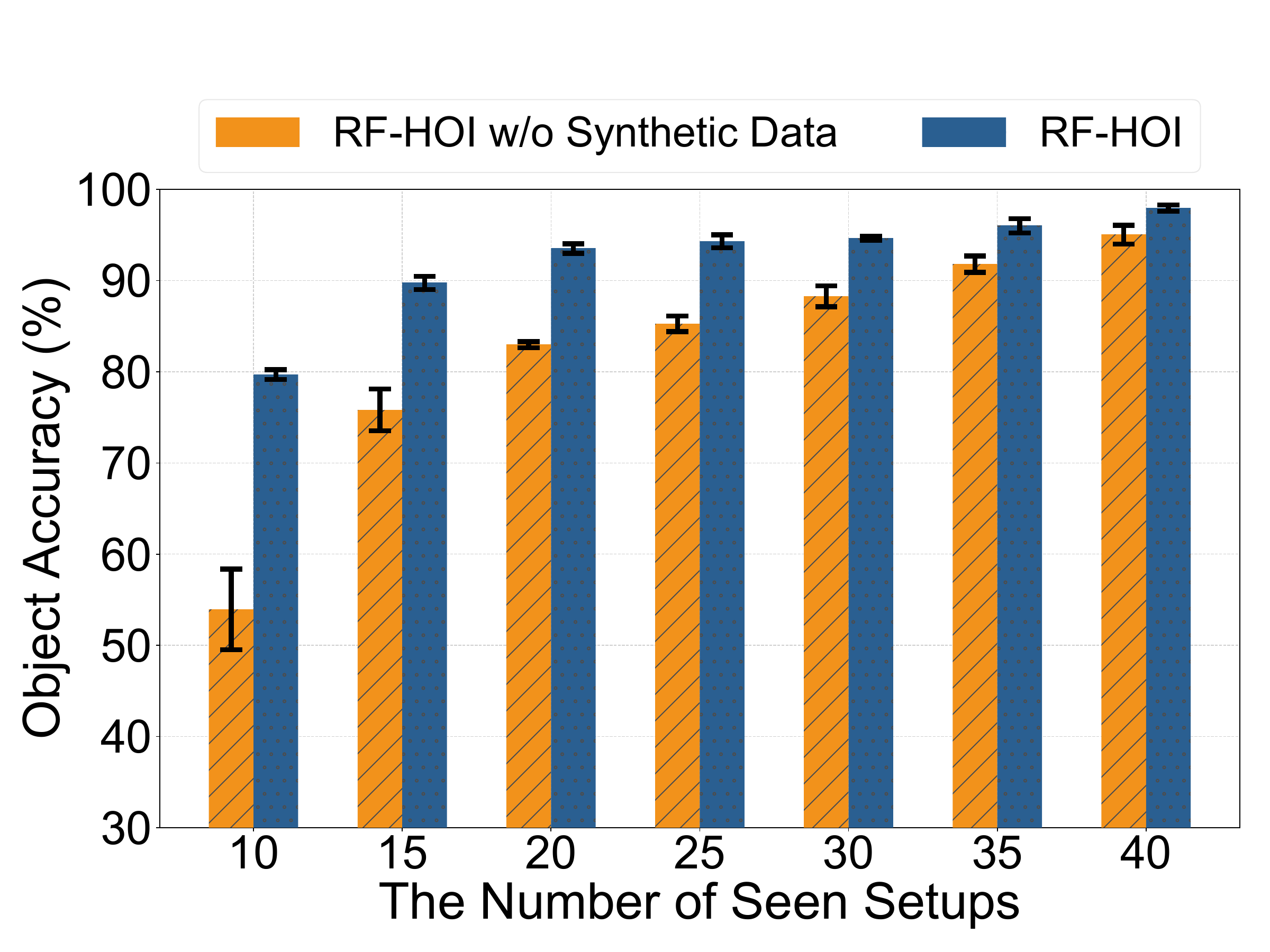}
}
\caption{Comparison between different numbers of unseen setups. \label{fig:exp_syn_seensetp}}
\Description{Comparison between unseen setups. }
\end{figure*}
\subsubsection{Performance Boost Under the Same Amount of Fine-Tuning Data}
Third, we demonstrate that pre-training with synthetic data significantly boosts \sysname{}'s performance, particularly when the available real-world training data is limited. Each experiment is repeated with five different random seeds to ensure reliability. By gradually increasing the percentage of real-world training samples, we report HOI accuracy, action accuracy, and object accuracy in Fig.~\ref{fig:exp_syn_size}. While both \sysname{} and the baseline improve as more real-world data are provided, \sysname{} consistently outperforms the baseline given the same number of samples, with average gains of 12.00\% in HOI accuracy, 1.97\% in action accuracy, and 12.48\% in object accuracy. The benefit is especially pronounced with fewer training samples: when only 17\% of real-world data is used, \sysname{} surpasses the baseline by 30.74\% in HOI accuracy, 6.00\% in action accuracy, and 36.01\% in object accuracy.

\subsubsection{Generalizability to Unseen Setups}
Finally, we show that incorporating synthetic data improves \sysname{}'s generalizability to unseen setups when the number of seen real-world training setups is limited. To evaluate this, we randomly select varying numbers of setups as seen for training, with the remaining setups reserved for testing. For each setup in the training set, five samples are included. We ensure that every possible value for most factors (including environment, user, and distance) appears in the training data, but not necessarily in all combinations. For the orientation factor, we specifically include $-90^\circ$, $0^\circ$, and $90^\circ$ instead of all orientations.
Each experiment is repeated with five different random seeds to ensure reliability.

As presented in Fig.~\ref{fig:exp_syn_seensetp}, when the number of seen setups increases, both \sysname{} and the baseline improve, but \sysname{} consistently achieves higher accuracy on the test (unseen) setups. On average, \sysname{} outperforms the baseline by 12.09\% in HOI accuracy, 5.50\% in action accuracy, and 10.40\% in object accuracy. The advantage is especially marked with fewer seen setups; for example, with only 20 seen setups, \sysname{} surpasses the baseline by 14.48\% in HOI accuracy, 7.51\% in action accuracy, and 10.54\% in object accuracy.

\subsection{Micro-benchmark Performance}
Here, we would evaluate the performance of \sysname{} under various system settings.

\begin{figure*}
    \centering
    \begin{minipage}[t]{0.28\linewidth}
	\centering 
               \includegraphics[width=\linewidth]{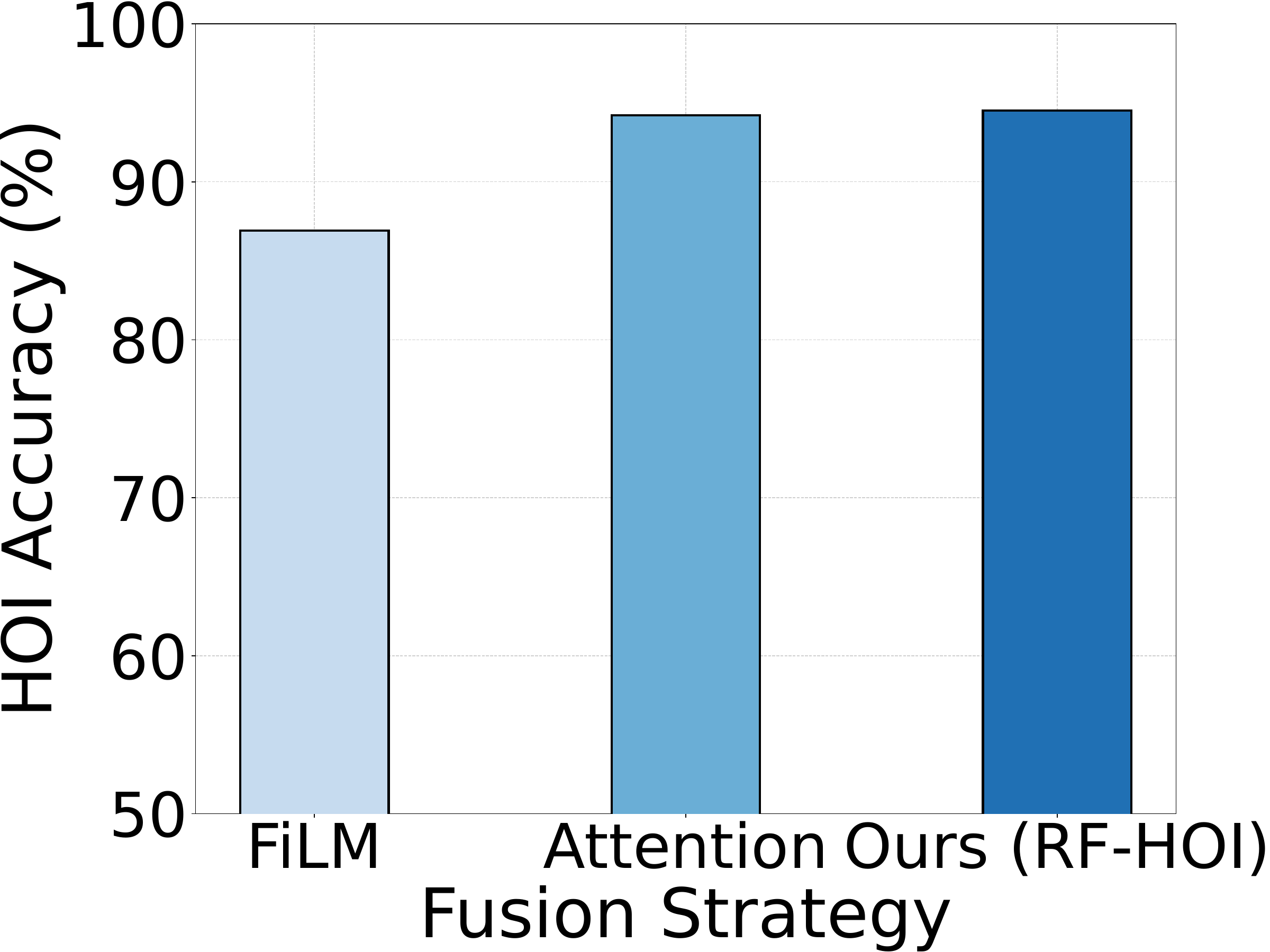}
        \caption{\edited Impact of fusion strategies. \label{fig:exp_micro_fusion}}
    \end{minipage}
    \hspace{0.03\linewidth}
    \begin{minipage}[t]{0.28\linewidth}
	\centering 
        \includegraphics[width=\linewidth]{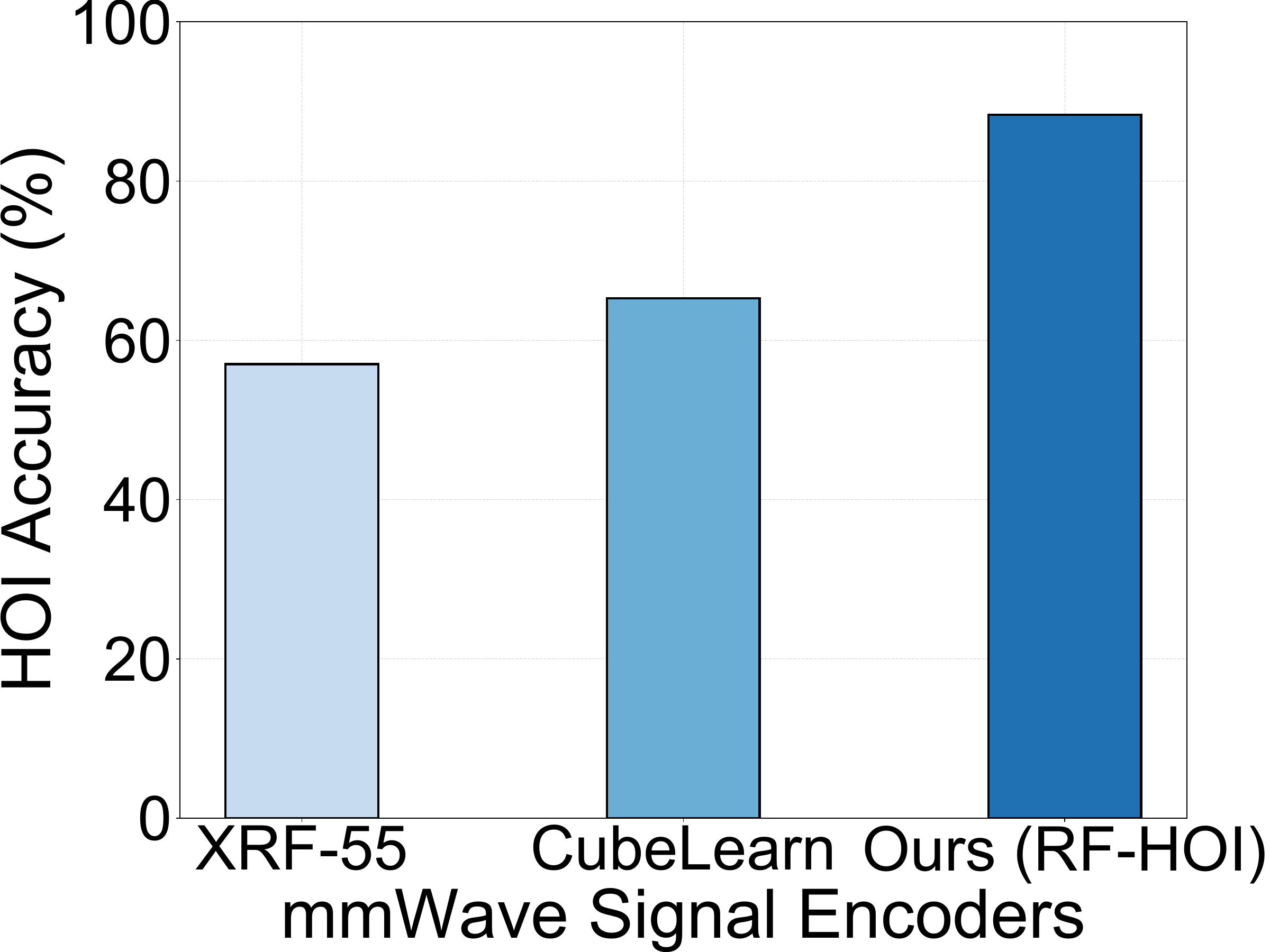}
        \caption{Comparison for different mmWave signal encoders.\label{fig:exp_micro_encoder}}
    \end{minipage}
    \hspace{0.03\linewidth}
    \begin{minipage}[t]{0.28\linewidth}
	\centering 
        \includegraphics[width=\linewidth]{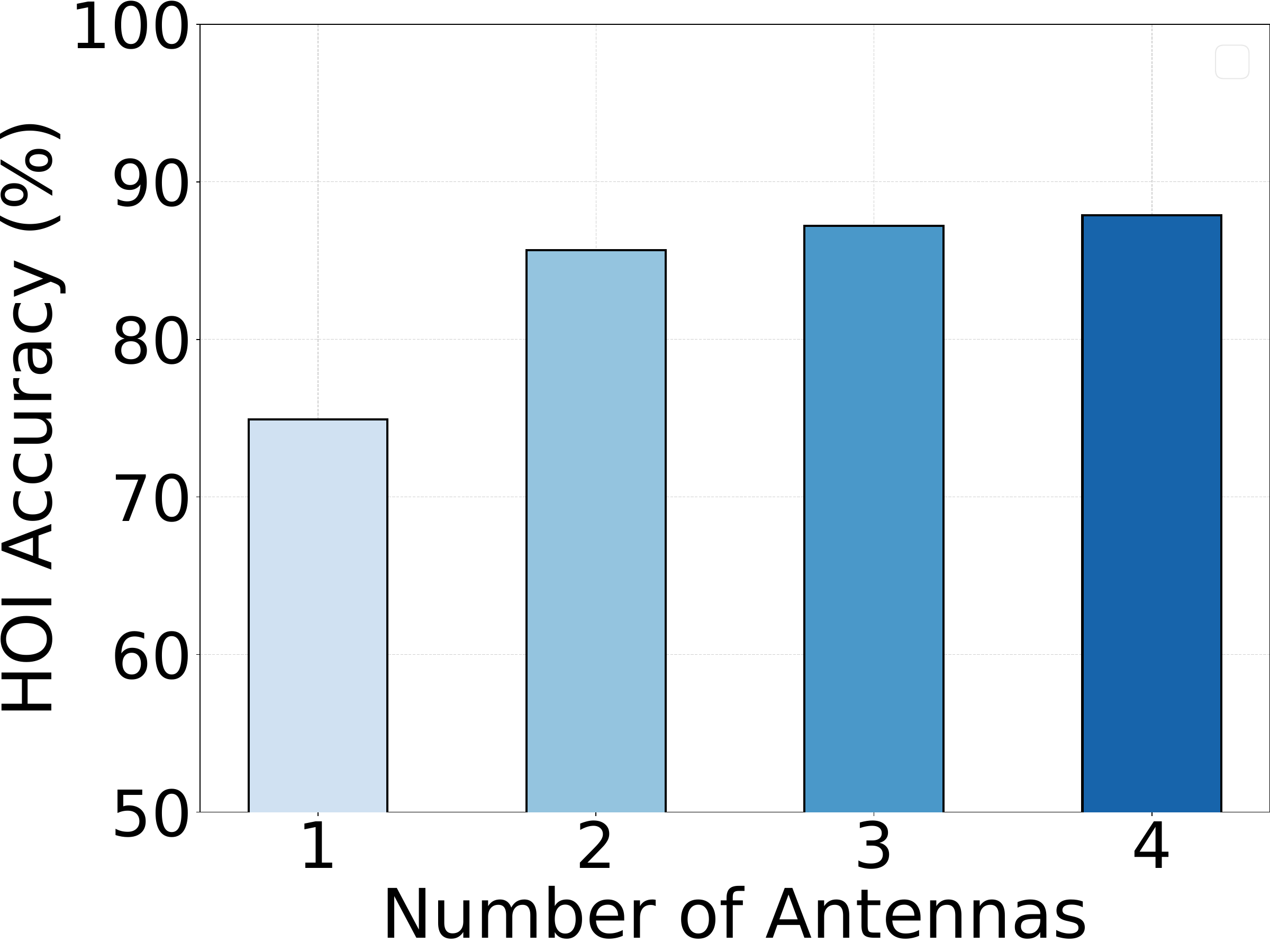}
        \caption{Comparison for different numbers of antenna.\label{fig:exp_micro_antenna}}
        
          \Description{Microbenchmark 2}
    \end{minipage}
\end{figure*}

{\edited 
\subsubsection{Impact of Fusion Strategies}
To validate the design choice of the modality fusion module in \sysname{}, we compare it with two widely adopted state-of-the-art fusion strategies for multimodal learning:
\\
(1) \textbf{FiLM}~\cite{perez2018film,fan2026mmpred}: 
Feature-wise Linear Modulation (FiLM), conditions the feature representations of one modality using linear layers as feature-wise affine transformations generated from the other modality.\\
(2) \textbf{Attention}~\cite{lu2019vilbert,imran2025swinlstm}: Attention-based fusion models cross-modal interactions by allowing one modality to attend to the other. We implement a cross-attention-based fusion module at the representation level.

For all baselines and \sysname{}, we adopt a consistent training pipeline. Specifically, all models are first pre-trained on the same synthetic dataset, and then fine-tuned and evaluated on the same real-world dataset. During fine-tuning, we use 67\% of the real-world training samples. To ensure a fair comparison, all models share identical modality-specific encoders and training protocols, with the fusion module being the only component that differs.

As illustrated in Fig.~\ref{fig:exp_micro_fusion}, \sysname{} achieves the highest accuracy among the three fusion strategies, while the performance gap between \sysname{} and attention-based fusion remains below 1\%. This observation indicates that, under the current experimental setting, more complex fusion mechanisms provide only marginal improvements over our lightweight design. We note that the proposed framework is modular and can readily incorporate more expressive fusion designs in the future.
}

\subsubsection{Impact of mmWave Encoders}
To validate our design for the mmWave branch, we compare \sysname{} with two state-of-the-art encoders proposed in previous HAR solutions:
\\
(1) \textbf{XRF-55}~\cite{wang2024xrf55}: a multimodal network for HAR, where we substitute its mmWave branch into our model. Since XRF-55 uses CNNs for temporal feature extraction, input samples are cropped to a fixed length for compatibility.\\
(2) \textbf{CubeLearn}~\cite{zhao2023cubelearn}: a model that processes raw complex-valued mmWave data via a learnable preprocessing module, followed by LSTM~\cite{hochreiter1997long} blocks to track temporal features. 

All models are trained from scratch using 67\% of the real-world training samples. As illustrated in Fig.~\ref{fig:exp_micro_encoder}, \sysname{} achieves 31.30\% higher accuracy than XRF-55, and surpasses CubeLearn by 23.04\%. The superior performance over XRF-55 can be attributed to \sysname{}'s transformer blocks, which handle variable-length samples more effectively. In comparison to CubeLearn, our design's efficient preprocessing module enables the model to learn decision boundaries more accurately when the dataset is limited.

\subsubsection{Impact of the Number of RFID Antennas}

{\edited 
In this section, we evaluate the impact of using different numbers of RFID antennas. The model is trained from scratch using 67\% of the real-world training samples, with the number of antennas varied from one to four. As shown in Fig.~\ref{fig:exp_micro_antenna}, the HOI accuracies corresponding to 1, 2, 3, and 4 antennas are 74.91\%, 85.66\%, 87.20\%, and 87.87\%, respectively.

Notably, increasing the number of antennas from one to two yields the largest performance improvement (10.75\%), while further increasing the array size from two to four results in only marginal gains (less than 3\%). This suggests that, although we use four antennas in our default configuration to achieve the best performance, the proposed system does not fundamentally rely on a large antenna array. In practice, a compact configuration with two antennas already achieves competitive performance, substantially reducing system complexity and hardware cost while retaining most of the recognition accuracy.
}

\begin{figure}[t]
    \centering
        \begin{minipage}[t]{0.4\linewidth}
	\centering
\includegraphics[width=\linewidth]{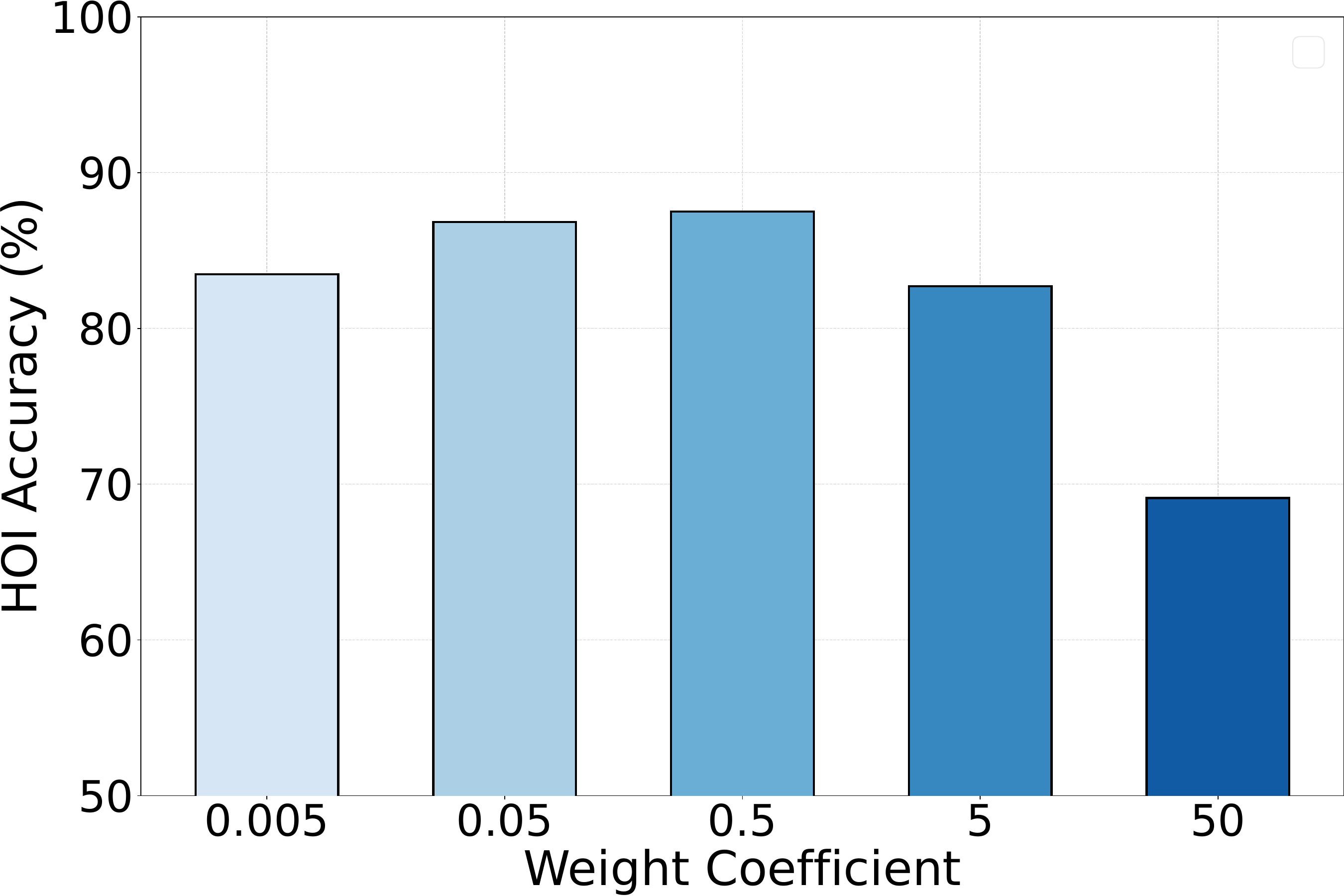}
        \caption{Impact of weight coefficient $\alpha$ of loss function.\label{fig:exp_micro_weight}}
    \end{minipage}
    \hspace{0.03\linewidth}
        \begin{minipage}[t]{0.4\linewidth}
        \centering      \includegraphics[width=\linewidth]{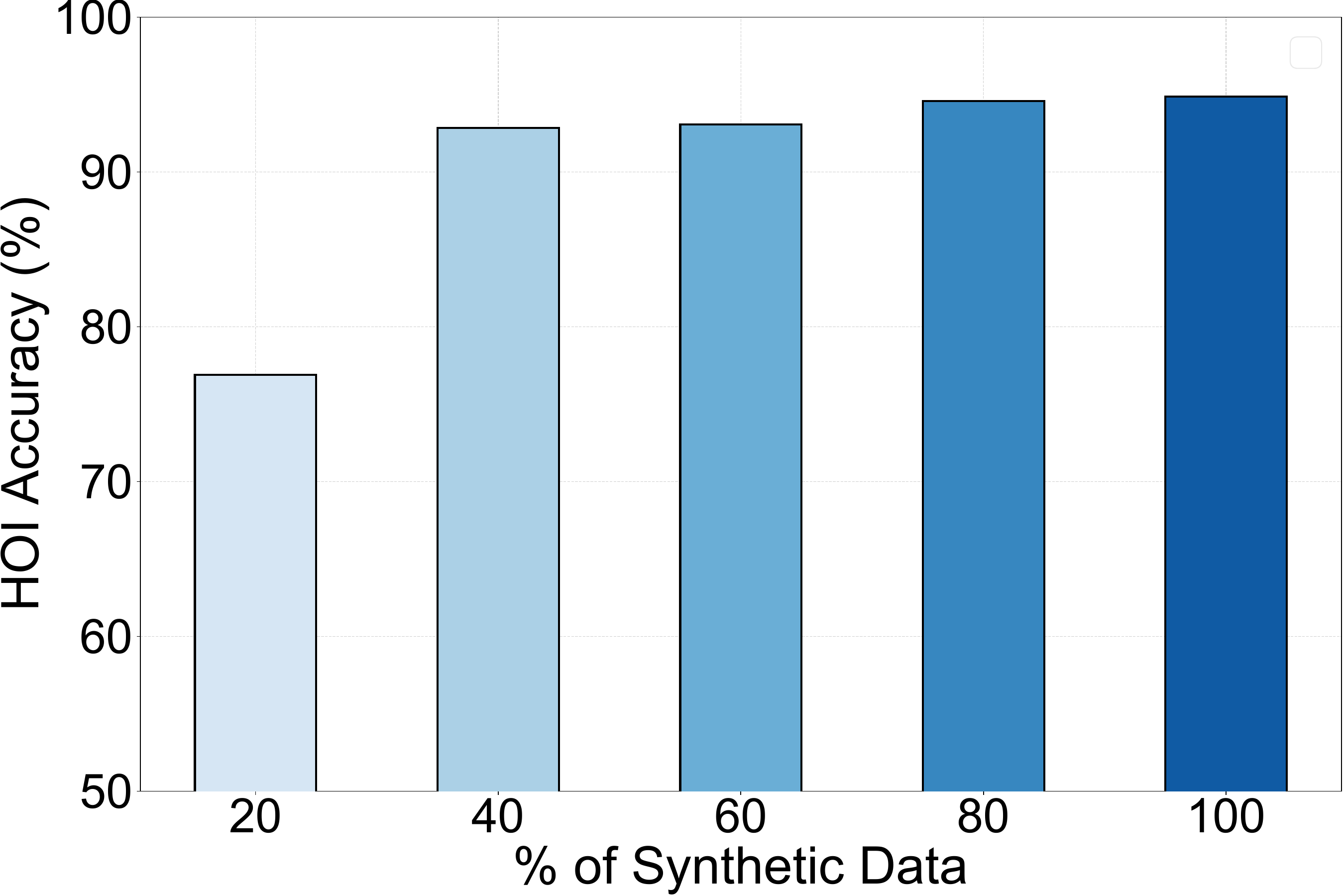}
        \caption{Impact of pre-training dataset size. \label{fig:exp_micro_synthetic}}
          \Description{Microbenchmark 2}
    \end{minipage}
\end{figure}
\subsubsection{Impact of Balancing Action Classification and Target Identification}

In this section, we evaluate the effect of $\alpha$, the hyperparameter controlling the balance between $loss_{action}$ and $loss_{target}$ in Equation~\ref{equ:loss}. The model is trained from scratch using 67\% of the real-world training samples. As shown in Fig.~\ref{fig:exp_micro_weight}, increasing $\alpha$ initially improves HOI accuracy, which then drops sharply beyond a certain point. More specifically, when the value of $\alpha$ increases from 0.005 to 50, starting at 83.47\%, the corresponding HOI accuracy first rises to 87.50\%  and then drops to 69.11\%. Notably, a larger (or smaller) $\alpha$ causes the model to prioritize action (or target) classification, respectively. We therefore select a moderate value $\alpha=0.5$, as both action and target identification are essential for robust HOI recognition.

\subsubsection{Impact of the Pre-Training Dataset Size}

In this section, we evaluate how the scale of the pre-training dataset affects performance. In this experiment, we vary the percentage of synthetic samples used for pre-training. After pre-training, all models are fine-tuned with 67\% of the real-world training samples and evaluated on the real-world test set. As shown in Fig.~\ref{fig:exp_micro_synthetic}, increasing the size of the pre-training dataset leads to a sharp improvement in HOI accuracy, especially when the number of synthetic training samples is limited. For example, the accuracy rises from 76.90\% to 92.84\% as the proportion of synthetic data increases from 20\% to 40\%. And the performance gains gradually saturate and plateau around 94\%. This result indicates that our synthesized dataset is sufficiently large for current experimental settings; however, expanding the synthetic dataset may be necessary if more HOI categories or setups are introduced in the future.

{\edited
\subsection{Robustness to Occlusion and Dynamic Environment}
\subsubsection{Impact of People Occlusion}

We evaluate the robustness of \sysname{} under three people-occlusion conditions: \\
(1) \textbf{No occlusion}: All RFID tags remain unobstructed. \\
(2) \textbf{Hand occlusion}: The user partially covers the RFID tag on the target object with fingers during interaction. \\
(3) \textbf{Body occlusion}: Tagged non-target objects are placed behind the user, resulting in body-induced occlusion along the antenna--user--object path. In the body-occlusion setting, the distance between the user and the occluded objects ranges from approximately 20 to 50\,cm.
\\
For all three conditions, we collect a total of 270 real-world samples, covering three user orientations and four actions that involve direct hand–object contact: pick, put, open, and close.

We first analyze the impact of people occlusion at the signal level.
Fig.~\ref{fig:exp_occ_rssi} compares the RSSI distributions of the target object under no occlusion and hand occlusion. When the tag is partially covered by the user’s hand, the RSSI distribution exhibits a clear left shift, with an average attenuation of 8.30\,dB. This result confirms that hand occlusion introduces substantial signal attenuation on the target tag, while still allowing the RFID signal to remain readable within the considered range. 
This observation is consistent with prior works~\cite{li2016deep,wang2018modeling,li2020reloc}, which show that RSSI is more susceptible to environmental interference than to motion patterns, and therefore provides limited discriminative power for interaction recognition.

\begin{figure}[t]
    \centering
        \begin{minipage}[t]{0.4\linewidth}
	\centering
        \includegraphics[width=\linewidth]{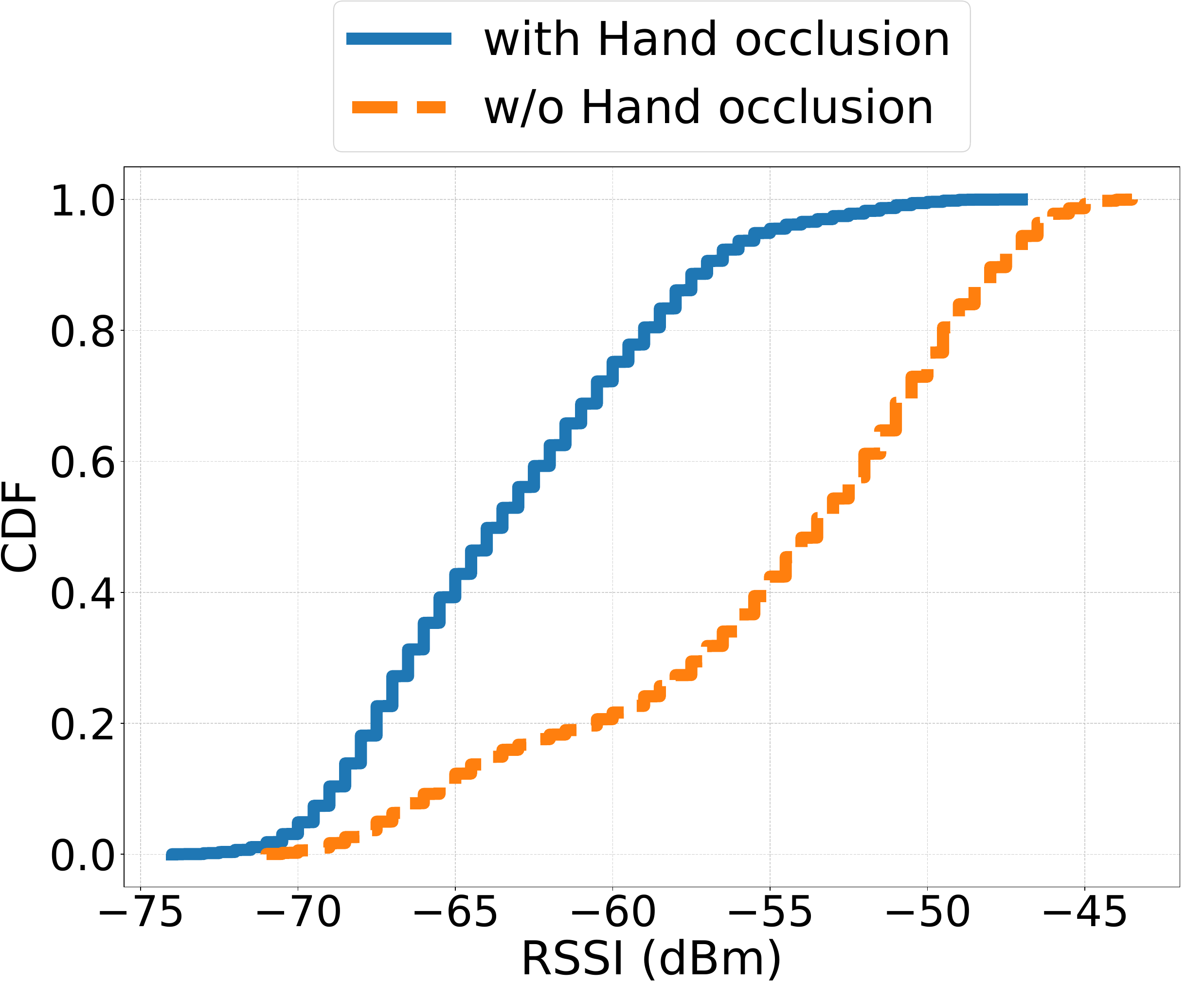}
        \caption{\edited Impact of hand occlusion on RFID RSSI.\label{fig:exp_occ_rssi}}
    \end{minipage}
    \hspace{0.03\linewidth}
        \begin{minipage}[t]{0.4\linewidth}
        \centering
        \includegraphics[width=\linewidth]{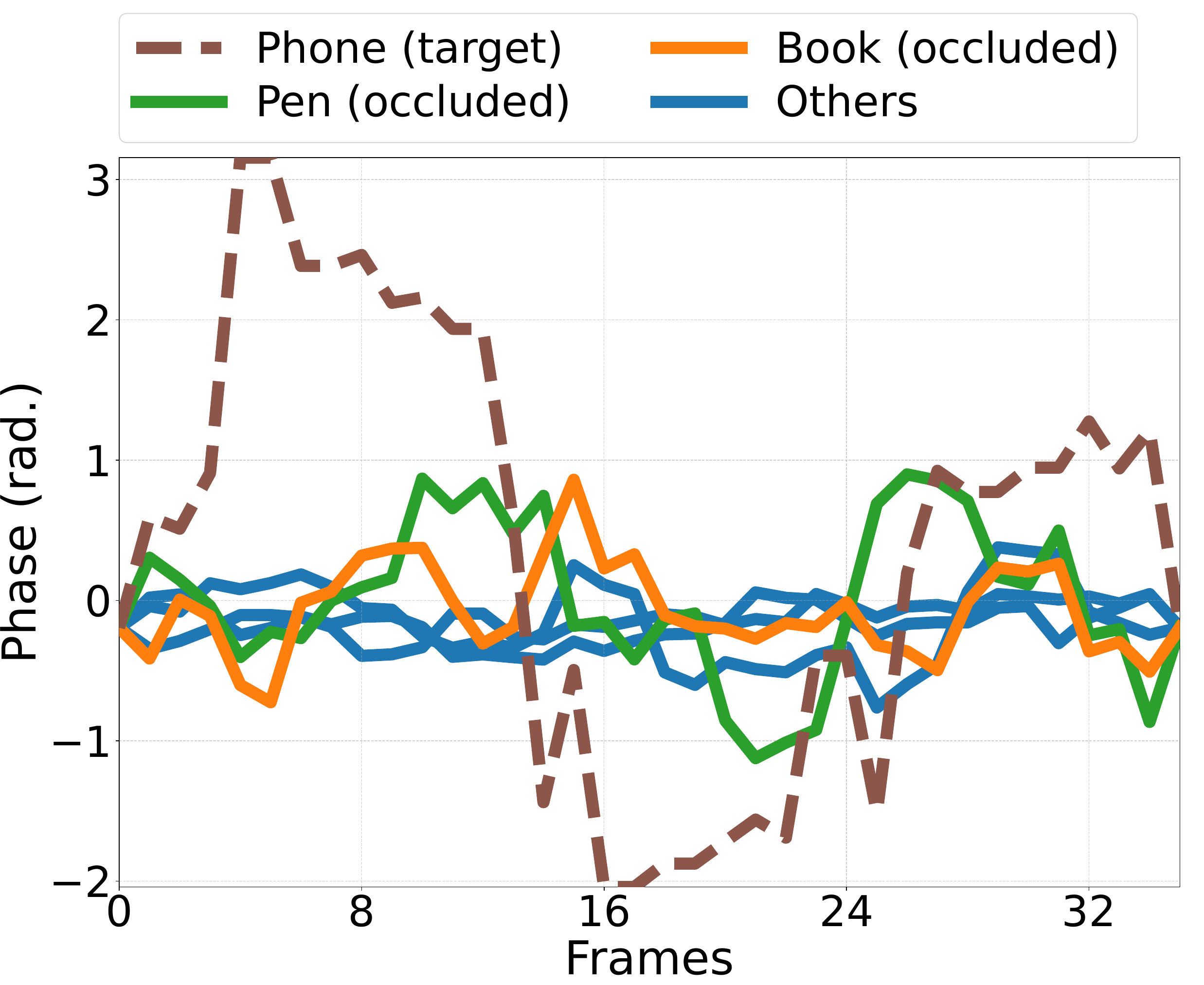}
        \caption{\edited Normalized tag phase differences under body occlusion. \label{fig:exp_occ_phase}}
          \Description{Microbenchmark 2}
    \end{minipage}
\end{figure}
\begin{figure}[t]
    \centering
        \begin{minipage}[t]{0.4\linewidth}
	\centering
        \includegraphics[width=\linewidth]{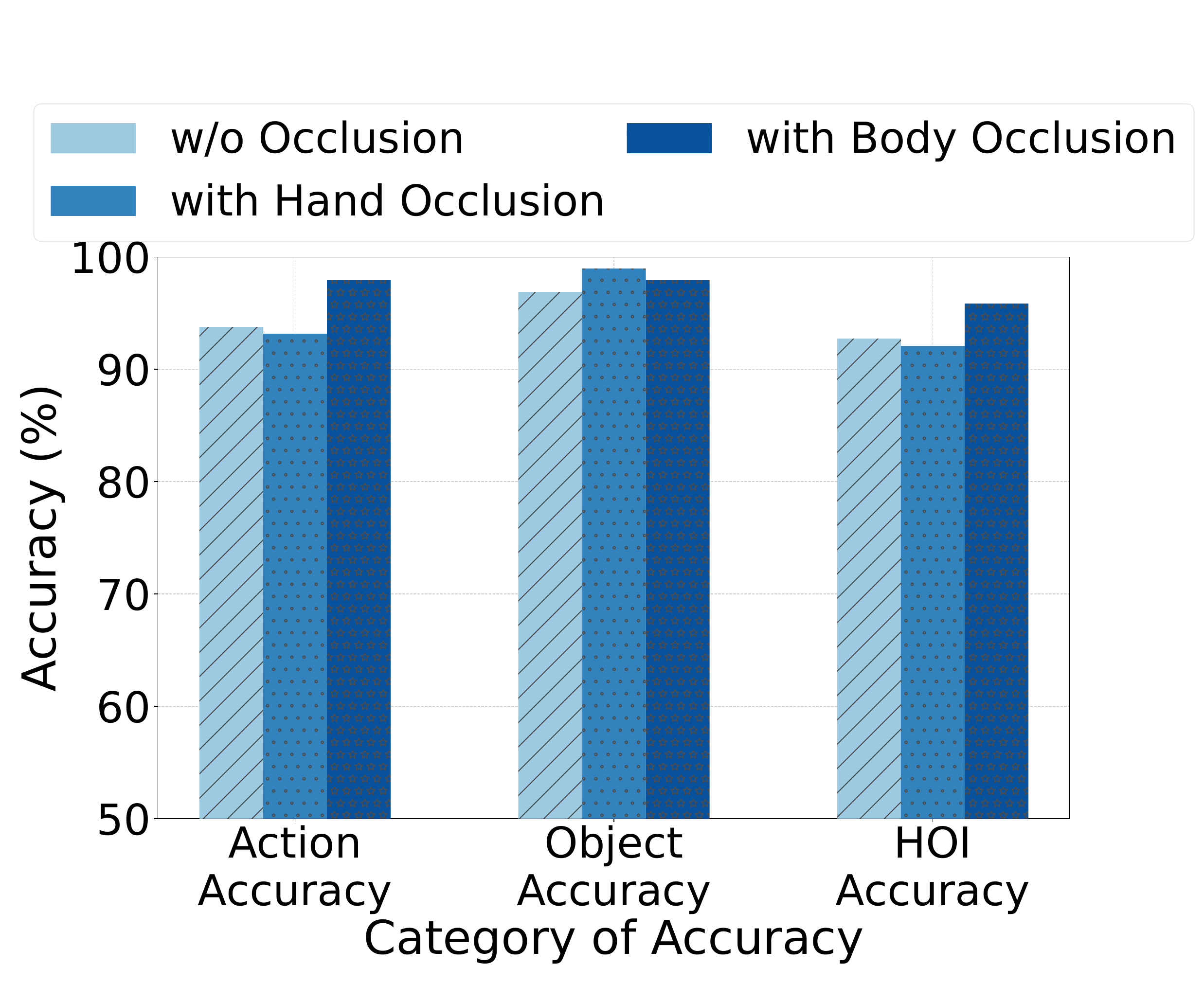}
        \caption{\edited Impact of human occlusion. \label{fig:exp_occlusion}}
    \end{minipage}
    \hspace{0.03\linewidth}
        \begin{minipage}[t]{0.4\linewidth}
        \centering
        \includegraphics[width=\linewidth]{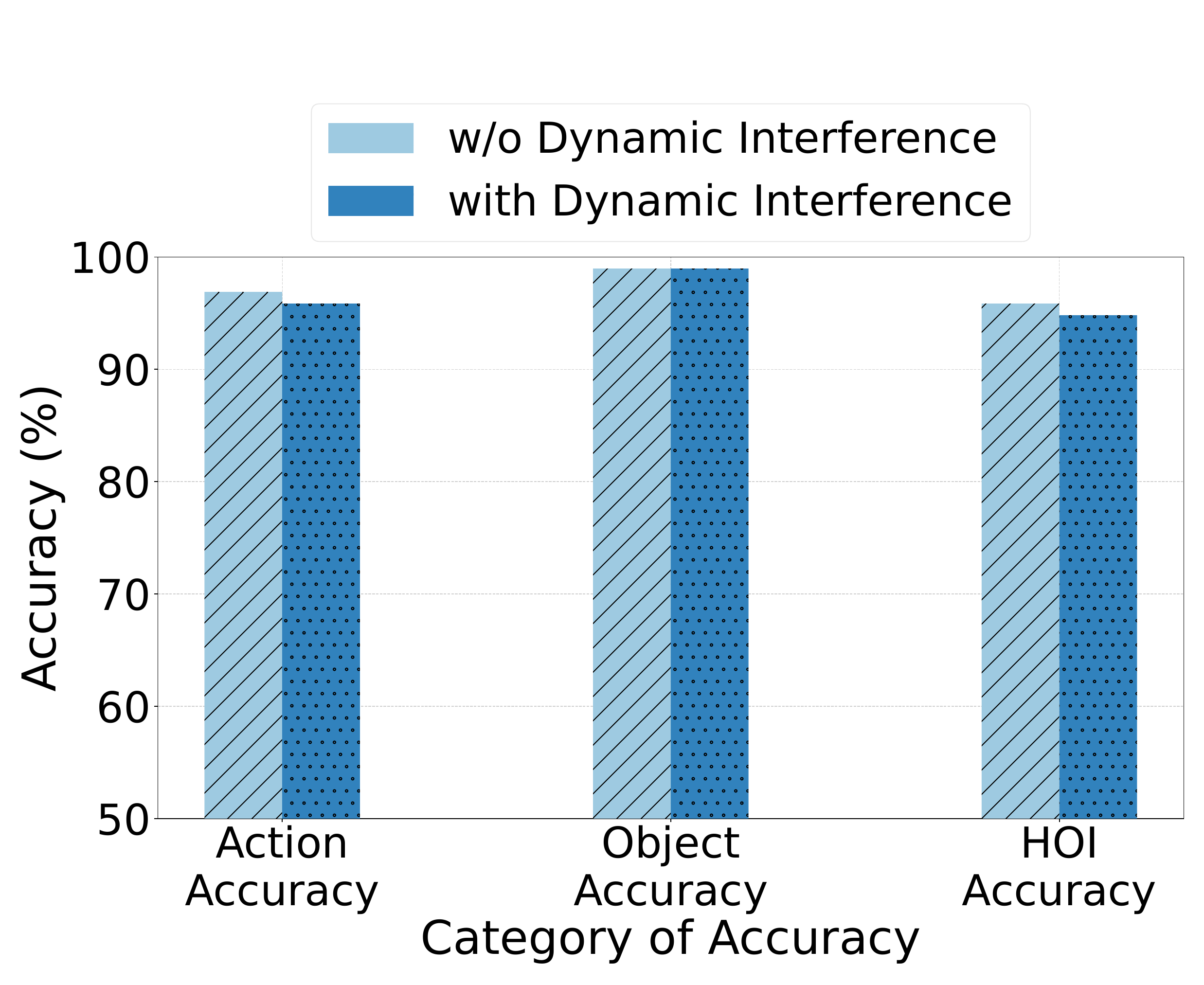}
        \caption{Test \sysname{} with dynamic inference. \label{fig:exp_env_test}}
          \Description{Microbenchmark 2}
    \end{minipage}
\end{figure}
We note that the design choice of \sysname{} to use phase features is primarily motivated by their higher sensitivity to object motion, which provides more discriminative cues for interaction recognition, rather than by occlusion considerations alone.
Fig.~\ref{fig:exp_occ_phase} visualizes one representative example of normalized RFID phase differences from one antenna to multiple objects under body occlusion. Although the non-target objects placed behind the user exhibit noticeable phase fluctuations due to body-induced dynamic occlusion, the target object (phone) consistently shows larger phase variations than the occluded non-target objects, such as pen and book. This preserves a clear contrast between the target and non-target objects.

We then present action, object, and HOI accuracy under the three occlusion conditions in Fig.~\ref{fig:exp_occlusion}. Across all settings, \sysname{} achieves over 90\% HOI accuracy, indicating strong robustness to people-induced occlusion. We observe that hand occlusion only slightly degrades HOI accuracy, likely because the phase variations of the target object’s tag remain observable, even though hand occlusion attenuates the RSSI.
In addition, body occlusion does not degrade performance and even leads to a slight improvement. A possible explanation is that placing non-target objects behind the user reduces their interference, while the mmWave radar’s line of sight to the human body remains unobstructed, allowing reliable capture of motion features.

Overall, these results demonstrate that \sysname{} can maintain stable HOI recognition performance under realistic people-occlusion scenarios, including both direct hand occlusion on the target tag and body-induced occlusion affecting non-target objects.

\subsubsection{Impact of Dynamic Multipath Interference}
We evaluate the robustness of \sysname{} under dynamic and multi-user environments, addressing the concern that non-target objects or surrounding people may introduce time-varying multipath effects in real-world settings. Although \sysname{} focuses on single-user HOI recognition, it does not assume a static environment and is designed to tolerate moderate dynamic interference.

In this experiment, we ask one volunteer to perform HOI interactions in 105 samples, covering three user orientations and seven actions. To introduce dynamic multipath interference, a second volunteer is asked to perform random daily activities involving non-target objects (e.g., drinking from a bottle or typing on a keyboard) around the interacting user at a distance of approximately 1--2 meters. These objects are not equipped with RFID tags and only serve as dynamic reflectors in the environment. As a baseline, we also collect 105 samples under identical setups without dynamic multipath interference.
Fig.~\ref{fig:exp_env_test} compares the performance of \sysname{} under the two conditions. The results show that the difference in HOI accuracy between static and dynamic multipath settings is within 1\%. This indicates that \sysname{} can maintain stable recognition performance even in the presence of additional users and mild environment dynamics.

Overall, this experiment demonstrates that \sysname{} is robust to slight dynamic multipath interference and can generalize beyond strictly static, single-user environments, supporting its applicability in more realistic real-world scenarios.
 }

\subsection{System Computation Overhead}

\subsubsection{Evaluation on Fine-Tuning Time Cost}

We evaluate the time required to fine-tune \sysname{} for real-world deployment while maintaining strong performance, highlighting its adaptability. Using 67\% of the real-world training samples, we fine-tune the pre-trained model for varying durations. As shown in Fig.~\ref{fig:exp_duration}, the accuracy of HOI initially increases with fine-tuning time and steadily levels off after approximately 20 minutes. Specifically, as the fine-tuning duration increases from 5 minutes to 25 minutes, the corresponding HOI accuracy improves from 84.49\% to 95.09\%, with the most swift gain occurring in the first 15 minutes, and then converges.
In practice, this enables rapid adaptation to new setups by collecting a small amount of real-world data for fine-tuning. Therefore, the low time cost for fine-tuning facilitates quick and flexible deployment of \sysname{} across various setups.

\subsubsection{Evaluation on System Latency}

We measure the system latency of \sysname{}, including both data preprocessing and model inference times. Although the MUSIC algorithm is computationally intensive, we implement the mmWave preprocessing pipeline in PyTorch to leverage GPU acceleration, reducing the average preprocessing time for one sample from 13.39 seconds using CPU to 1.75 seconds using GPU on a Dell G15 laptop~\cite{laptop}.

\begin{figure}[t]
    \centering
        \begin{minipage}[t]{0.28\linewidth}
        \centering
        \includegraphics[width=\linewidth]{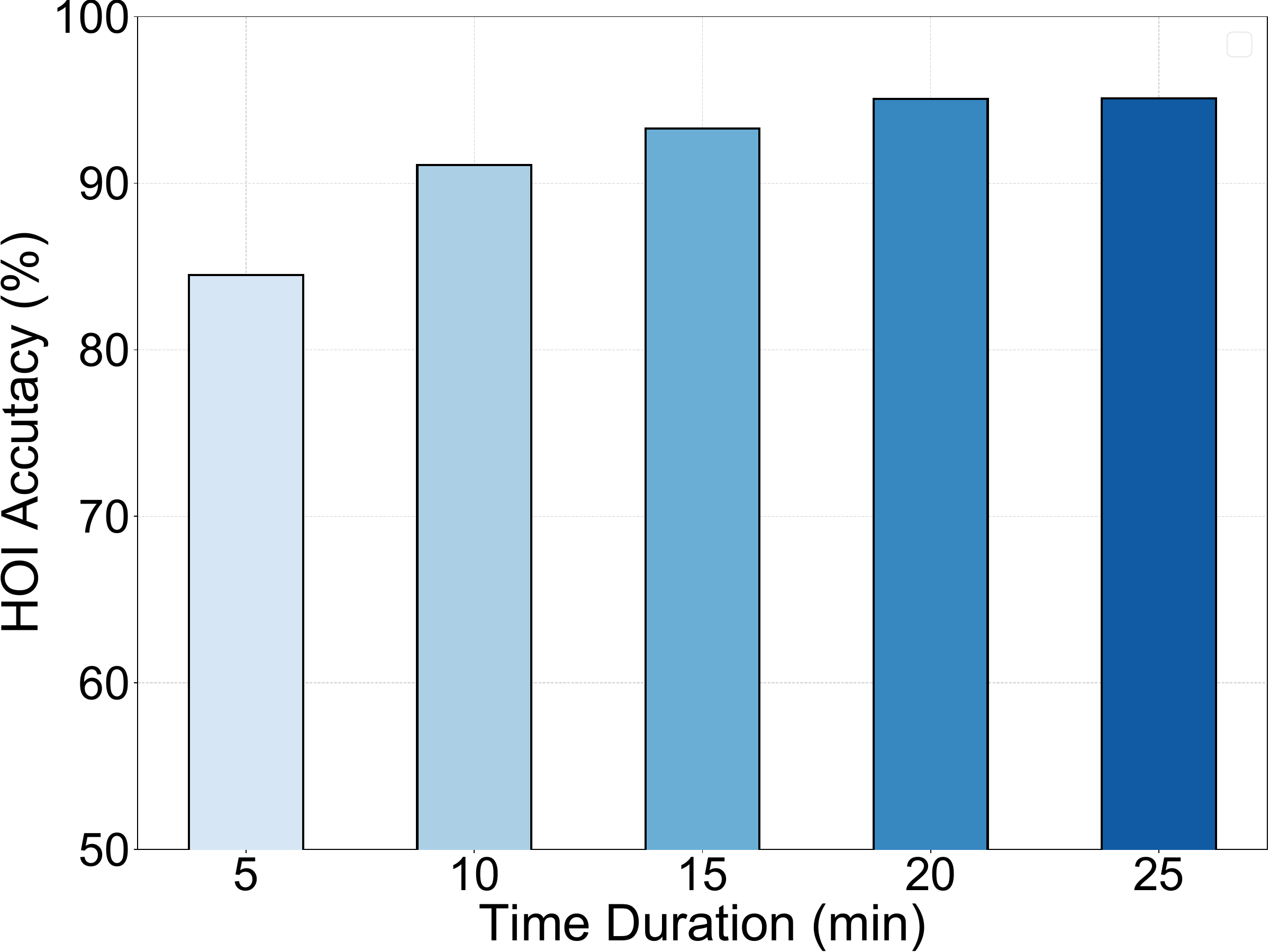}
        \caption{Impact of fine-tuning duration.}\label{fig:exp_duration}
    \Description{Impact of fine-tuning duration.}
    \end{minipage}
      \hspace{0.03\linewidth}
    \begin{minipage}[t]{0.28\linewidth}
        \centering
        \includegraphics[width=\linewidth]{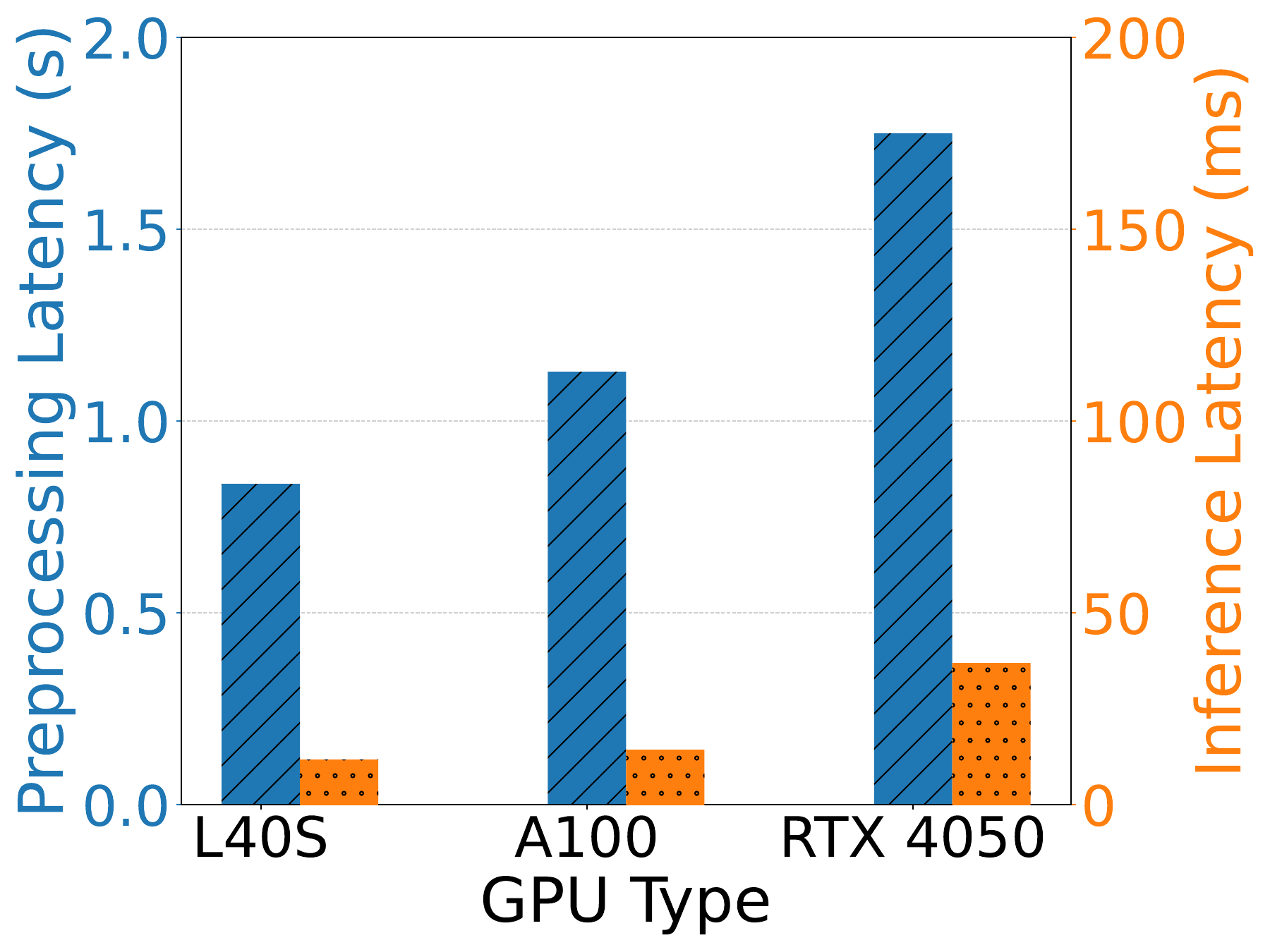}
        \caption{\edited Sample level latency on preprocessing and model inference.}\label{fig:exp_decoupled_latency}
    \Description{Inference Latency}
    \end{minipage}
          \hspace{0.03\linewidth}
    \begin{minipage}[t]{0.28\linewidth}
        \centering
        \includegraphics[width=\linewidth]{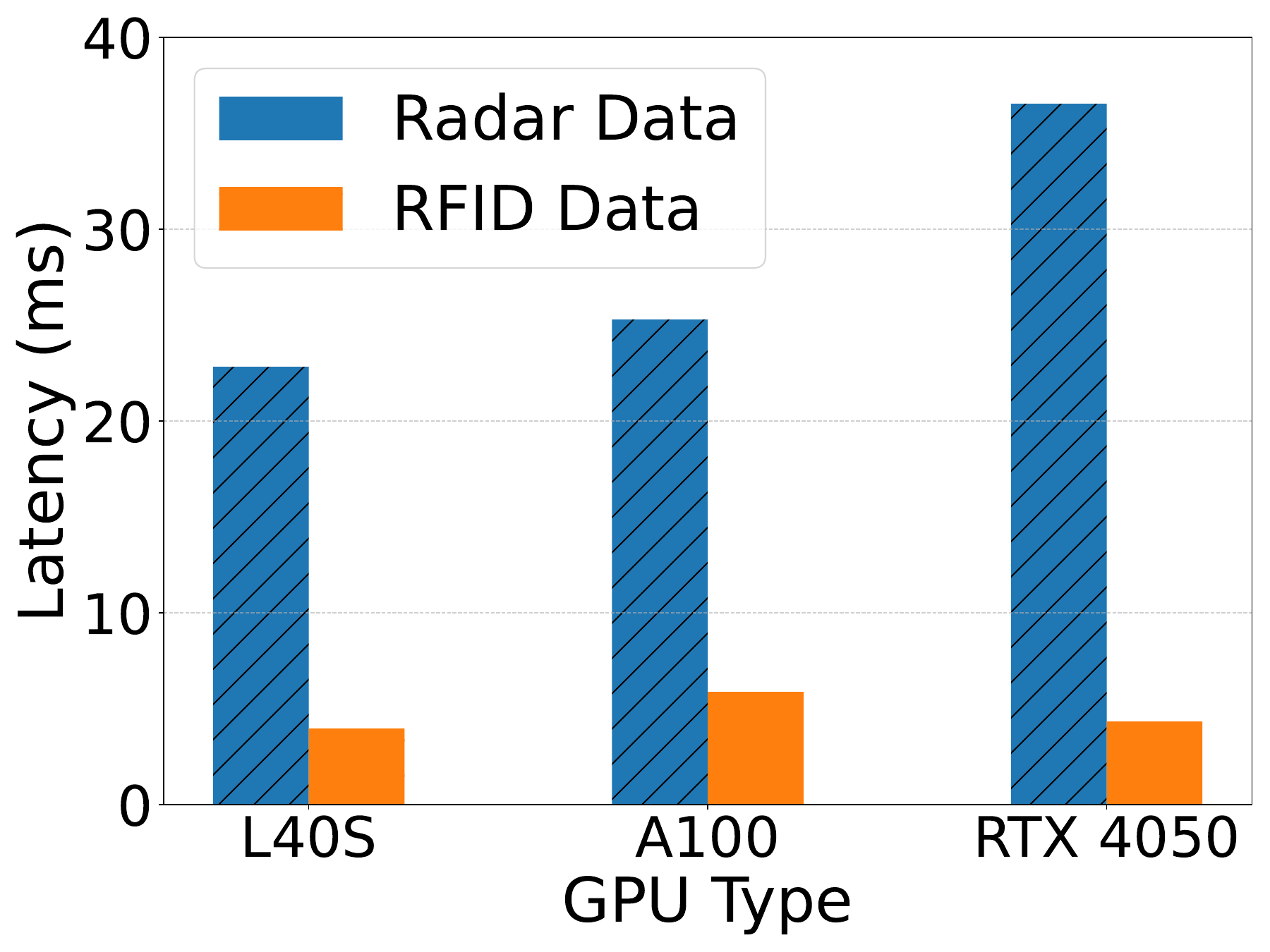}
        \caption{\edited Frame level preprocessing latency on mmWave radar data and RFID data. }\label{fig:exp_frame_latency}
    \Description{Inference Latency}
    \end{minipage}
\end{figure}

\begin{figure*}[t]
  \centering
\includegraphics[width=0.7\linewidth]{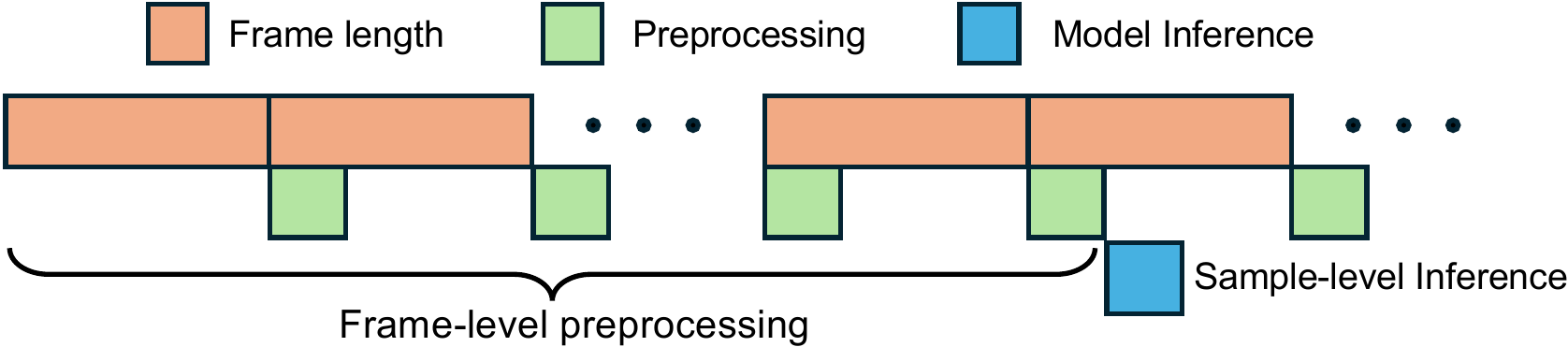}
  \caption{\edited Streaming pipeline of \sysname{} for real-time processing. \label{fig:realtime_pipeline}}
  \Description{HOI Categories. }
\end{figure*}

{\edited 
We then analyze the latency on systems equipped with either an NVIDIA L40S, NVIDIA A100, or (on the laptop) an NVIDIA RTX 4050 GPU.
All experiments on the server are conducted using a single GPU, with 8 CPU cores and 128 GB RAM allocated per job.
As illustrated in Fig.~\ref{fig:exp_decoupled_latency}, both model inference and preprocessing are fastest on the L40S, followed by the A100, with the RTX 4050 being the slowest.
On the laptop, the end-to-end latency averages 1.79 seconds per sample. Since the shortest sample duration in our dataset is 2.40 seconds, these results demonstrate the feasibility of \sysname{} for supporting real-time HOI recognition. 
In addition, Fig.~\ref{fig:exp_decoupled_latency} shows that the system latency is dominated by the preprocessing stage, while the model inference itself is relatively lightweight, taking only 36.86~ms per sample on the RTX~4050 GPU. We note that GPU performance is workload-dependent: in our preprocessing-dominated pipeline, the L40S shows better efficiency, likely due to architectural differences and optimizations for inference-oriented workloads.

To further assess the system’s responsiveness in an online setting, we analyze the frame-level preprocessing latency of each sensing modality in Fig.~\ref{fig:exp_frame_latency}. As shown in the figure, the per-frame preprocessing cost of mmWave radar data ranges from 22.83~ms to 36.52~ms across different GPUs, while the corresponding cost for RFID data remains below 6~ms. 
The effective acquisition interval of an RFID frame—corresponding to one complete read of all tags—is not strictly fixed, but is typically close to 100~ms in our system. With a 100~ms frame interval for the mmWave radar and a comparable effective acquisition interval for RFID data, the frame-level preprocessing latency is well within the real-time budget. Therefore, by performing preprocessing incrementally as each frame arrives in parallel to data acquisition, the system can further improve its responsiveness in an online setting, as illustrated in Fig.~\ref{fig:realtime_pipeline}.

}

\subsection{Case Study: Goal Inference}
To demonstrate the real-world applictions of \sysname{}, we apply \sysname{} to infer humans' goals in household scenarios. Such goal inference capacity is crucial for engineering assistive home systems to help humans without the need for privacy-intrusive sensors, e.g., cameras.
As is illustrated in Fig~\ref{fig:intro}, given a sequence of human-object interactions observed so far, we ask a goal inference model to infer the goal of the human. Unlike prior work, we leverage \sysname{} to recognize the HOI sequence from privacy-preserving sensors instead of relying on video inputs. Following the baselines proposed in \cite{ying2025siftom}, we take LLMs as the goal inference models. We provide a predefined goal space for LLMs to choose from \cite{puig2023nopa} and provide some pairs between HOI sequences and goals for few-shot prompting. The goal inference accuracy is defined as the proportion of correctly inferred sequences to the total number of sequences.  We present our prompt as follows:

\begin{tcolorbox}
Human has been performing a task that involves executing a sequence of interactions. The goal of task can only be one of the following: \{\texttt{Candidate goals}\}. 

You are a helpful assistant. In order to help human, please infer the most corresponding goals selected from the goal space for the following interaction sequences, in order: \{\texttt{Test HOI sequences}\}. 

Hints:
There is a set of example pairs showing the relationships between interaction sequences and their corresponding goals: \{\texttt{Few-shot prompting pairs between HOI sequences and goals}\}. 

The order of actions in each sequence is very important. Each sequence has a one-to-one correspondence with its goal.

Output requirement:
For each provided sequence, you have to choose the most relevant goal from the goal space, even if you are uncertain.
\end{tcolorbox}

\begin{table}[h]
\centering
\caption{Five examples of HOI sequences and corresponding goals.\label{tab:goalspace}}
\begin{tabular}{cc}
\toprule
HOI sequences & Goals \\
\midrule
\multirow{2}{*}{
\makecell{\texttt{[}<stand up, chair>, <pick up, book>,\\ <open, book>, <pick up, pen>\texttt{]}}}
    & \multirow{2}{*}{standing and taking notes in book with pen }\\ \\
\midrule
\multirow{2}{*}{
\makecell{\texttt{[}<sit down, chair>, <stand up, chair>,\\ <pick up, phone>, <put down, phone>\texttt{]}}}
    & \multirow{2}{*}{standing and answering the phone} \\ \\
\midrule
\multirow{2}{*}{
\makecell{\texttt{[}<open, microwave>, <close, microwave>,\\ <open, microwave>, <close, microwave>\texttt{]}}}
    & \multirow{2}{*}{using microwave} \\ \\
\midrule
\multirow{2}{*}{
\makecell{\texttt{[}<pick up, bottle>, <open, cabinet>,\\ <put down, bottle>, <close, cabinet>\texttt{]}}}
    & \multirow{2}{*}{trying to put bottle in cabinet} \\ \\
\midrule
\multirow{2}{*}{
\makecell{\texttt{[}<wipe, table>, <pick up, bottle>, \\<put down, bottle>, <wipe, table>\texttt{]}}}
    & \multirow{2}{*}{cleaning table} \\ \\
\bottomrule
\end{tabular}
\end{table}

In total, we define 30 different pairs between HOI sequences and goals, where each HOI sequence may have four or five HOIs. Table \ref{tab:goalspace} shows five pairs as an illustration. 
Among all 30 pairs, eight pairs are for few-shot prompting, which is designed to help the LLM understand the goal inference task. The remaining 22 pairs are for testing.
During the test, we ask the LLM to infer the corresponding goals for the HOI sequences in the test pairs from all 30 candidate goals.

We ask one volunteer to perform according to the 22 test HOI sequences in both the working space and the kitchen. We collect the corresponding RF data and get the recognized HOIs by \sysname{}. We compare three types of input for the goal inference:\\
(1) \textbf{Action-only} only inputs sequences of actions recognized by \sysname{} to the LLM.\\
(2) \textbf{Object-only} only inputs sequences of objects recognized by \sysname{} to the LLM.\\
(3) \textbf{\sysname{}} inputs sequences of full HOI categories recognized by \sysname{} to the LLM.

We evaluate the performance of the three methods on three LLMs: Gemini 2.5 Flash \cite{google_gemini25_flash}, GPT-5 \cite{openai_gpt5_chatgpt}, and DeepSeek-V3.2 \cite{deepseek_website}, and present the results in Table \ref{tab:caseres}. We can see that for all LLMs, \sysname{} brings higher goal inference accuracy than Action-only and Object-only. Specifically, on average, \sysname{} outperforms action-only by 68.18\% and object-only by 22.72\% in goal inference accuracy.

We observe that, in the action-only setting, most failures stem from misidentifying the target object, whereas in the object-only setting, errors primarily arise from misinterpreting the intended action. For example, a sequence like [<open>, <close>, <open>, <close>] can lead the LLM to confuse using the microwave with opening and closing a cabinet. Conversely, a sequence like [<book>, <book>, <book>, <book>] can cause the LLM to infer merely checking books rather than reading them, due to the absence of the <open> action. These results demonstrate the effectiveness of \sysname{} when adapted to downstream applications and highlight the importance of jointly recognizing actions and objects for a better user understanding.

\begin{table*}[h]
\centering
\caption{Goal Inference Accuracy \label{tab:caseres}}
\begin{tabular}{cccc}
\toprule
\diagbox{LLM}{Input} &  Action-only & Object-only & \sysname{}  \\
\midrule
Gemini 2.5 Flash & 18.18\% & 59.09\% &95.45\% \\
GPT-5 & 27.27\%&77.27\%&86.36\% \\
DeepSeek-V3.2 & 18.18\%&63.64\%&86.36\% \\
\bottomrule
\end{tabular}
\end{table*}

\section{Related Work}
\subsection{Vision-based Human-Object Interaction Recognition}
Over the past decade, significant effort has been devoted to camera-based HOI recognition~\cite{gupta2009observing,yao2010modeling,chao2015hico,luo2023detection}, given its importance for scene and human understanding. Many subsequent works~\cite{gkioxari2018detecting,hu2018relation,qi2018learning} extend object detection frameworks, typically structuring HOI recognition as a multi-task problem using a two-branch network: one branch detects humans and objects, while the other classifies actions and infers target objects through human-object relationships. Recent advancements include the use of sophisticated model architectures such as Transformers~\cite{li2024disentangled} and Mamba~\cite{xu2025hoimamba}. Other approaches introduce multimodal techniques, incorporating modalities such as text prompts~\cite{yang2024open}. 
Specifically, RF-camera~\cite{liu2021rfid} fuses vision and RFID modalities to recognize human gestures and achieve human-object matching. 
In addition, there is growing interest in collecting 3D HOI scene mesh data~\cite{diller2024cg, xu2025interact,peng2025hoi,huang2025hoigpt,jiang2024scaling,jiang2024autonomous}, advancing HOI understanding from 2D to 3D environments. 

However, vision-based approaches raise privacy concerns for in-home user sensing, and often suffer under poor lighting conditions. In contrast, \sysname{} leverages RF modalities, which are less invasive to privacy and robust to ambient lighting variations.

{\edited
\subsection{Wireless Sensing}
The advancement of deep learning has facilitated rapid progress in wireless sensing systems for human activity recognition~\cite{wang2015understanding,singh2019radhar,wang2021m,zhang2021widar3,liu2022mtranssee,yang2023slnet,zhao2023cubelearn,cao2024mmclip,wang2024xrf55,zhang2024super}. 
To achieve vision-level perception while preserving user privacy, prior work has explored various wireless modalities, including mmWave~\cite{singh2019radhar,wang2021m,liu2022mtranssee,zhao2023cubelearn,cao2024mmclip,zhang2024super}, RFID~\cite{wang2018modeling,huang2019id,sun2024lodihar,wang2024xrf55,feng2026rfusion}, and WiFi~\cite{wang2015understanding,zhang2021widar3,yang2023slnet}, for sensing human activities. 
To improve user convenience, most systems operate in device-free settings, although some RFID-based approaches require tags attached to the user’s body~\cite{wang2016toward,jin2018towards,yang2022environment,wang2025generative,wu2026rf,wang2026rfgat}. 
However, identifying specific objects in such settings remains highly challenging~\cite{liu2019concealed,zhang2020mmeye}, let alone jointly recognizing both human actions and their corresponding target objects.

To address the challenge of object identification, a line of prior work adopts device-based approaches by attaching RFID tags or similar markers to objects and inferring user actions from changes in the communication channel between tags and readers~\cite{li2015idsense,spielberg2016rapid,pradhan2017rio,wang2018modeling,zhang2019shopeye,gao2019livetag}. These approaches avoid instrumenting users and have been widely explored in RFID-based sensing. However, they often struggle to distinguish actions that do not induce significant channel variations, such as interactions without object movement or direct tag coverage, thereby limiting the diversity of recognizable actions.
RF-Diary~\cite{fan2020home} combines radar sensing with a pre-defined floor map of object locations to recognize HOI. While effective, maintaining accurate object maps in everyday environments remains challenging.

Other works~\cite{laput2019sensing,lee2024echowrist,lee2025grab} employ acoustic signals collected by wearable devices to sense HOI, yet these approaches require wearables and fail to capture whole body movements, such as stand and sit. Recent works~\cite{EgoADL,XRFV2} explore HOI recognition with modality fusion, but their solutions require access to personal devices, such as audio and IMU data from smartphone, while \sysname{} does not assume any devices on users.

In contrast, \sysname{} combines the strengths of both strategies by fusing mmWave and RFID data, enabling effective recognition of both human actions and target objects without compromising user convenience.
 }
\subsection{Wireless Signal Synthesis for Sensing}
Recent studies have explored the use of synthetic wireless signals to augment training data for human activity recognition. Technically, these works employ either ray tracing-based methods~\cite{korany2019xmodal,deng2023midas++,rfgen,cao2024mmclip,li2024sbrf} or data-driven approaches~\cite{ahuja2021vid2doppler,deng2023midas,chi2024rf}. 
As a physics-based method, ray tracing models signal propagation by computing paths
affected by reflection,  diffraction, and scattering 
using physics and geometry formulas. In contrast, the simulation of data-driven approaches usually relies on empirical models, which lack physical interpretability, and may produce unrealistic outputs without physical priors.

However, prior research typically models only the human body as the primary signal reflector, treating other objects in the environment as interference. This makes it difficult to extend such techniques to HOI recognition tasks. In contrast, \sysname{} explicitly simulates separate signals from both the human body and target objects, enabling the generation of multimodal synthetic data tailored for HOI recognition.

\section{Discussion and Future Work}

\subsection{Increase the Size and Diversity of the Data}
As an initial effort to explore RF-based sensing for HOI recognition, \sysname{} collects data at a moderate scale and focuses on a set of common HOI categories encountered in daily life under diverse setups and sensing conditions.
{\minor Compared with large-scale vision-based HOI datasets, our real-world dataset remains limited in scale, particularly in the number of participants, object categories, and interaction types. 
This limitation comes from the cost of collecting mmWave--RFID data under various real-world setups, and motivates future efforts toward larger community benchmarks for RF-based HOI recognition.}
We acknowledge that expanding the number of HOI categories and participants is an important direction toward broader real-world deployment.
However, extending to entirely new action categories in the current simulation pipeline is constrained by the availability of realistic motion priors in the underlying TRUMANS dataset, which limits the ability to synthesize physically plausible RF data for additional action categories.
The selected action categories (e.g., pick, put, open, and close) are therefore intentionally chosen as fundamental interaction primitives.
These primitives are sufficient to support a range of downstream tasks under state-of-the-art frameworks, such as goal inference~\cite{puig2023nopa,zhang2025autotom,ying2025siftom}. Rather than solely increasing the category count in the training set, we evaluate \sysname{} along more challenging generalization axes, including unseen object categories and unseen action-object combinations, without using any additional real-world training data for new categories.

In future works, the collected data can be further expanded in both size and diversity and can include more HOI categories to support broader applications and more comprehensive evaluations.
We observe that there is a growing number of motion capture-based mesh datasets \cite{jiang2024autonomous,zhang2024hoi,lu2025humoto}.  In parallel, recent advances in HOI mesh synthesis~\cite{diller2024cg, xu2025interact,peng2025hoi,huang2025hoigpt,yao2026hosig,mu2026fantasyhsi,liconta} enable generative models to more faithfully represent the dynamics of human-object interactions, paving the way for large-scale RF data generation to support HOI recognition tasks. Accordingly, \sysname{} has the potential to enlarge its synthetic data by utilizing more mesh datasets from either motion capture or generative models.
Furthermore, in terms of the real-world data, we envision that \sysname{} will inspire future research on RF-based HOI recognition using custom-built real-world test sets. 
Specifically, the decoupled representation of actions and objects enables systematic expansion of HOI categories by combining existing primitives with new object categories, reducing the need for exhaustive data collection. 
The proposed recognition model can serve as a baseline to quantify the difficulty of the dataset, while the insights from our data collection provide practical guidance for designing future RF-based HOI datasets.
Although the data that a single work can contribute may be limited, we may integrate these efforts into a universe benchmark, including more HOI categories and setups.

\subsection{Downstream Application}
HOI recognition serves as a crucial component in various downstream applications, including VR, AR, and assistive robots. 
In VR \cite{canales2020performance}, recognizing an HOI enables the system to generate appropriate sensory feedback—such as the sound of a closet opening—enhancing user immersion. 
In AR \cite{jain2023ubi}, the recognized HOI can trigger updates to the virtual environment, such as overlaying the text to write when the user picks up a pen. In general, \sysname{} can enhance the sense of realism in both VR and AR. 
Besides, as shown in our case study, HOI sequences can help robots to infer the user's goal. Then assistive robots can provide help correspondingly to reach the goal.
For instance, observing the sequence [<sit down, chair>, <stand up, chair>, <pick up, book>, <put down, book>] may lead the robot to infer that the user intends to check books, prompting it to deliver another book to the user for checking. 
These scenarios demonstrate the potential of integrating \sysname{} as a core module in interactive systems that require semantic understanding of HOI.
{\minor The above applications are more readily deployable in environments where RFID infrastructure is already available or can be naturally integrated, such as smart homes, healthcare facilities, assisted living spaces, and retail environments with tagged objects.
However, deploying \sysname{} in fully uninstrumented environments would require additional tagging effort and reader placement, which may limit its immediate applicability. 
Future work can reduce this deployment burden by exploring opportunistic RFID infrastructure, lower-cost tagging strategies, or complementary object-identification cues.
}

\section{Conclusion}
This paper proposes \sysname{}, a novel HOI recognition system that leverages only RF signals. By fusing mmWave and RFID data, \sysname{} harnesses the strengths of different modalities, making it particularly well-suited for HOI recognition applications. To enhance generalizability, \sysname{} uses a physics-based simulator to generate synthetic mmWave and RFID data, pre-trains the model on these synthetic data, and subsequently fine-tunes it with a small amount of real-world data. Extensive experiments demonstrate that \sysname{} consistently outperforms single-modality baselines, with synthetic data yielding significant performance improvements when real-world training samples or setups are limited, thus reducing data collection costs. We anticipate that \sysname{} will inspire further research in wireless sensing for HOI recognition.

\begin{acks}
We thank the anonymous reviewers for their constructive feedback. We also thank Zhehao Zhang and Jienan Chen for their help with data collection.
\end{acks}

\bibliographystyle{ACM-Reference-Format}
\bibliography{aa-ref}

\end{document}